\documentclass[journal]{IEEEtran}
\usepackage{lineno}
\modulolinenumbers[5]

\usepackage{xpatch}

\usepackage{bm}
\usepackage{array}
\usepackage{graphicx}
\usepackage{amsmath,amssymb,amsthm}
\usepackage{siunitx}
\usepackage{algpseudocode}
\usepackage{algorithmicx}
\usepackage{algorithm}
\usepackage{booktabs}
\usepackage{adjustbox}
\usepackage{color}
\usepackage{changepage}
\usepackage{xr}
\usepackage{xr-hyper}
\usepackage{placeins}

\usepackage{graphicx}

\usepackage{tabularx}
\usepackage{subfig}
\usepackage{caption}
\usepackage{multirow}
\usepackage{float}
\usepackage{soul}
\usepackage[hidelinks]{hyperref}
\usepackage[numbers,sort&compress]{natbib}
\usepackage{bigstrut} 
\usepackage[table]{xcolor} 
\usepackage{enumitem} 
\usepackage{subfig}
\usepackage{listings} 
\usepackage{subcaption}
\definecolor{dkgreen}{rgb}{0,0.5,0}
\definecolor{gray}{rgb}{0.5,0.5,0.5}
\definecolor{mauve}{rgb}{0.58,0,0.82}
\definecolor{dkblue}{rgb}{0,0,0.6}

\renewcommand{\algorithmicrequire}{\textbf{Input:}}
\renewcommand{\algorithmicensure}{\textbf{Output:}}

\AtBeginDocument{
 \abovedisplayskip=5pt plus 4pt minus 2pt
 \abovedisplayshortskip=5pt plus 4pt minus 4pt
 \belowdisplayskip=5pt plus 4pt minus 2pt
 \belowdisplayshortskip=5pt plus 4pt minus 4pt
}

\ifCLASSINFOpdf
\else
\fi

\begin{document}

\captionsetup{font={footnotesize}}
\captionsetup[table]{labelformat=simple, labelsep=newline, textfont=sc, justification=centering}

\title{Unleashing the Potential of Differential Evolution through Individual-Level Strategy Diversity}

\author{Chenchen Feng, Minyang Chen, Zhuozhao Li,
        and Ran Cheng
        \thanks{
        Chenchen Feng is with the Department of Computer Science and Engineering, Southern University of Science and Technology, Shenzhen 518055, China. E-mail: chenchenfengcn@gmail.com. 
        }
        \thanks{
        Minyang Chen was with the Department of Computer Science and Engineering, Southern University of Science and Technology, Shenzhen 518055, China. E-mail: cmy1223605455@gmail.com. }
        \thanks{
        Zhuozhao Li is with the Department of Computer Science and Engineering, Southern University of Science and Technology, Shenzhen 518055, China. E-mail: lizz@sustech.edu.cn. 
        }
        \thanks{
        Ran Cheng is with the Department of Data Science and Artificial Intelligence, and the Department of Computing, The Hong Kong Polytechnic University, Hong Kong SAR, China. E-mail: ranchengcn@gmail.com. (\emph{Corresponding author: Ran Cheng}).
        }
        }

\markboth{Bare Demo of IEEEtran.cls for IEEE Journals}
{Shell \MakeLowercase{\textit{et al.}}: Bare Demo of IEEEtran.cls for IEEE Journals}

\maketitle

\begin{abstract}
Since Differential Evolution (DE) is sensitive to strategy choice, most existing variants pursue performance through adaptive mechanisms or intricate designs.
While these approaches focus on adjusting strategies over time, the structural benefits that static strategy diversity may bring remain largely unexplored.
To bridge this gap, we study the impact of individual-level strategy diversity on DE’s search dynamics and performance, and introduce \textbf{iStratDE} (DE with individual-level strategies), a minimalist variant that assigns mutation and crossover strategies independently to each individual at initialization and keeps them fixed throughout the evolutionary process.
By injecting diversity at the individual level without adaptation or feedback, iStratDE cultivates persistent behavioral heterogeneity that is especially effective with large populations.
Moreover, its communication-free construction possesses intrinsic concurrency, thereby enabling efficient parallel execution and straightforward scaling for GPU computing.
We further provide a convergence analysis of iStratDE under standard reachability assumptions, which establishes the almost-sure convergence of the best-so-far fitness.
Extensive experiments on the CEC2022 benchmark suite and robotic control tasks demonstrate that iStratDE matches or surpasses established adaptive DE variants.
These results highlight individual-level strategy assignment as a straightforward yet effective mechanism for enhancing DE’s performance.
The source code of iStratDE is publicly accessible at: \url{https://github.com/EMI-Group/istratde}.
\end{abstract}


\IEEEpeerreviewmaketitle\section{Introduction}
\IEEEPARstart{D}{ifferential} Evolution (DE), introduced by Storn and Price~\cite{DE1997}, is a widely adopted population-based algorithm for continuous optimization. Its appeal lies in the simplicity of its operators (mutation, crossover, and selection), which are governed by only three parameters: population size (\(N\)), scaling factor (\(F\)), and crossover rate (\(CR\)). Owing to this minimalist design, DE has delivered reliable performance across a broad spectrum of optimization problems~\cite{DEsurvey2011, DEsurvey2016, pant2020differential, chiong2012variants} and remains a strong contender in benchmark competitions, including those organized by the IEEE Congress on Evolutionary Computation (CEC)~\cite{sun2021learning}.A persistent challenge, however, lies in choosing suitable parameters and strategies. The No Free Lunch (NFL) theorems imply that no single configuration is uniformly optimal across problem domains~\cite{NFL}. In response, a substantial body of work has sought to enhance performance and adaptability through algorithmic variation. Representative examples include SaDE~\cite{SaDE2008}, JADE~\cite{JADE2009}, EPSDE~\cite{mallipeddi2011differential}, EDEV~\cite{EDEV2018}, and the LSHADE family~\cite{SHADE2013, LSHADE2014, JSO2017, LSHADE-RSP2018}. These approaches typically incorporate adaptation mechanisms that dynamically tune parameters or switch among multiple strategies based on historical success: SaDE maintains an adaptively selected strategy pool; JADE introduces historical learning via an archive; and LSHADE and its successors further improve robustness through population resizing, parameter memories, and ensemble-style search.

Although these methods have achieved considerable success, they often add layers of algorithmic complexity. Operating predominantly at the \emph{population level}, they depend on centralized feedback and bookkeeping, which leads to several practical limitations. 
First, the proliferation of interacting components and hyperparameters can undermine generality and robustness when prior knowledge is scarce or problem characteristics vary widely.
Second, population-wide adaptation tends to lose effectiveness in high-dimensional or large-scale regimes, where cross-individual dependencies grow rapidly and global statistics become less informative~\cite{yang2007differential, mahdavi2015metaheuristics}. 
Third, synchronization barriers and reliance on global archives hinder fine-grained parallelism on modern accelerators, which results in poor utilization and control-flow divergence on GPUs~\cite{wang2013parallel}.

These observations motivate a re-examination of DE’s core design assumptions, particularly those concerning {strategy assignment}. 
Whereas most methods employ a uniform or adaptively shared strategy at the \emph{population level}, the alternative of assigning strategies independently at the \emph{individual level} (even in a non-adaptive manner) has been largely overlooked. 
This perspective suggests a path to behavioral diversity with minimal algorithmic complexity and strong potential for parallel scalability.

We revisit strategy assignment in DE by {studying} the effect of individual-level strategy diversity on search dynamics and performance, and {operationalize} this perspective via \textbf{iStratDE} (DE with individual-level strategy). Concretely, iStratDE assigns each individual its mutation and crossover strategies (together with their parameters)  \emph{randomly and independently} at initialization and then keeps them fixed throughout evolution. 
This construction induces \emph{individual-level structural diversity}: heterogeneous search behaviors arise without feedback, archives, or dynamic tuning, while the lack of cross-individual dependencies makes it naturally compatible with fine-grained parallel execution.The design of iStratDE yields a simple intuition: a population of independently randomized strategies functions as an implicit ensemble, thereby diversifying trajectories, broadening exploration, and mitigating premature convergence through decentralized variation. 
Rather than layering additional adaptive machinery, {iStratDE} attains behavioral richness by injecting fixed heterogeneity at the structural level. 
Moreover, the same independence that sustains diversity also confers intrinsic concurrency: the absence of global adaptation, archives, and historical statistics enables {iStratDE} to perform communication-free and fully parallel updates, thereby sidestepping the synchronization bottlenecks typical of adaptive variants.
In summary, the main contributions of this work are as follows:
\begin{itemize}
    \item We {study} individual-level strategy diversity in DE and {instantiate} this perspective as {iStratDE}, a simple yet effective DE variant that assigns mutation and crossover strategies (and parameters) randomly and independently to each individual at initialization, thereby achieving algorithmic diversity without feedback, archives, or dynamic tuning.
    \item We leverage iStratDE's communication-free design to develop a GPU-accelerated implementation that exploits the resulting intrinsic concurrency, thereby enabling fully parallel updates with minimal synchronization and high hardware utilization.
    \item  We provide a convergence analysis of iStratDE under standard assumptions and conduct extensive experiments on the CEC2022 benchmark suite and robotic control tasks, where {iStratDE} matches or surpasses strong adaptive DE variants and exhibits favorable scaling behavior.
\end{itemize}

The remainder of the paper is organized as follows. Section~\ref{section_Preliminary} reviews the related work. Section~\ref{section_The proposed iStratDE} presents iStratDE and its convergence analysis. Section~\ref{section_Experimental_study} reports experimental results. Section~\ref{section Conclusion} concludes the work.\section{Related Work}\label{section_Preliminary}
\subsection{Overview of DE}\label{subsection_DE overview}Differential Evolution (DE) is a population-based optimization algorithm introduced by Storn and Price~\cite{DE1997}. It is designed to solve complex optimization problems by iteratively improving candidate solutions through three core operations: mutation, crossover, and selection.
DE variants adhere to a standardized naming scheme, \texttt{DE/x/y/z}, where \texttt{x} denotes the base vector used for mutation, \texttt{y} specifies the number of differences involved, and \texttt{z} represents the crossover method applied.
For instance, the DE variant described in Algorithm \ref{Alg_DE} is referred to as \texttt{DE/rand/1/bin}.
We now detail the core components of DE and provide a step-by-step breakdown of its operational process.\begin{algorithm}
\small
\caption{Differential Evolution (DE)}
\label{Alg_DE}
\begin{algorithmic}[1]
  \Require $D$, $N$, $G_{\max}$, $F$, $CR$, $\boldsymbol{lb}$, $\boldsymbol{ub}$
  \State Initialize $\boldsymbol{X} = \{\boldsymbol{x}_1, \dots, \boldsymbol{x}_{N}\}$ within $[\boldsymbol{lb}, \boldsymbol{ub}]$
  \State Evaluate $\boldsymbol{y} = \{ f(\boldsymbol{x}_1), \dots, f(\boldsymbol{x}_{N}) \}$
  \State $g \gets 0$
  \While{$g < G_{\max}$}
    \For{$i = 1$ to $N$}
      \State Select $r_1, r_2, r_3 \in \{1, \dots, N\} \setminus \{i\}$
      \State $\boldsymbol{v}_i \gets \boldsymbol{x}_{r_1} + F\,(\boldsymbol{x}_{r_2} - \boldsymbol{x}_{r_3})$
      \State Choose $j_{\text{rand}} \in \{1, \dots, D\}$
      \For{$j = 1$ to $D$}
        \If{$\operatorname{rand}(0, 1) \leq CR$ \textbf{ or } $j = j_{\text{rand}}$}
          \State $\boldsymbol{u}_i[j] \gets \boldsymbol{v}_i[j]$
        \Else
          \State $\boldsymbol{u}_i[j] \gets \boldsymbol{x}_i[j]$
        \EndIf
      \EndFor
      \If{$f(\boldsymbol{u}_i) \leq f(\boldsymbol{x}_i)$}
        \State $\boldsymbol{x}_i \gets \boldsymbol{u}_i$
      \EndIf
    \EndFor
    \State $g \gets g + 1$
  \EndWhile
  \State \Return $\boldsymbol{x}_{\text{best}}$, $f(\boldsymbol{x}_{\text{best}})$
\end{algorithmic}
\end{algorithm}
\subsubsection{Initialization}

The DE algorithm commences with the initialization of a population of $N$ individuals. Each individual $\boldsymbol{x}_i$, where $i \in \{1, 2, \dots, N\}$, represents a potential solution in a $D$-dimensional search space. The initial population is generated randomly within predefined lower and upper bounds:\begin{align}
\boldsymbol{x}_i \in [\boldsymbol{lb}, \boldsymbol{ub}], \quad \forall i = 1, 2, \dots, N.
\end{align}Here, $\boldsymbol{lb}$ and $\boldsymbol{ub}$ represent the lower and upper bounds for each dimension, thereby ensuring that all solutions are feasible within the problem's constraints. The choice of initialization technique can significantly affect the diversity and convergence speed of DE, as discussed extensively in existing literature \cite{kazimipour2014review}.
Each individual in the population is initialized as:\begin{align}
\boldsymbol{x}_i = \left[ x_{i, 1}, x_{i, 2}, \dots, x_{i, D} \right].
\end{align}The fitness of each individual is evaluated using an objective function $f(\boldsymbol{x}_i)$, which quantifies solution quality. The algorithm is designed to minimize (or maximize) this fitness function over generations:\begin{align}
f: \mathbb{R}^D \rightarrow \mathbb{R}.
\end{align}This initial population provides the starting point for the evolutionary process.

\subsubsection{Mutation}

Mutation introduces diversity into the population by perturbing existing solutions, thereby enabling the algorithm to explore new areas of the search space. Several mutation strategies are available, each balancing exploration and exploitation differently \cite{price2006differential}. The following sections introduce several common mutation strategies used in DE.

\paragraph{DE/rand/1}
In the \texttt{DE/rand/1} strategy, the mutant vector is generated using three randomly selected individuals:\begin{align}
\boldsymbol{v}_i = \boldsymbol{x}_{r_1} + F\,(\boldsymbol{x}_{r_2} - \boldsymbol{x}_{r_3}),
\end{align}where $r_1, r_2, r_3 \in \{1, 2, \dots, N\}$ are distinct indices, and $F \in [0, 2]$ is the scaling factor that controls the magnitude of the difference vector. This strategy promotes exploration by introducing randomness into the search.

\paragraph{DE/best/1}
In contrast, the \texttt{DE/best/1} strategy biases the search toward the best-known solution:\begin{align}
\boldsymbol{v}_i = \boldsymbol{x}_{\text{best}} + F\,(\boldsymbol{x}_{r_1} - \boldsymbol{x}_{r_2}),
\end{align}where $\boldsymbol{x}_{\text{best}}$ is the individual with the best fitness in the current population. This approach enhances exploitation, but it risks premature convergence.

\paragraph{DE/current/1}
In the \texttt{DE/current/1} strategy, the mutation is focused around the current individual:\begin{align}
\boldsymbol{v}_i = \boldsymbol{x}_{i} + F\,(\boldsymbol{x}_{r_1} - \boldsymbol{x}_{r_2}),
\end{align}where $\boldsymbol{x}_{i}$ is the individual undergoing mutation. This method promotes local exploration but limits the search's global scope.

\paragraph{DE/current-to-pbest/1}
To address this limitation, the \texttt{DE/current-to-pbest/1} strategy combines both exploration and exploitation by using the current individual and a top-performing individual selected from the best $p\%$ of the population:\begin{align}
\boldsymbol{v}_i = \boldsymbol{x}_{i} 
+ F\,(\boldsymbol{x}_{\text{pbest}} - \boldsymbol{x}_{i}) 
+ F\,(\boldsymbol{x}_{r_1} - \boldsymbol{x}_{r_2}),
\end{align}where $\boldsymbol{x}_{\text{pbest}}$ is randomly selected from the top $p\%$ of the population, and $\boldsymbol{x}_{i}$ is the current individual. This strategy achieves a balance between local and global search.

\paragraph{DE/current-to-rand/1}
The \texttt{DE/current-to-rand/1} strategy enhances exploration by combining the current individual with a randomly selected individual from the population:\begin{align}
\boldsymbol{v}_i = \boldsymbol{x}_i 
+ F\,(\boldsymbol{x}_{r_1} - \boldsymbol{x}_i) 
+ F\,(\boldsymbol{x}_{r_2} - \boldsymbol{x}_{r_3}),
\end{align}where $\boldsymbol{x}_i$ denotes the current individual, and $\boldsymbol{x}_{r_1}$, $\boldsymbol{x}_{r_2}$, and $\boldsymbol{x}_{r_3}$ are mutually distinct random individuals selected from the population. This mutation scheme significantly improves global search performance by guiding the individual toward random directions, thereby reducing the risk of premature convergence.

These mutation strategies allow DE to be flexible, thereby adapting its behavior to suit different types of optimization problems by tuning the balance between exploration and exploitation.

\subsubsection{Crossover}
After mutation, DE applies crossover to combine the mutant vector $\boldsymbol{v}_i$ with the current target vector $\boldsymbol{x}_i$, thereby creating a trial vector $\boldsymbol{u}_i$ by exchanging elements. Crossover aims to produce superior offspring by integrating parental components, thereby promoting diversity and balancing exploration and exploitation. Common DE crossover techniques, such as binomial, exponential, and arithmetic, significantly influence search behavior \cite{zaharie2009influence}.

\paragraph{Binomial Crossover}
The binomial crossover \texttt{bin} operates element-wise between 
$\boldsymbol{x}_i$ and $\boldsymbol{v}_i$:\begin{align}
u_{i, j} =
\begin{cases}
v_{i, j}, & \text{if } \text{rand}(0, 1) \leq CR \text{ or } j = j_{\text{rand}} \\
x_{i, j}, & \text{otherwise},
\end{cases}
\end{align}where $CR \in [0, 1]$ is the crossover rate, which controls the likelihood of taking elements from the mutant vector $\boldsymbol{v}_i$. The index $j_{\text{rand}}$ ensures that at least one element from $\boldsymbol{v}_i$ is included in $\boldsymbol{u}_i$.

\paragraph{Exponential Crossover}
The exponential crossover \texttt{exp} selects a continuous segment from the mutant vector $\boldsymbol{v}_i$, starting from a random position $j_{\text{rand}}$. It copies exactly $L$ consecutive elements into the trial vector $\boldsymbol{u}_i$, and the remaining elements are copied from the target vector $\boldsymbol{x}_i$:\begin{align}
u_{i, j} =
\begin{cases}
v_{i, j}, & j \in \{j_{\text{rand}}, j_{\text{rand}}+1, \dots, j_{\text{rand}}+L-1\} \\
x_{i, j}, & \text{otherwise},
\end{cases}
\end{align}where indices are taken modulo $D$, and the number of elements \(L\) is determined probabilistically based on the crossover rate \(CR\).

\paragraph{Arithmetic Crossover}
The arithmetic crossover, i.e., \texttt{arith}, produces a trial vector by linearly combining $\boldsymbol{v}_i$ and $\boldsymbol{x}_i$:\begin{align}
\boldsymbol{u}_i = \alpha \boldsymbol{v}_i + (1 - \alpha) \boldsymbol{x}_i,
\end{align}where $\alpha \in [0, 1]$ is a random weight controlling the contribution from each vector. This crossover promotes smooth transitions between solutions, which makes it particularly useful for continuous optimization tasks.

The selection among these crossover techniques, and the meticulous setting of the $CR$ parameter, substantially shape DE's search dynamics and overall efficacy. While binomial crossover is widely adopted due to its component-wise recombination efficacy and generally robust performance across various problems, exponential and arithmetic crossovers offer alternative mechanisms for information exchange that might be advantageous for specific problem structures or search phases \cite{zaharie2009influence, mezura2006comparative}. Ultimately, these crossover techniques provide different ways of balancing exploration and exploitation, thereby enhancing the search capability of DE by influencing the diversity and evolutionary trajectory of the population.

\subsubsection{Selection}

The final step in each generation is the selection process, where the algorithm determines whether the trial vector $\boldsymbol{u}_i$ or the target vector $\boldsymbol{x}_i$ survives to the next generation. This process ensures that only better-performing solutions are retained, thereby gradually improving the overall population quality.

The selection mechanism is formulated as:\begin{align}
\boldsymbol{x}_i^{(t+1)} =
\begin{cases}
\boldsymbol{u}_{i}, & \text{if } f(\boldsymbol{u}_i) \leq f(\boldsymbol{x}_{i}) \\
\boldsymbol{x}_{i}, & \text{otherwise},
\end{cases}
\end{align}where $f(\boldsymbol{x})$ represents the objective function used to evaluate the quality of the solution $\boldsymbol{x}$. The trial vector $\boldsymbol{u}_i$ replaces the target vector $\boldsymbol{x}_i$ only if it yields a better objective value.

This greedy selection ensures fast convergence by retaining the best solutions from the current and previous generations, while maintaining diversity through mutation and crossover. By repeating this process over multiple generations, the population gradually evolves toward optimal solutions. 

However, as the dimensionality or population size increases, the computational burden escalates, which highlights the need for efficient parallelization. This challenge is addressed through GPU acceleration, as discussed in Section \ref{subsec:gpu_ea}.

\subsection{Adaptive and Distributed DE}\label{subsection:ada_dis_de}
To address the challenges of strategy selection and parameter tuning inherent in canonical DE, two dominant research paradigms have emerged: Adaptive Differential Evolution and Distributed Differential Evolution.
Adaptive DE departs from the traditional fixed-parameter approach by introducing mechanisms to dynamically adjust control parameters (e.g., $F$ and $CR$) and mutation strategies based on feedback gathered during the evolutionary process \cite{jDE2006, SaDE2008, JADE2009}.
Distributed DE, in contrast, employs a multi-population framework that partitions the main population into several sub-populations. These sub-populations evolve in parallel and periodically exchange individuals via a migration policy, which is a mechanism intended to promote population diversity and search efficacy \cite{weber2010study, weber2011study}.

Recent research in Distributed DE has focused on leveraging this multi-population structure for sophisticated algorithmic enhancements \cite{DEpapersurvey2020}. 
One prominent direction uses the framework to ensemble various DE variants or operators, as demonstrated in algorithms such as MPEDE \cite{MPEDE2015}, EDEV \cite{EDEV2018}, and IMPEDE \cite{IMPEDE}.
Another active research area employs these distributed populations for adaptive resource allocation, as exemplified by DDE-AMS \cite{DDE-AMS} and DDE-ARA \cite{DDE-ARA}.

The contemporary landscape of Adaptive DE is characterized by refined methodologies built upon foundational success-history mechanisms, as established by SHADE \cite{SHADE2013} and LSHADE \cite{LSHADE2014}. 
Recent advancements further refine these concepts: LSHADE-RSP \cite{LSHADE-RSP2018} introduces mechanisms for selective pressure, IMODE \cite{IMODE2020} focuses on integrating multiple distinct DE variants, and LADE \cite{LADE2023} automates the adaptation learning process.

\subsection{Paradigms of Strategy Assignment in DE}\label{subsection:non_adaptive_de}

Non-adaptive DE variants employ a fixed set of strategies and control parameters that are assigned at initialization and are not adjusted during the evolutionary process. A key advantage of this approach is its simplicity and low computational overhead. Lacking a continuous adaptation mechanism, these algorithms are computationally efficient, which makes them attractive for problems with tight computational budgets or limited resources.

However, a significant limitation of non-adaptive DE algorithms is their lack of flexibility in responding to dynamic changes in the optimization landscape. These algorithms are prone to premature convergence, particularly in complex or high-dimensional search spaces, as they cannot adjust to the evolving needs of the search. Without adaptation, they may either overexplore certain areas of the solution space or become trapped in local optima, which ultimately reduces their overall effectiveness.In response to these limitations, adaptive DE variants have become increasingly popular. These approaches introduce mechanisms for dynamically adjusting control parameters or strategies based on feedback from the evolutionary process. Although such adaptations have often led to performance improvements, they also introduce substantial algorithmic complexity. Consequently, the performance gains of many DE variants have been accompanied by a progressive increase in algorithmic complexity, a trend that deviates from the simplicity originally emphasized by the Differential Evolution paradigm.

As an alternative direction within the non-adaptive paradigm, iStratDE adopts a structurally decentralized approach. Rather than employing uniform or globally shared configurations, it assigns strategies and control parameters individually and randomly at initialization. These per-individual assignments remain fixed throughout evolution, thereby avoiding internal adaptation mechanisms. While iStratDE shares the non-adaptive nature of its predecessors, it differs in the granularity of strategy deployment across the population. The structural implications of this design are explored in Section \ref{section_The proposed iStratDE}.

\subsection{GPU-Accelerated Evolutionary Algorithms}\label{subsec:gpu_ea}
The population-based structure of evolutionary algorithms (EAs) naturally aligns with the capabilities of modern parallel computing architectures, particularly GPUs. In DE, for instance, core operations such as mutation, crossover, and selection are typically applied to each member of the population independently. This inherent characteristic directly facilitates concurrent computation across a large number of threads.

This structural independence has motivated growing interest in applying GPU acceleration to EAs, especially for high-dimensional or complex problems~\cite{tan2015survey, liang2024gpu, liang2025bridging}. When implemented appropriately, each individual in the population can be assigned to a separate GPU thread or block, thereby enabling efficient execution of the algorithm at scale.

Despite this potential, the extent to which a DE variant can benefit from GPU acceleration depends on its internal structure. Adaptive DE algorithms often incorporate feedback-based mechanisms, which introduce inter-individual dependencies that limit parallel efficiency due to synchronization requirements and control flow divergence. Although asynchronous evolutionary algorithms have been proposed to mitigate idle time and improve resource utilization in parallel environments~\cite{scott2015understanding}, they do not fundamentally eliminate the need for cross-individual communication. For example, the Fuzzy Adaptive Differential Evolution (FADE) algorithm~\cite{liu2005fuzzy} employs a fuzzy rule system to dynamically adjust parameters. Although this approach improves flexibility, it introduces global dependencies that make it less suitable for efficient parallel execution. In contrast, non-adaptive DE variants are generally more suitable for parallel execution due to their simpler control logic. When strategies and parameters are fixed per individual at initialization and remain unchanged during evolution, each individual's update can be computed independently, thereby reducing coordination overhead. Nevertheless, frameworks such as MetaDE~\cite{metade} have demonstrated that GPU acceleration can be successfully applied to DE, achieving high performance through parallel computation.

\section{Method}\label{section_The proposed iStratDE}\subsection{Motivation}\label{section_Motivation}

The field of DE has witnessed substantial progress, with numerous variants achieving strong performance on complex optimization problems by incorporating sophisticated strategies designed and adapted at the population level. 
Such designs steer the entire ensemble of solutions under a unified set of rules or dynamically adjusted parameters, thereby fostering coordinated progression toward promising regions of the search space~\cite{DEsurvey2011,DEsurvey2016}.

While this population-level coherence can be powerful, it also entails inherent limitations, particularly on challenging multimodal problems. 
When the behavior of the entire population is synchronized under a single strategy, even an adaptive one, exploration directions and tempos become highly correlated. 
As a result, the population is prone to being drawn collectively into one dominant region of attraction, which in multimodal landscapes often corresponds to a local optimum. 
Once trapped, premature convergence ensues, as the population loses its exploratory vigor and becomes unable to escape toward the global optimum.

To address these limitations, the design of iStratDE is predicated on deploying strategies at an alternative granularity: the individual level. 
The core tenet of our methodology is to endow each solution with a unique and immutable behavioral profile at initialization. 
Consequently, rather than being governed by a monolithic strategic directive, the evolutionary search in iStratDE emerges from the collective interplay of these heterogeneous, individual-specific behaviors. 
This approach is motivated by a central hypothesis: a population of individuals, each characterized by a simple, fixed, and distinct behavior, can inherently maintain sufficient persistent diversity to ensure robust convergence toward the global optimum. 
A primary objective of this design is therefore to intrinsically mitigate the risk of premature convergence, thereby obviating the need for complex adaptive control systems or explicit inter-individual coordination.

The design is also guided by two key principles. 
First, in line with DE’s original ethos of simplicity~\cite{DE1997}, iStratDE deliberately pursues algorithmic simplicity as the means to achieve diversity, rather than resorting to additional adaptive layers or control mechanisms. 
Second, a design based on independent individuals with fixed strategies is inherently conducive to parallel processing~\cite{tan2015survey,cantu2000efficient,wang2013parallel}, which offers clear advantages for scalability when handling large populations or computationally intensive problems. 
These considerations form the foundation of our proposal and shape both the conceptual underpinnings and the practical realization of iStratDE.

\subsection{Framework}\label{section_Framework}

\begin{figure}[!h]
    \centering
    \includegraphics[width=\columnwidth]{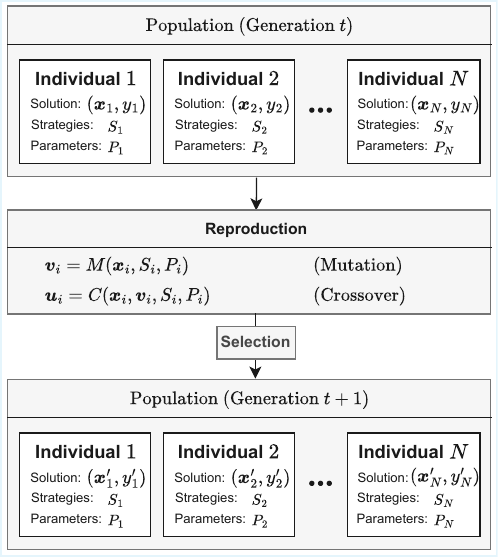}
    \caption{Framework of iStratDE. Each individual is initialized with a unique combination of mutation strategy \( S_i \) and control parameters \( P_i \). During reproduction, mutation and crossover are applied according to these assigned configurations, producing a trial vector \( \boldsymbol{u}_i \) that competes with its parent solution \( \boldsymbol{x}_i \); the fitter candidate survives into the next generation. Although the solution vector \( \boldsymbol{x}_i \) and its fitness \( y_i \) evolve over time, the assigned strategy \( S_i \) and parameters \( P_i \) remain fixed, thereby sustaining persistent diversity across the population.}
    \label{iStratDE_Structure}
\end{figure}

The fundamental framework of the iStratDE algorithm, illustrated in Fig.~\ref{iStratDE_Structure}, centers on the deployment of strategies at the \emph{individual level}, where each member of the population operates under a unique and persistent strategy. 
The process begins with the creation of an initial population, in which solution vectors are randomly generated within the search space. 
As shown in the figure, each individual $i$ is associated with a solution vector $\boldsymbol{x}_i$, whose fitness $y_i$ is obtained by evaluating the objective function. 
In addition, iStratDE assigns to each individual a dedicated and permanent strategy set $S_i$ and a parameter set $P_i$ during initialization. 
This one-time assignment of distinct and fixed strategic attributes establishes the structural diversity described in the previous section, ensuring that every individual embarks on the search with a unique behavioral profile.The cornerstone of this framework is the persistence of strategic assignments: an individual's strategy set $S_i$ and parameter set $P_i$ remain fixed throughout the evolutionary process. 
As depicted in the central and bottom panels of Fig.~\ref{iStratDE_Structure}, the generational workflow executes this principle through individual-level reproduction and selection.
In each generation, every individual applies its assigned mutation and crossover operations to produce a trial vector $\boldsymbol{u}_i$.
A one-to-one greedy Selection follows, in which the trial vector competes directly with its parent $\boldsymbol{x}_i$. the fitter (or equally fit) candidate survives into the next generation. 
Consequently, while the solution vector $\boldsymbol{x}_i$ may be replaced, its underlying strategic attributes $S_i$ and $P_i$ persist.

This static assignment of strategies deliberately severs the link between runtime performance feedback and strategy adjustment that characterizes many adaptive DE variants. 
In iStratDE, diversity is therefore not a dynamically managed state but a structural property embedded at initialization. 
This design results in a decentralized population of fixed-strategy individuals operating in parallel, in which simplicity and inherent concurrency underpin both scalability and robustness. 
The specific contents of the individual strategy sets $S_i$ are detailed in the following section.\subsection{Strategy Pool}

A core principle of iStratDE is to establish individual-level strategy diversity by assigning each individual a unique and fixed search strategy at population initialization, which it retains throughout the entire optimization process. This design produces a heterogeneous population from the outset, where each individual acts as a specialist that consistently follows its own methodology. This diversity is realized through a strategy pool constructed from several fundamental components.

The foundation of each strategy is a generalized mutation operator constructed from two base vectors: a left base vector ($\boldsymbol{x}_{bl}$) and a right base vector ($\boldsymbol{x}_{br}$). These vectors are selected from four canonical choices (\texttt{rand}, \texttt{best}, \texttt{pbest} with $p=0.05$, and \texttt{current}) to determine the search direction, thereby balancing global exploration with local exploitation. The mutation vector $\boldsymbol{v}$ is then defined as\begin{equation}
\boldsymbol{v} = \boldsymbol{x}_{bl} + F \cdot (\boldsymbol{x}_{br} - \boldsymbol{x}_{bl}) + F \cdot (\Delta_1 + \Delta_2 + \dots + \Delta_{dn}),
\end{equation}where each $\Delta_i$ denotes the difference between two distinct individuals not previously selected. The overall magnitude of the perturbation is further controlled by the number of differential vectors, $dn \in \{1,2,3,4\}$.

To complete the strategy definition, a crossover scheme (\texttt{bin}, \texttt{exp}, or \texttt{arith}) is selected to determine how the mutant vector is combined with its parent. In addition, each individual is assigned its own scaling factor $F$ and crossover rate $CR$, independently sampled from a uniform distribution $U(0,1)$, which introduces another layer of variability.

Accordingly, each individual's fixed strategy can be represented in the format \texttt{DE/bl-to-br/dn/cs}. The combinations listed in Table~\ref{tab:strategy-pool} yield a pool of 192 distinct configurations. When coupled with the per-individual random sampling of $F$ and $CR$, this combinatorial richness produces a diverse initialization space. Moreover, the large population size employed in iStratDE ensures that this random sampling statistically provides comprehensive and balanced coverage of the entire strategy space, which secures robust structural diversity across the population.

\begin{table}[htbp]
\centering
\caption{Constituents of the strategy pool in iStratDE. The Cartesian product of these components yields 192 distinct configurations, from which each individual is randomly assigned a fixed strategy at initialization.}
\label{tab:strategy-pool}
\renewcommand{\arraystretch}{1.25} 
\resizebox{\columnwidth}{!}{%
\begin{tabular}{@{}>{\centering\arraybackslash}m{0.5\linewidth} l@{}}
\toprule
\textbf{Component} & \textbf{Candidate Values} \\
\midrule
Base Vector Left ($bl$) & \texttt{rand}, \texttt{best}, \texttt{pbest}, \texttt{current} \\
Base Vector Right ($br$) & \texttt{rand}, \texttt{best}, \texttt{pbest}, \texttt{current} \\
Differential Vector Number ($dn$) & \texttt{1}, \texttt{2}, \texttt{3}, \texttt{4} \\
Crossover Scheme ($cs$) & \texttt{bin}, \texttt{exp}, \texttt{arith} \\
\bottomrule
\end{tabular}}
\end{table}

To further justify the composition of our strategy pool, we conducted extensive sensitivity analyses and component exclusion experiments, which are presented in the Supplementary Material. We examined the impact of varying the strategy pool size (e.g., 1, 5, 20, and 192) and observed a positive correlation where larger pools yield superior performance. This finding confirms that iStratDE's robustness stems fundamentally from the \emph{sufficient richness} of the strategy source rather than from the fine-tuning of a specific combination. Additionally, tests involving the removal of specific mutation or crossover types (e.g., retaining only exploitation-heavy strategies) highlight the necessity of the full pool to maintain a critical balance between exploration and exploitation. Comprehensive data and discussions are provided in the Supplementary Document.

\subsection{Convergence Analysis}

We analyze the convergence of iStratDE under standard assumptions following the Markov chain framework widely used in evolutionary algorithm theory~\cite{rudolph1994convergence, hu2014finite}. Let $P^{(t)}=\{\boldsymbol{x}_i^{(t)}\}_{i=1}^N$ denote the population at iteration $t$, which constitutes the state of a time-homogeneous Markov chain on the state space $\mathcal{S}=\mathcal{X}^N$, where $\mathcal{X}\subset\mathbb{R}^D$ is the decision space and $N$ is the population size. We assume throughout that $\mathcal{X}$ is compact (closed and bounded)~\cite{nocedal1999numerical} and that the objective function $f:\mathcal{X}\to\mathbb{R}$ is continuous, which ensures the existence of at least one global optimum $\boldsymbol{x}^*\in\mathcal{X}$.

Define the best individual in $P^{(t)}$ as\begin{equation}
    \boldsymbol{x}_{\text{best}}^{(t)}=\arg\min_{1\le i\le N} f(\boldsymbol{x}_i^{(t)}).
\end{equation}iStratDE employs one-to-one greedy selection, which ensures that the best-so-far fitness never deteriorates:\begin{equation}
    f(\boldsymbol{x}_{\text{best}}^{(t+1)}) \le f(\boldsymbol{x}_{\text{best}}^{(t)}).
\end{equation}Thus, $\{f(\boldsymbol{x}_{\text{best}}^{(t)})\}_{t=0}^\infty$ is monotonically non-increasing. Because $f$ is bounded below on compact $\mathcal{X}$, this sequence is guaranteed to converge.

We assume that the variation operators of iStratDE satisfy a \emph{uniform global reachability} property: for any $\epsilon>0$, there exists a constant $p_\epsilon>0$ such that, from any population state $P\in\mathcal{S}$, the probability of generating at least one offspring in the $\epsilon$-ball $B_\epsilon(\boldsymbol{x}^*)$ around a global optimum is at least $p_\epsilon$. This condition guarantees that, regardless of the current state, the algorithm always has a positive, state-independent probability of generating an individual arbitrarily close to $\boldsymbol{x}^*$.

For any $\eta>0$, define the level set\begin{equation}
    \mathcal{A}_\eta=\{P\in\mathcal{S}\mid f(\boldsymbol{x}_{\text{best}}(P)) \le f(\boldsymbol{x}^*)+\eta\}.
\end{equation}By the monotonicity of $f(\boldsymbol{x}_{\text{best}}^{(t)})$, once the process enters $\mathcal{A}_\eta$, it cannot leave. Consequently, $\mathcal{A}_\eta$ is an absorbing set. By the continuity of $f$, there exists an $\epsilon(\eta)>0$ such that $\boldsymbol{x} \in B_{\epsilon(\eta)}(\boldsymbol{x}^*)$ implies $f(\boldsymbol{x})\le f(\boldsymbol{x}^*)+\eta$. This result, combined with the uniform reachability assumption, implies that $\mathcal{A}_\eta$ is reached with probability one in finite time. As this holds for any $\eta>0$, we conclude that\begin{equation}
    \Pr\!\Bigl(\lim_{t\to\infty} f(\boldsymbol{x}_{\text{best}}^{(t)}) = f(\boldsymbol{x}^*)\Bigr)=1.
\end{equation}Therefore, under the assumptions of compactness, continuity, and uniform global reachability, iStratDE converges almost surely to the global optimum as the number of iterations tends to infinity. Although computational resources are finite in practice, this analysis provides a theoretical foundation for the robust convergence behavior observed empirically. The structural diversity inherent to iStratDE helps balance exploration and exploitation, thereby mitigating premature convergence. These properties are further demonstrated in Section~\ref{section_Experimental_study}.

\subsection{Discussion}
As described in Section~\ref{section_Framework}, iStratDE introduces population-wide structural heterogeneity by assigning strategies independently at the individual level. This decentralized assignment fosters a diverse set of search behaviors that collectively enhance the algorithm’s robustness and resistance to premature convergence. Unlike traditional DE variants that typically impose a uniform or adaptively shared strategy, iStratDE allows individuals to follow distinct and persistent trajectories, thereby broadening exploration across multimodal landscapes~\cite{sallam2017landscape}. The procedural framework of iStratDE is summarized in Algorithm~\ref{Alg_iStratDE}.

\begin{algorithm}
\small
\caption{iStratDE}
\label{Alg_iStratDE}
\begin{algorithmic}[1]
  \Require $D$, $N$, $G_{\max}$, $\boldsymbol{lb}$, $\boldsymbol{ub}$
  \State Initialize $\boldsymbol{X} = \{\boldsymbol{x}_1, \dots, \boldsymbol{x}_{N}\}$ uniformly within $[\boldsymbol{lb}, \boldsymbol{ub}]$
  \State Randomly assign a strategy set $S_i$ and parameter set $P_i$ to each individual $\boldsymbol{x}_i$
  \State Evaluate fitness $\boldsymbol{y} = \{ f(\boldsymbol{x}_1), \dots, f(\boldsymbol{x}_{N}) \}$
  \State $g \gets 0$
  \While{$g < G_{\max}$}
    \For{$i = 1$ to $N$}
      \State $\boldsymbol{v}_i \gets M(\boldsymbol{x}_i, S_i, P_i)$
      \State $\boldsymbol{u}_i \gets C(\boldsymbol{x}_i, \boldsymbol{v}_i, S_i, P_i)$
      \If{$f(\boldsymbol{u}_i) \leq f(\boldsymbol{x}_i)$}
        \State $\boldsymbol{x}_i \gets \boldsymbol{u}_i$
      \EndIf
    \EndFor
    \State $g \gets g + 1$
  \EndWhile
  \State \Return $\boldsymbol{x}_{\text{best}}$, $f(\boldsymbol{x}_{\text{best}})$
\end{algorithmic}
\end{algorithm}

Through this mechanism, iStratDE maintains individuals that simultaneously pursue exploration and exploitation. For instance, configurations such as \texttt{DE/rand/1/bin} introduce large perturbations and enhance global exploration, whereas strategies such as \texttt{DE/current-to-pbest/1/bin} emphasize local refinement near promising areas. The coexistence of such complementary behaviors prevents stagnation, ensuring that new areas of the search space are continuously discovered while existing optima are refined. This perspective resonates with prior studies highlighting the importance of population diversity for evolutionary search, such as Ursem’s diversity-guided evolutionary algorithms~\cite{ursem2002diversity}.

From a computational perspective, the time complexity of iStratDE per generation, as detailed in Algorithm~\ref{Alg_iStratDE}, is $O(N \cdot (D + T(f)))$, where $N$ is the population size, $D$ is the problem dimension, and $T(f)$ is the cost of one function evaluation. This complexity is identical to that of canonical DE. The algorithm achieves this equivalence because its only additional operation is a one-time, negligible $O(N)$ cost incurred at initialization. As shown in Algorithm~\ref{Alg_iStratDE} (line 2), this cost is for randomly assigning the fixed strategies and parameters to each individual. Crucially, the main evolutionary loop does not introduce any complex adaptive mechanisms, such as historical archive maintenance or feedback-based parameter adjustments. Moreover, considering the GPU architecture, the independence of individuals allows for fully parallel computation. Consequently, even as the population size increases significantly, the actual wall-clock time per generation does not rise proportionally. Within the effective parallel capacity of the GPU, the expansion of the population size has a limited impact on the runtime, which makes the deployment of massive populations computationally feasible.

An additional advantage of iStratDE lies in its \emph{scalable simplicity}. Whereas advanced methods like jSO~\cite{JSO2017} rely on success-history adaptation and multiple control layers, iStratDE achieves robustness by implicitly sustaining population heterogeneity through randomized, fixed strategies. This design dispenses with extra control components and synchronization, which yields a streamlined yet versatile framework that adapts well across different problem domains.

Conceptually, the emphasis on structural diversity aligns with the philosophy of Quality Diversity (QD) algorithms~\cite{pugh2016quality}, most notably the MAP-Elites framework~\cite{mouret2015illuminating}. QD methods aim to produce collections of high-performing yet behaviorally distinct solutions. This principle has been shown to enhance adaptability, for instance, in robots recovering from damage~\cite{cully2015robots}. While iStratDE does not explicitly construct a behavior–performance map or employ feature descriptors, its randomized and fixed assignment of strategies naturally encourages the preservation of diverse behavioral niches, enabling individuals to specialize in distinct regions of the search space. In this sense, iStratDE attains a similar effect, i.e., maintaining behavioral variety, without the bookkeeping typically required in QD algorithms.

A similar parallel can be drawn with Novelty Search (NS)~\cite{lehman2011abandoning}, which rewards behavioral novelty to drive exploration. Although iStratDE does not measure novelty explicitly, its decentralized design guarantees that a subset of the population continuously explores along diverse trajectories. This persistent heterogeneity reduces the risk of premature convergence and improves the algorithm’s capacity to traverse complex, high-dimensional landscapes. Thus, while iStratDE is not an explicit QD or NS method, it converges philosophically with both by treating structural diversity as a first-class mechanism for resilience and adaptability.\section{Experimental Study}\label{section_Experimental_study}This section reports a comprehensive experimental evaluation of iStratDE to assess both its effectiveness and practical applicability. The study consists of three parts. First, iStratDE is benchmarked against representative DE variants on the CEC2022 test suite~\cite{CEC2022SO} to evaluate relative performance. Second, iStratDE is compared with the top-performing algorithms from the CEC2022 Competition on Single-objective Bound-constrained Numerical Optimization. Finally, iStratDE is applied to robotic control tasks to demonstrate its utility in real-world scenarios.

We conducted all experiments on a workstation equipped with an Intel Core i9-10900X CPU and an NVIDIA RTX 3090 GPU. To ensure a fair comparison and to exploit GPU acceleration, we implemented all algorithms and test functions within the EvoX framework~\cite{evox}.

\subsection{Experimental Setup}

We evaluate the efficacy of iStratDE on the CEC2022 benchmark suite~\cite{CEC2022SO}, a comprehensive set of twelve test functions designed to challenge evolutionary algorithms across diverse levels of difficulty. The suite comprises basic functions ($F_1$–$F_5$), hybrid functions ($F_6$–$F_8$), and composition functions ($F_9$–$F_{12}$), which collectively cover a wide range of modalities, separability, and ruggedness.

For comparison, we selected several representative Differential Evolution variants. These include the classical \texttt{DE/rand/1/bin} and a set of widely recognized adaptive or ensemble-based algorithms: JADE~\cite{JADE2009}, SHADE~\cite{SHADE2013}, CoDE~\cite{CoDE2011}, SaDE~\cite{SaDE2008}, LSHADE-RSP~\cite{LSHADE-RSP2018}, and EDEV~\cite{EDEV2018}. These methods are competitive across many benchmark studies and thus serve as strong baselines. To ensure a consistent implementation, all algorithms were implemented within the EvoX framework~\cite{evox}.

Unless otherwise specified, the baseline algorithms used a population size of 100, which follows common practice, as larger populations often lead to stagnation or degraded performance in conventional DE variants. We configured their control parameters, such as the scaling factor $F$ and crossover rate $CR$, according to the recommendations in the original studies.

In contrast, iStratDE deliberately departs from this convention. As discussed in Section~\ref{section_The proposed iStratDE}, its architecture eliminates manual parameter tuning and is explicitly designed to exploit large-scale populations in parallel computing environments. To test this central hypothesis, we set the population size for iStratDE to 100,000. Although such a scale is rarely feasible for traditional DE algorithms due to computational constraints, iStratDE leverages this massive population to mitigate the risk of premature convergence and demonstrate its performance advantages. To ensure a rigorous and fair comparison, we additionally evaluated the baselines under varying population sizes (see Section~\ref{sec:population_scaling}), thereby enabling a more meaningful assessment of their intrinsic scalability.

For the time controlled experiments, we used a fixed time budget of 60 seconds. In a GPU accelerated environment, this duration enables tens of millions of function evaluations, a scale that substantially exceeds standard CPU based experiments. Empirical observations indicate that baseline algorithms typically stagnate within the first 5 to 10 seconds. Consequently, this 60 second budget is sufficient to ensure that all methods can fully exhaust their search potential.

For numerical stability, results with an error smaller than $10^{-8}$ were rounded to zero. Each experiment was independently repeated 31 times to ensure statistical reliability.

\subsection{Performance Comparison under Equal Time}\label{sec:expereiment_time}For this experiment, we established a fixed runtime of \SI{60}{\second} as the termination criterion and a uniform population size of 100 for all algorithms. This configuration ensures a fair comparison of convergence behavior under identical time budgets. The evaluation was conducted on both the 10D and 20D versions of the CEC2022 benchmark suite, which contains highly multimodal and composition functions.\begin{figure}[htbp]
\centering
\includegraphics[width=0.95\columnwidth]{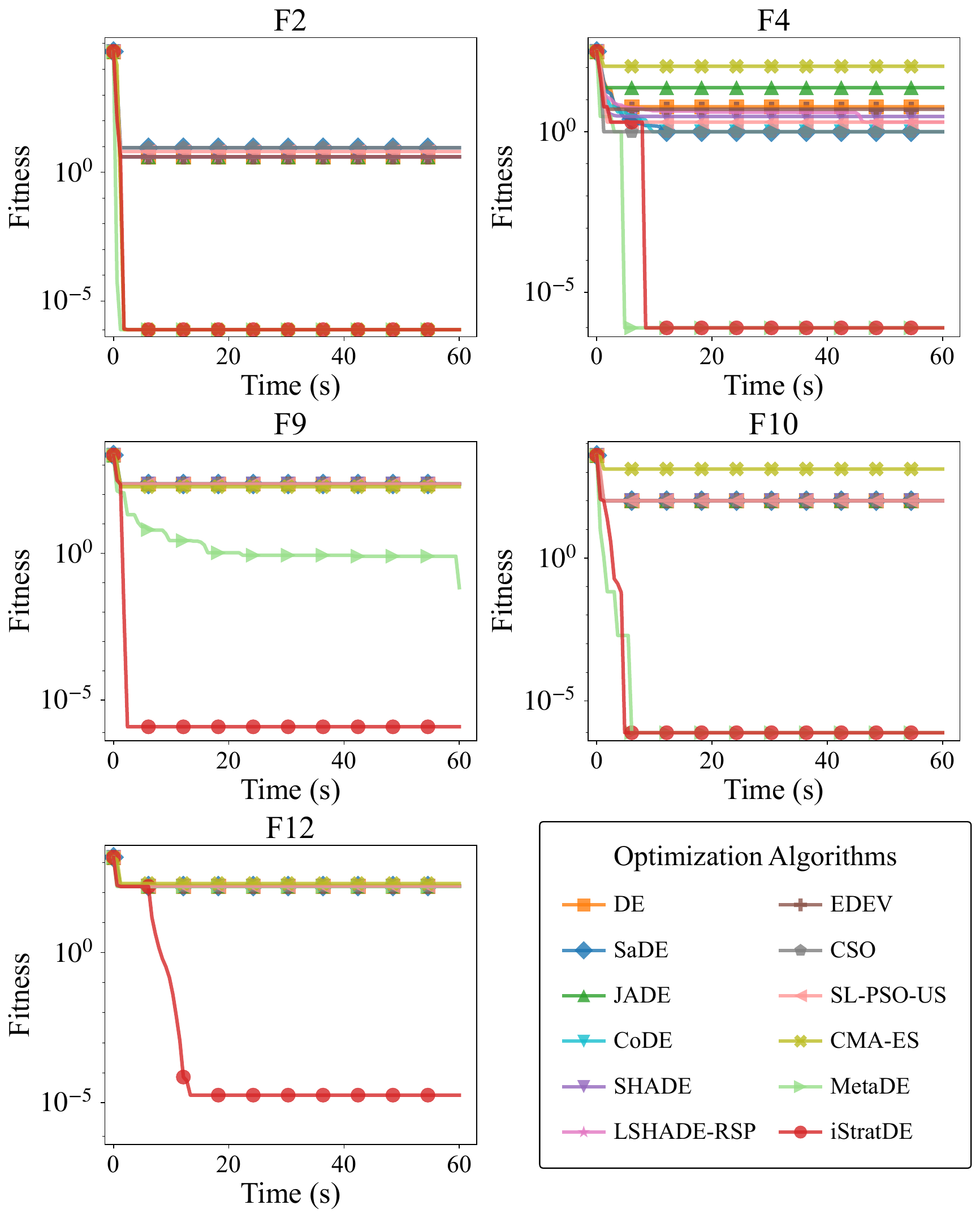}
\caption{Convergence curves on 10D problems in the CEC2022 benchmark suite.}
\label{Figure_convergence_10D}
\end{figure}Fig.~\ref{Figure_convergence_10D} presents the convergence trajectories for five representative functions in the 10D setting. Comprehensive results for all functions, including the 20D cases, are provided in the Supplementary Document. Across these problems, iStratDE exhibits consistently fast and stable convergence, often achieving the lowest fitness values. On more challenging instances, its advantage is particularly evident, as iStratDE avoids stagnation in local optima and steadily progresses toward globally optimal regions.

Notably, iStratDE does not outperform competitors on all instances, such as the 10D $F_{11}$ case in the Supplementary Material. This reflects the \emph{No Free Lunch} theorem: while greedy, adaptive algorithms excel on simple landscapes through aggressive exploitation, iStratDE avoids such rapid convergence. This represents a deliberate trade-off, sacrificing maximal precision on simple problems to prioritize robustness on the complex, multimodal landscapes that constitute the primary focus of this study.

This performance stems from iStratDE’s population-wide structural diversity. By assigning strategies and parameters randomly at the individual level, the algorithm maintains a heterogeneous search behavior, thereby improving exploration and reducing the likelihood of premature convergence. This mechanism enhances global coverage of the search space and enables better adaptation to different landscape characteristics.

\begin{figure}[!h]
\centering
\includegraphics[width=0.95\columnwidth]{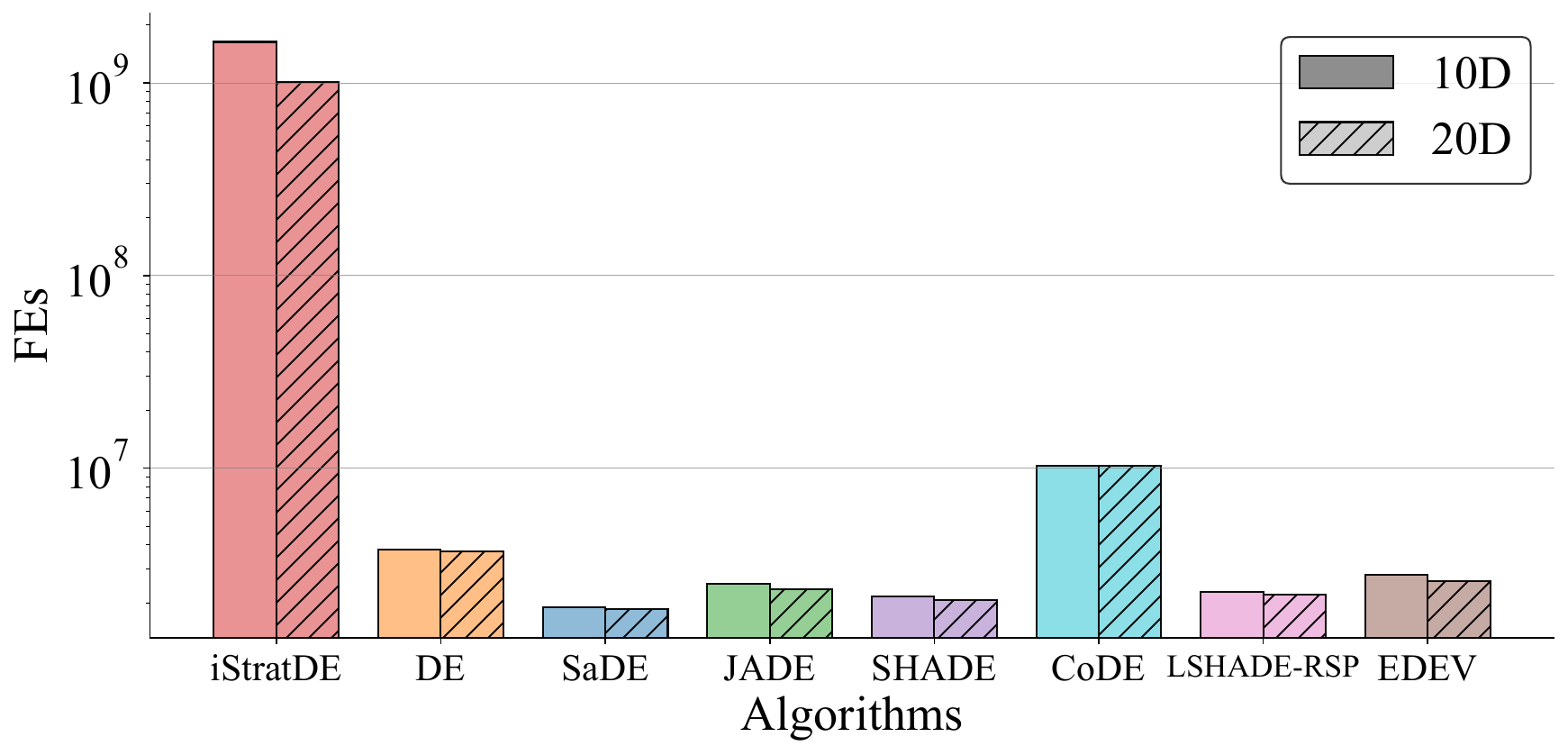}
\caption{Average number of function evaluations (FEs) completed within \SI{60}{\second} on 10D and 20D problems in the CEC2022 benchmark suite.}
\label{fig:maxFEs}
\end{figure}

In addition to effectiveness, iStratDE also demonstrates strong execution efficiency. Fig.~\ref{fig:maxFEs} reports the number of function evaluations (FEs) completed by conventional DE variants within the 60-second budget. iStratDE performs on the order of \(10^9\) evaluations—two to three orders of magnitude higher than the \(10^6\)–\(10^7\) evaluations achieved by competing algorithms. This substantial gap reflects the algorithm’s intrinsic concurrency: its fully independent update mechanism maps naturally onto GPU architectures, enabling massive parallelism without additional synchronization overhead. 
By contrast, although other algorithms are also implemented with GPU support, their complex adaptive mechanisms introduce control dependencies and synchronization barriers, thereby limiting the extent to which GPU acceleration can be fully exploited.\subsection{Performance Comparison under Equal FEs}

To further assess optimization performance independent of runtime, we compared iStratDE and baseline differential evolution algorithms under an equal number of function evaluations (FEs). The FE limit for each algorithm was set to match the number of evaluations performed by iStratDE within 60 seconds in the previous experiment (Section~\ref{sec:expereiment_time}). Due to space constraints, Table~\ref{tab:sameFEs} reports representative results on a subset of challenging functions from the CEC2022 suite. Full results across all problems are provided in the Supplementary Document. The table presents the mean and standard deviation of final errors, with the best results highlighted. Statistical significance was assessed using the Wilcoxon rank-sum test at a 0.05 significance level.
\begin{table*}[htbp]
\centering
\caption{Comparisons between iStratDE and other DE variants under equal FEs. 
The mean and standard deviation (in parentheses) of the results over multiple runs are displayed. 
Best mean values are highlighted.}
\resizebox{\linewidth}{!}{
\renewcommand{\arraystretch}{1.2}
\renewcommand{\tabcolsep}{2pt}
\begin{tabular}{ccccccccccc}
\toprule
\multicolumn{2}{c}{Func} & iStratDE  & DE    & SaDE  & JADE  & CoDE  & SHADE & LSHADE-RSP & EDEV&MetaDE\\
\midrule
\multirow{6}[2]{*}{10D} 
  & $F_{2}$  & \textbf{0.00E+00 (0.00E+00)}  & 5.95E+00 (2.37E+00)$-$ & 5.95E+00 (4.48E+00)$-$ & 6.38E+00 (3.21E+00)$-$ & 5.36E+00 (2.15E+00)$-$ & 4.75E+00 (3.68E+00)$-$ & 3.87E+00 (3.25E+00)$-$ & 5.04E+00 (2.94E+00)$-$ &\textbf{0.00E+00 (0.00E+00)}$\approx$\\
  & $F_{4}$  & \textbf{0.00E+00 (0.00E+00)} & 7.90E+00 (3.35E+00)$-$ & 7.30E-01 (5.71E-01)$-$ & 2.49E+01 (1.21E+01)$-$ & 6.63E-01 (5.93E-01)$-$ & 2.92E+00 (5.71E-01)$-$ & 1.78E+00 (6.60E-01)$-$ & 4.61E+00 (1.76E+00)$-$ &\textbf{0.00E+00 (0.00E+00)$\approx$}\\
  & $F_{6}$  & 1.33E-02 (8.36E-03) & 1.08E-01 (1.92E-01)$-$ & 3.42E+00 (3.10E+00)$-$ & 5.83E-01 (4.15E-01)$-$ & \textbf{0.00E+00 (0.00E+00)}$+$ & 1.46E+00 (4.16E+00)$-$ & 4.20E-06 (1.54E-05)$+$ & 4.30E-01 (5.44E-01)$-$ & 5.50E-04(3.96E-04)$+$\\
  & $F_{9}$  & \textbf{1.75E-06 (0.00E+00)}  & 2.29E+02 (3.92E-06)$-$ & 2.29E+02 (8.77E-06)$-$ & 2.29E+02 (4.33E-06)$-$ & 2.29E+02 (6.19E-06)$-$ & 2.29E+02 (0.00E+00)$-$ & 2.29E+02 (8.65E-06)$-$ & 2.29E+02 (4.84E-06)$-$ & 3.36E+00(1.77E+01)$-$\\
  & $F_{10}$ & \textbf{0.00E+00 (0.00E+00)}  & 1.00E+02 (4.65E-02)$-$ & 1.20E+02 (2.65E+01)$-$ & 1.08E+02 (2.93E+01)$-$ & 1.00E+02 (0.00E+00)$-$ & 1.03E+02 (2.58E+01)$-$ & 1.00E+02 (0.00E+00)$-$ & 1.01E+02 (3.04E+01)$-$ &\textbf{0.00E+00 (0.00E+00)}$\approx$\\
  & $F_{12}$ & \textbf{6.35E+01 (7.77E+01)} & 1.62E+02 (1.52E+00)$-$ & 1.63E+02 (1.70E+00)$-$ & 1.62E+02 (2.31E+00)$-$ & 1.59E+02 (3.68E-01)$-$ & 1.63E+02 (8.33E-01)$-$ & 1.64E+02 (1.56E+00)$-$ & 1.63E+02 (9.38E-01)$-$ & 1.39E+02(4.63E+01)$-$\\
\midrule
\multirow{6}[2]{*}{20D} 
  & $F_{2}$  & \textbf{0.00E+00 (0.00E+00)}  & 4.70E+01 (1.90E+00)$-$ & 4.91E+01 (1.96E+00)$-$ & 4.68E+01 (1.27E+00)$-$ & 4.91E+01 (4.20E-06)$-$ & 4.55E+01 (1.96E+00)$-$ & 4.50E+01 (2.00E+00)$-$ & 4.47E+01 (1.55E+01)$-$ &1.26E-02(3.74E-02)$-$\\
  & $F_{4}$  & 1.04E+01 (2.41E+00) & 2.11E+01 (9.69E+00)$-$ & 8.56E+00 (2.46E+00)$+$ & 7.31E+01 (3.48E+01)$-$ & 9.95E+00 (1.82E+00)$\approx$ & 1.27E+01 (3.17E+00)$-$ & 8.69E+00 (2.34E+00)$+$ & 1.86E+01 (6.63E+00)$-$ &\textbf{2.02E+00(8.56E-01)$+$}\\
  & $F_{6}$  & 3.53E-01 (3.05E-01) & 7.16E-01 (5.31E-01)$-$ & 4.05E+01 (2.41E+01)$-$ & 6.33E+01 (2.80E+01)$-$ & 1.05E+01 (1.04E+01)$-$ & 5.06E+01 (3.06E+01)$-$ & 5.52E+00 (3.44E+00)$-$ & 3.62E+01 (2.83E+01)$-$ &\textbf{1.16E-01(2.79E-02)$+$}\\
  & $F_{9}$  & \textbf{1.00E+02 (0.00E+00)}  & 1.81E+02 (4.27E-06)$-$ & 1.81E+02 (7.79E-06)$-$ & 1.81E+02 (4.47E-06)$-$ & 1.81E+02 (1.10E-05)$-$ & 1.81E+02 (8.82E-06)$-$ & 1.81E+02 (9.73E-06)$-$ & 1.81E+02 (8.14E-07)$-$ &1.07E+02(1.98E+01)$-$\\
  & $F_{10}$ & \textbf{0.00E+00 (0.00E+00)} & 1.25E+02 (8.32E+01)$-$ & 1.00E+02 (1.68E-02)$-$ & 1.01E+02 (3.81E+01)$-$ & 1.00E+02 (2.09E-02)$-$ & 1.07E+02 (2.97E+01)$-$ & 1.06E+02 (2.89E+01)$-$ & 1.17E+02 (4.28E+01)$-$ &\textbf{0.00E+00 (0.00E+00)$\approx$}\\
  & $F_{12}$ & 2.31E+02 (1.40E+00) & 2.37E+02 (3.64E+00)$-$ & 2.42E+02 (6.10E+00)$-$ & 2.35E+02 (4.44E+00)$-$ & 2.34E+02 (3.38E+00)$-$ & 2.37E+02 (3.26E+00)$-$ & 2.40E+02 (4.38E+00)$-$ & 2.44E+02 (4.51E+00)$-$ & \textbf{2.29E+02(6.08E-01)$+$}\\
\midrule
\multicolumn{2}{c}{$+$ / $\approx$ / $-$} & --  & 0/0/12 & 1/0/11 & 0/0/12 & 1/1/10 & 0/0/12 & 2/0/10 & 0/0/12 &4/4/4\\
\bottomrule
\end{tabular}}
\label{tab:sameFEs}
\footnotesize
\textsuperscript{*} Wilcoxon rank-sum tests ($\alpha=0.05$) were performed between iStratDE and each comparator. The bottom row reports the number of functions where the comparator performed significantly better ($+$), similar ($\approx$), or worse ($-$) than iStratDE.
\end{table*}Overall, iStratDE consistently delivers superior results under equal FEs. On difficult cases such as $F_2$, $F_9$, and $F_{10}$, it achieves near-zero average error across multiple runs, which clearly demonstrates its ability to avoid local optima and converge reliably to global solutions. Even where other algorithms show progress, their improvements are typically modest compared to iStratDE.

These findings emphasize that the benefit of iStratDE is not simply the volume of evaluations it can process, but also the effectiveness with which those evaluations are translated into quality improvements. The algorithm’s structural heterogeneity allows it to exploit large evaluation budgets more productively, which leads to rapid error reduction even on rugged landscapes where competing methods stagnate. Importantly, the advantage is observed across different categories of functions (i.e., basic, hybrid, and composition), which highlights the broad applicability of iStratDE as a general-purpose optimizer, particularly in settings where ample computational resources are available.

\subsection{Performance Comparison under High Dimension}
To further evaluate the scalability and robustness of iStratDE in solving large-scale optimization problems, we extended our experiments to high-dimensional landscapes. Specifically, we tested the algorithms on rotated and shifted 200-dimensional ($200D$) versions of representative multimodal functions: Schwefel, Rastrigin, and Ackley. In this experimental setup, the population size for iStratDE was set to 10,000, while the population size for all comparative algorithms was maintained at 100. To ensure the difficulty of the optimization task and mimic realistic black-box scenarios, all functions were subjected to shift and rotation operations.

\begin{figure}[htbp]
\centering
\includegraphics[width=0.95\columnwidth]{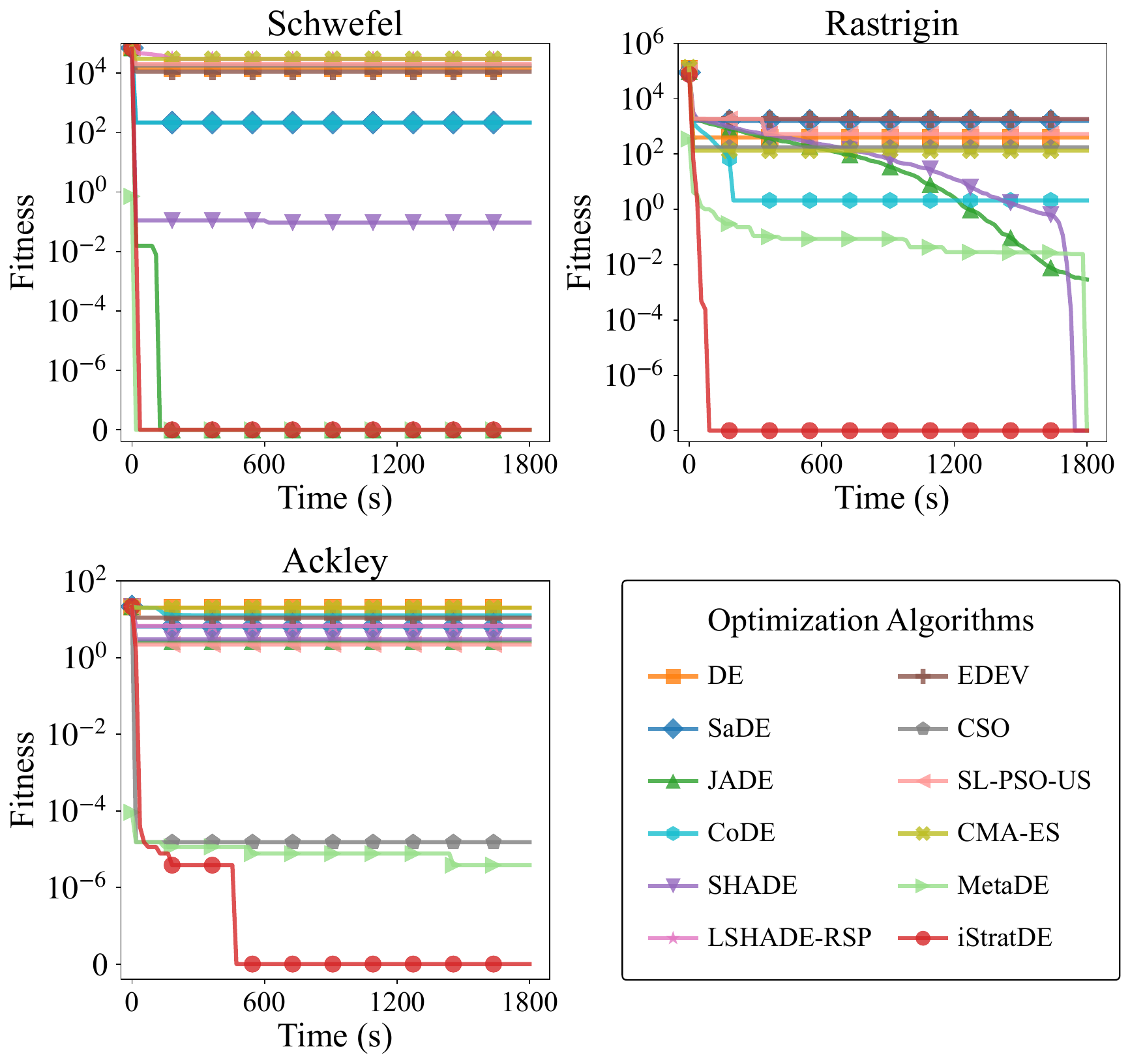}
\caption{Convergence curves on 200D rotated and shifted Schwefel, Rastrigin, and Ackley functions.}
\label{Figure_convergence_200D}
\end{figure}

Fig.~\ref{Figure_convergence_200D} illustrates the convergence trajectories of iStratDE and the comparative algorithms over time. The results indicate that the \emph{curse of dimensionality} severely impacts most traditional adaptive DE variants (e.g., SaDE, JADE, and SHADE) and standard baselines (e.g., CMA-ES), which leads to early stagnation or slow convergence rates. In contrast, iStratDE exhibits substantial scalability. Its individual-level strategy diversity and large population size, enabled by GPU parallelism, allow iStratDE to maintain a continuous downward trend in fitness values, thereby effectively escaping local optima that trap other algorithms.

iStratDE is also highly competitive when compared with other specialized high-dimensional or GPU-accelerated algorithms. As observed in Fig.~\ref{Figure_convergence_200D}, iStratDE significantly outperforms established large-scale optimizers such as CSO and SL-PSO-US. Moreover, it exhibits superior robustness compared to MetaDE, a state-of-the-art learning-based algorithm published in 2025 \cite{metade} that is similarly designed based on GPU architecture. While both algorithms achieve rapid convergence on Schwefel and Rastrigin, iStratDE distinguishes itself on the Ackley function, where it successfully locates the global optimum, whereas MetaDE shows slightly weaker performance than iStratDE. This demonstrates that even without complex learning mechanisms, the structural diversity intrinsic to iStratDE provides a highly competitive and scalable solution for high-dimensional optimization.

\subsection{Population Scaling Performance}\label{sec:population_scaling}

\begin{figure}[htbp]
    \centering
    \includegraphics[width=\columnwidth]{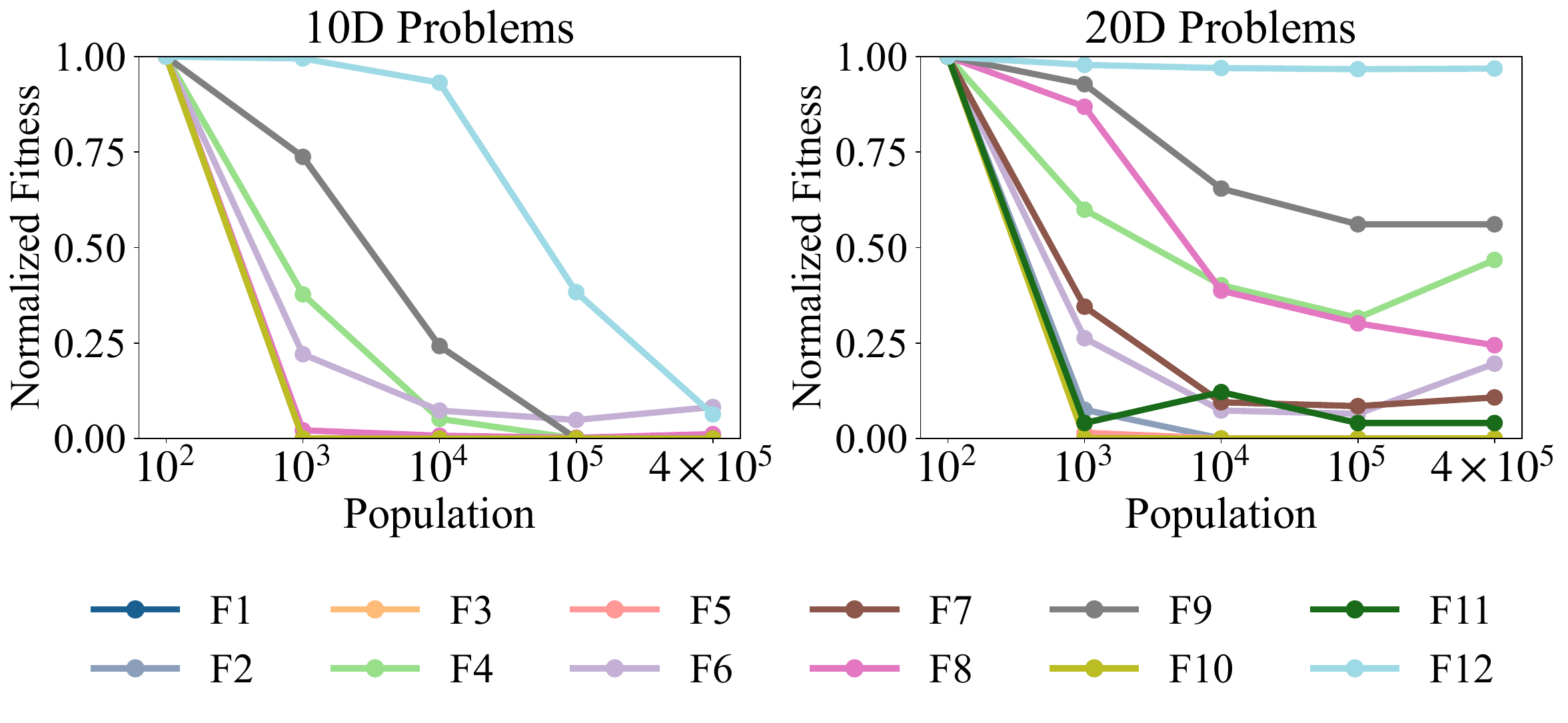}
    \caption{Normalized average fitness on 10D and 20D CEC2022 problems under different population sizes. The left panel shows results for 10D problems, and the right panel shows results for 20D problems.}
    \label{combined_pop_trend}
\end{figure}

We next examine the impact of population size on optimization performance. Consistent with the experimental setup in Section~\ref{sec:expereiment_time}, all experiments in this section use a fixed time budget of 60 seconds as the termination condition. Fig.~\ref{combined_pop_trend} presents the normalized average fitness values of iStratDE on 10D and 20D CEC2022 problems under different population sizes. The results reveal a clear and consistent trend: the performance of iStratDE improves steadily as the population grows, which confirms its suitability for large-population settings. This trend is attributable to the core design of iStratDE: assigning randomized strategies and parameters at the individual level becomes increasingly effective at scale, as larger populations sustain greater behavioral diversity and exhibit stronger resistance to premature convergence~\cite{cantu2000efficient}.

\begin{figure*}[htbp]
    \centering
    \includegraphics[width=\textwidth]{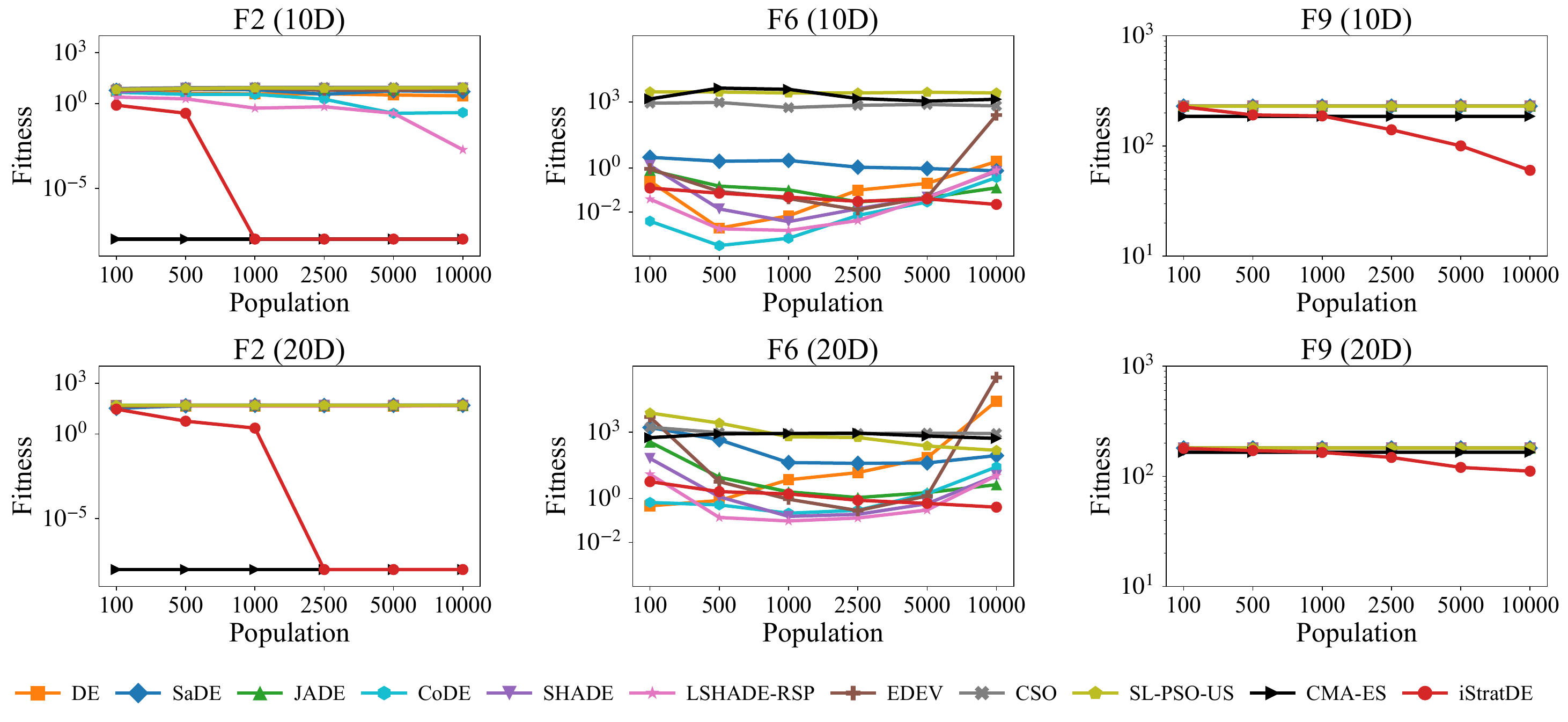}
    \caption{Performance sensitivity of baseline DE variants to population size on selected CEC2022 functions.}
    \label{baseline_pop_change}
\end{figure*}

The benefits are particularly pronounced on hybrid and composition functions such as $F_9$ and $F_{12}$, where the search space contains multiple competing basins of attraction. Such complex landscapes demand extensive exploration, and the heterogeneous population dynamics of iStratDE enable it to navigate them more effectively.To validate our design and highlight the limitations of conventional DE algorithms, we analyzed the performance of baseline variants under increasing population sizes. Fig.~\ref{baseline_pop_change} reports representative results on selected functions. For direct comparison, the performance of iStratDE is also plotted on the same axes. In contrast to iStratDE, most baselines fail to exhibit significant improvements as population size increases. On difficult problems such as $F_9$, performance remains stagnant regardless of scale, whereas on simpler functions (e.g., $F_6$ in the 10D case), only minor gains are observed. Even with populations as large as 10,000, baseline algorithms fail to achieve substantial improvements. 
This stagnation reflects an inherent structural ceiling in traditional DE designs~\cite{mahdavi2015metaheuristics,yang2007differential}, which prevents them from effectively exploiting large-scale parallelism.Taken together, these results demonstrate a fundamental contrast: whereas conventional DE variants face a performance ceiling and fail to leverage large populations, iStratDE scales robustly and benefits directly from increasing population size. Its concurrency-friendly architecture allows it to transform additional individuals into meaningful search diversity, thereby avoiding stagnation and unlocking scalability in ways that traditional DE cannot.

\subsection{Comparisons with Top Algorithms in CEC2022 Competition}\label{section_Comparison with Top Algorithms of CEC Competition}\begin{table*}[!t]
  \centering
  
  \caption{{Comparisons between iStratDE and the top 3 algorithms from CEC2022 Competition (10D). 
The mean and standard deviation (in parentheses) of the results over multiple runs are displayed in pairs. 
Results with the best mean values are highlighted. }
  }
\footnotesize
\begin{tabular}{cccccc}
\toprule
Func  & iStratDE & EA4eig & NL-SHADE-LBC & NL-SHADE-RSP \\
\midrule
$F_{1}$ & \textbf{0.00E+00 (0.00E+00)} & \boldmath{}\textbf{0.00E+00 (0.00E+00)$\approx$}\unboldmath{} & \boldmath{}\textbf{0.00E+00 (0.00E+00)$\approx$}\unboldmath{} & \boldmath{}\textbf{0.00E+00 (0.00E+00)$\approx$}\unboldmath{} \\
$F_{2}$ & \textbf{0.00E+00 (0.00E+00)} & 1.33E+00 (1.88E+00)$-$ & \boldmath{}\textbf{0.00E+00 (0.00E+00)$\approx$}\unboldmath{} & \boldmath{}\textbf{0.00E+00 (0.00E+00)$\approx$}\unboldmath{} \\
$F_{3}$ & \textbf{0.00E+00 (0.00E+00)} & \boldmath{}\textbf{0.00E+00 (0.00E+00)$\approx$}\unboldmath{} & \boldmath{}\textbf{0.00E+00 (0.00E+00)$\approx$}\unboldmath{} & \boldmath{}\textbf{0.00E+00 (0.00E+00)$\approx$}\unboldmath{} \\
$F_{4}$ & \textbf{0.00E+00 (0.00E+00)} & 1.66E+00 (9.38E-01)$-$ & \boldmath{}\textbf{0.00E+00 (0.00E+00)$\approx$}\unboldmath{} & 3.23E+00 (4.31E-01)$-$ \\
$F_{5}$ & \textbf{0.00E+00 (0.00E+00)} & \boldmath{}\textbf{0.00E+00 (0.00E+00)$\approx$}\unboldmath{} & \boldmath{}\textbf{0.00E+00 (0.00E+00)$\approx$}\unboldmath{} & \boldmath{}\textbf{0.00E+00 (0.00E+00)$\approx$}\unboldmath{} \\
$F_{6}$ & 1.33E-02 (8.36E-03) & \boldmath{}\textbf{4.50E-04 (4.44E-04)$+$}\unboldmath{} & 6.13E-02 (8.03E-02)$-$ & 1.28E-02 (8.82E-03)$\approx$ \\
$F_{7}$ & \textbf{0.00E+00 (0.00E+00)} & \boldmath{}\textbf{0.00E+00 (0.00E+00)$\approx$}\unboldmath{} & 2.17E-06 (4.33E-06)$-$ & \boldmath{}\textbf{0.00E+00 (0.00E+00)$\approx$}\unboldmath{} \\
$F_{8}$ & 7.45E-03 (7.36E-03) & \boldmath{}\textbf{2.09E-05 (7.77E-06)$+$}\unboldmath{} & 6.81E-04 (6.80E-04)$+$ & 3.02E-03 (2.19E-03)$+$ \\
$F_{9}$ & 1.75E-06 (0.00E+00) & 1.86E+02 (0.00E+00)$-$ & 2.29E+02 (0.00E+00)$-$ & \boldmath{}\textbf{0.00E+00 (0.00E+00)$+$}\unboldmath{} \\
$F_{10}$ & \textbf{0.00E+00 (0.00E+00)} & 1.00E+02 (0.00E+00)$-$ & 1.00E+02 (0.00E+00)$-$ & 2.00E+01 (4.00E+01)$-$ \\
$F_{11}$ & 7.71E-06 (9.16E-06) & \boldmath{}\textbf{0.00E+00 (0.00E+00)$+$}\unboldmath{} & \boldmath{}\textbf{0.00E+00 (0.00E+00)$+$}\unboldmath{} & \boldmath{}\textbf{0.00E+00 (0.00E+00)$+$}\unboldmath{} \\
$F_{12}$ & \textbf{6.35E+01 (7.77E+01)} & 1.50E+02 (6.39E+00)$-$ & 1.65E+02 (3.60E-03)$-$ & 1.65E+02 (1.81E-01)$-$ \\
\midrule
$+$ / $\approx$ / $-$ & --    & 3/4/5 & 2/5/5 & 3/6/3 \\
\bottomrule
\end{tabular}%

\footnotesize
\textsuperscript{*} The Wilcoxon rank-sum tests (with a significance level of 0.05) were conducted between iStratDE and each algorithm individually.
The final row displays the number of problems where the corresponding algorithm performs statistically better ($+$), similar ($\thickapprox$), or worse ($-$) compared to iStratDE.\\

\label{tab:vsCECTop 10D}%
\end{table*}%
\begin{table*}[!t]
  \centering
  
  \caption{{Comparisons between iStratDE and the top 3 algorithms from CEC2022 Competition (20D). 
The mean and standard deviation (in parentheses) of the results over multiple runs are displayed in pairs. 
Results with the best mean values are highlighted.
  }
  }
  \footnotesize
\begin{tabular}{ccccc}
\toprule
Func  & iStratDE & EA4eig & NL-SHADE-LBC & NL-SHADE-RSP \\
\midrule
$F_{1}$ & \textbf{0.00E+00 (0.00E+00)} & \boldmath{}\textbf{0.00E+00 (0.00E+00)$\approx$}\unboldmath{} & \boldmath{}\textbf{0.00E+00 (0.00E+00)$\approx$}\unboldmath{} & \boldmath{}\textbf{0.00E+00 (0.00E+00)$\approx$}\unboldmath{} \\
$F_{2}$ & \textbf{0.00E+00 (0.00E+00)} & \boldmath{}\textbf{0.00E+00 (0.00E+00)$\approx$}\unboldmath{} & 4.91E+01 (0.00E+00)$-$ & \boldmath{}\textbf{0.00E+00 (0.00E+00)$\approx$}\unboldmath{} \\
$F_{3}$ & \textbf{0.00E+00 (0.00E+00)} & \boldmath{}\textbf{0.00E+00 (0.00E+00)$\approx$}\unboldmath{} & \boldmath{}\textbf{0.00E+00 (0.00E+00)$\approx$}\unboldmath{} & \boldmath{}\textbf{0.00E+00 (0.00E+00)$\approx$}\unboldmath{} \\
$F_{4}$ & 1.04E+01 (2.41E+00) & 7.36E+00 (1.85E+00)$+$ & \boldmath{}\textbf{3.18E+00 (9.75E-01)$+$}\unboldmath{} & 2.83E+01 (3.31E+00)$-$ \\
$F_{5}$ & \textbf{0.00E+00 (0.00E+00)} & \boldmath{}\textbf{0.00E+00 (0.00E+00)$\approx$}\unboldmath{} & \boldmath{}\textbf{0.00E+00 (0.00E+00)$\approx$}\unboldmath{} & 9.09E-02 (1.82E-01)$-$ \\
$F_{6}$ & 3.53E-01 (3.05E-01)& 2.54E-01 (3.83E-01)$\approx$ & 1.23E+00 (7.97E-01)$-$ & \boldmath{}\textbf{1.58E-01 (5.64E-02)$+$}\unboldmath{} \\
$F_{7}$ & 1.09E+00 (1.07E+00) & 1.17E+00 (8.71E-01)$\approx$ & \boldmath{}\textbf{6.24E-02 (1.25E-01)$+$}\unboldmath{} & 5.37E+00 (7.95E+00)$-$ \\
$F_{8}$ & 7.21E+00 (9.40E+00) & 2.02E+01 (1.15E-01)$-$ & \boldmath{}\textbf{6.32E+00 (7.68E+00)$\approx$}\unboldmath{} & 1.73E+01 (5.38E+00)$-$ \\
$F_{9}$ & \textbf{1.00E+02 (0.00E+00)} & 1.65E+02 (0.00E+00)$-$ & 1.81E+02 (0.00E+00)$-$ & 1.81E+02 (0.00E+00)$-$ \\
$F_{10}$ & \textbf{0.00E+00 (0.00E+00)} & 1.23E+02 (4.58E+01)$-$ & 1.00E+02 (2.84E-02)$-$ & \boldmath{}\textbf{0.00E+00 (0.00E+00)$\approx$}\unboldmath{} \\
$F_{11}$ & 1.41E-05 (2.37E-05) & 3.20E+02 (4.00E+01)$-$ & 3.00E+02 (0.00E+00)$-$ & \boldmath{}\textbf{0.00E+00 (0.00E+00)$+$}\unboldmath{} \\
$F_{12}$ & 2.31E+02 (1.40E+00) & \boldmath{}\textbf{2.00E+02 (2.32E-04)$+$}\unboldmath{} & 2.35E+02 (6.60E-01)$-$ & 2.37E+02 (3.31E+00)$-$ \\
\midrule
$+$ / $\approx$ / $-$ & --    & 2/6/4 & 2/4/6 & 2/4/6 \\
\bottomrule
\end{tabular}%

\footnotesize
\textsuperscript{*} The Wilcoxon rank-sum tests (with a significance level of 0.05) were conducted between iStratDE and each algorithm individually.
The final row displays the number of problems where the corresponding algorithm performs statistically better ($+$), similar ($\thickapprox$), or worse ($-$) compared to iStratDE.\\

\label{tab:vsCECTop 20D}%
\end{table*}%
To further assess the competitiveness of iStratDE, we compare its performance with the top three algorithms from the CEC2022 competition under an identical computational budget. The FE limit was set to the number of evaluations completed by iStratDE within 60 seconds. The baselines include EA4eig~\cite{EA4eig}, NL-SHADE-LBC~\cite{NL-SHADE-LBC}, and NL-SHADE-RSP~\cite{NL-SHADE-RSP}, all of which represent state-of-the-art designs.

As summarized in Table~\ref{tab:vsCECTop 10D} and Table~\ref{tab:vsCECTop 20D}, iStratDE consistently attains comparable or superior mean fitness values across both 10D and 20D benchmark functions, which underscores its robustness across diverse problem types.
These results confirm that iStratDE is not only highly competitive against classical and adaptive DE variants but also matches or surpasses the most advanced entries from the CEC2022 competition. 

Notably, the competing algorithms are highly sophisticated: EA4eig employs eigen-decomposition to accelerate convergence on non-separable and ill-conditioned problems, while NL-SHADE-LBC and NL-SHADE-RSP extend the SHADE framework with advanced features such as adaptive population sizing, parameter control, and learning-based mechanisms for balancing exploration and exploitation. In contrast, iStratDE relies solely on a simple yet scalable mechanism of individual-level diversity and yet achieves highly competitive performance under identical evaluation budgets. This comparison highlights that structural diversity, rather than increasingly complex adaptive machinery, can serve as a powerful design principle for advancing evolutionary algorithms.

\subsection{Key Mechanisms in iStratDE}

\subsubsection{Implicit Elitism}

A central mechanism underlying this behavior is \emph{implicit elitism}. 
Unlike traditional DE variants that enforce synchronized convergence through population-wide adaptation, iStratDE does not require uniform progress across all individuals. 
Instead, it capitalizes on the disproportionate influence of a small subset of elite individuals that emerge naturally through the random assignment of strategy configurations. 
Once such elites arise, they are preferentially referenced during mutation (e.g., via \texttt{best} or \texttt{pbest}), thereby steering the population toward promising regions.\begin{figure}[htbp]
\centering
\includegraphics[width=0.95\columnwidth]{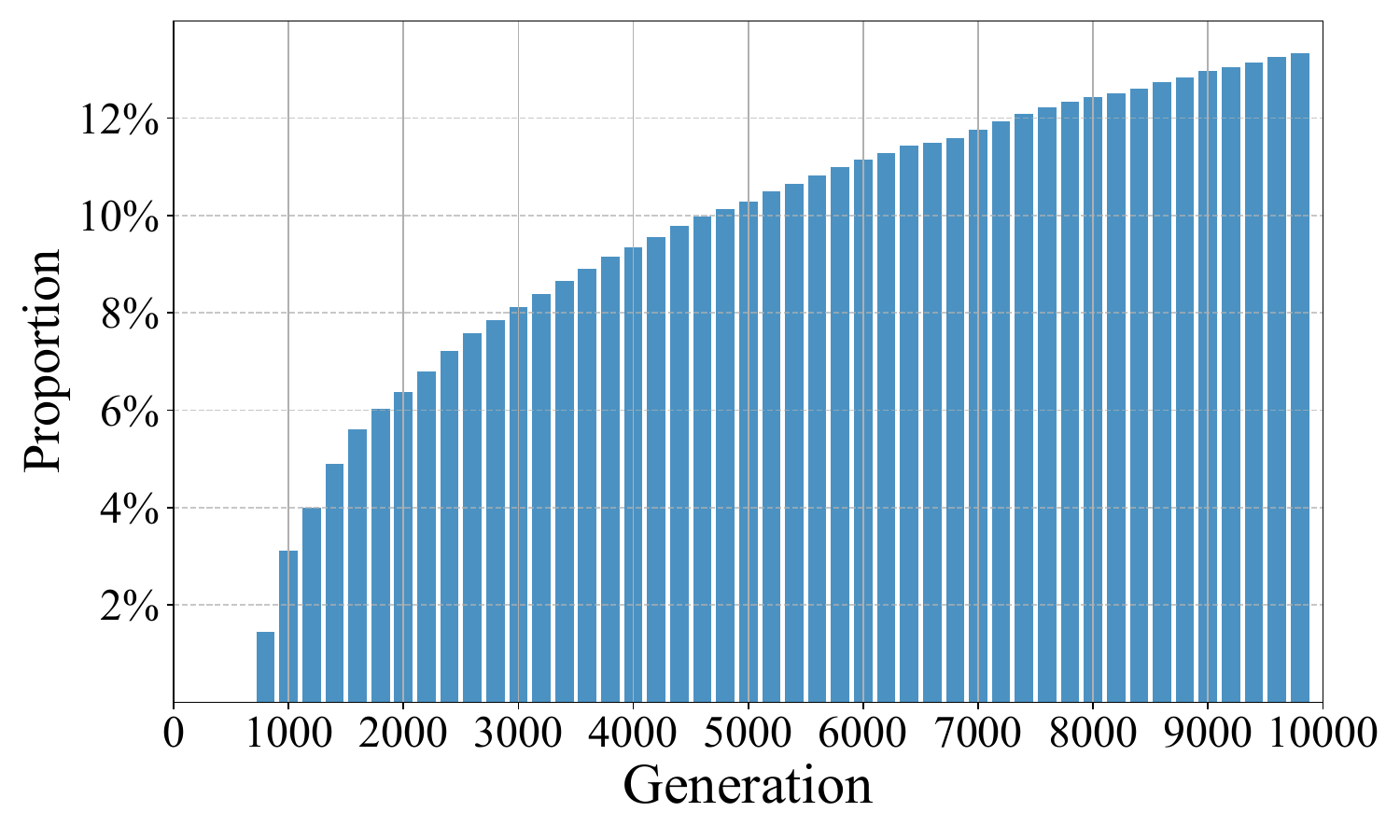}
\caption{Proportion of individuals reaching the global optimum over generations in iStratDE.}
\label{fig:elitism_bar}
\end{figure}As shown in Fig.~\ref{fig:elitism_bar}, the proportion of individuals attaining the global optimum steadily increases over generations, which indicates that only a few strong performers appearing early in the search can propel population-wide progress. This result is obtained on the relatively difficult $F_2$ problem, and similar effects are observed on other challenging functions. Although most individuals are not optimally configured, they remain valuable for sustaining structural diversity, while the well-configured few drive the search forward in a decentralized manner. 

\subsubsection{Asynchronous Convergence}

As analyzed in the previous section, implicit elitism explains how a few strong individuals can drive population-wide progress without explicit adaptation rules. 
This observation directly connects to iStratDE’s second key mechanism: \emph{asynchronous convergence}.\begin{figure}[htbp]
\centering
\includegraphics[width=0.95\columnwidth]{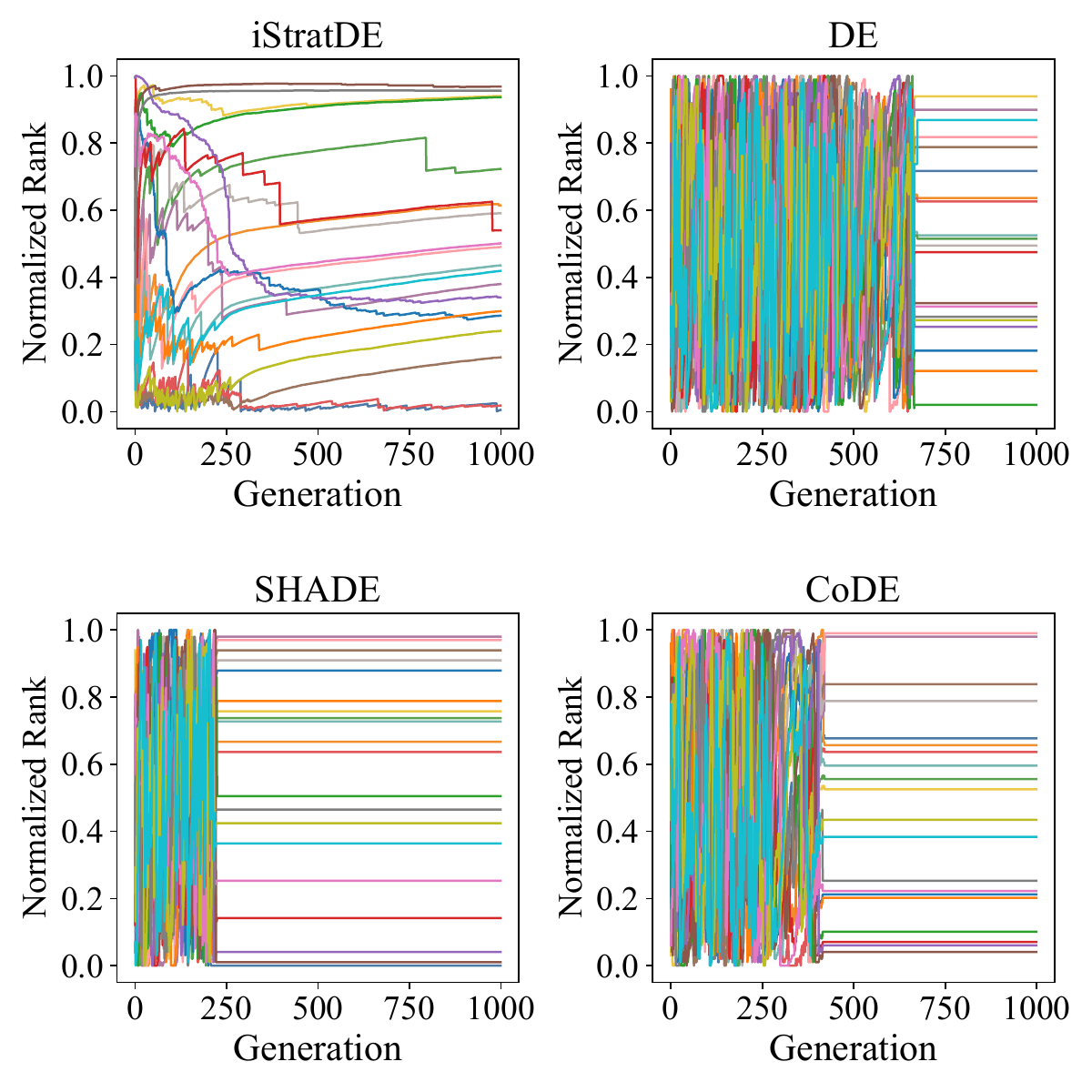}
\caption{
Evolution of normalized individual fitness ranks for 20 randomly selected individuals under different DE algorithms on a 10D benchmark problem.
The `Normalized Rank' (Y-axis) is scaled from 0.0 (best fitness rank) to 1.0 (worst fitness rank).
iStratDE exhibits asynchronous convergence: individuals improve at different rates and stabilize at different times.
In contrast, other DE variants show synchronized convergence, where most individuals settle simultaneously, reflecting stronger convergence pressure and reduced resilience to premature stagnation.
}
\label{rank_evolution_all}
\end{figure}

As illustrated in Fig.~\ref{rank_evolution_all}, the fitness rank trajectories of iStratDE individuals exhibit substantial temporal diversity: some converge rapidly within a few hundred generations, while others progress more slowly or remain stagnant for extended periods. 
This pattern demonstrates that iStratDE allows individuals to evolve independently along distinct temporal trajectories, whereas most DE variants impose strong convergence pressure that drives the entire population to converge simultaneously. 
This behavior is evident in the tightly clustered and flat trajectories of DE, SHADE, and CoDE, where individuals converge together and then stagnate. 
Such synchronized dynamics indicate strong interdependencies and limited temporal diversity, thereby increasing the risk of search collapse in complex landscapes.

By contrast, iStratDE’s randomized and fixed per-individual configurations decouple search behaviors, which prevents premature synchronization and allows each individual to explore and exploit independently. 
As a result, the population maintains distributed activity across the landscape rather than collapsing into a single region. 
This asynchronous dynamic also gives rise to two informal roles within the population: \textit{early leaders}, which quickly identify promising regions, and \textit{late leaders}, which explore broadly before contributing high-quality solutions at later stages. 
These distinct temporal contributions improve robustness and sustain exploration potential over long horizons.

\subsection{Application to Robot Control}\label{sec:expereiment_brax}

To further evaluate the real-world applicability of iStratDE and its capability to solve high-dimensional optimization problems, we apply it to three robot control tasks within the Brax framework~\cite{brax2021github}. 
Evolutionary optimization of neural networks has shown strong potential in robotic control, which makes this an important application domain for Evolutionary Reinforcement Learning (EvoRL)~\cite{such2017deep, stanley2019designing, bai2023evolutionary}. 
The selected tasks are \textit{Swimmer}, \textit{Reacher}, and \textit{Hopper}. Each involves evolving approximately 1,500 parameters of a multilayer perceptron (MLP) policy to maximize cumulative rewards\footnote{See Supplementary Document for additional details.}. 
In this setting, iStratDE employs a population of 10,000, thereby leveraging substantial concurrency to efficiently explore the high-dimensional solution space.\begin{figure}[htbp]
\centering
\includegraphics[width=\linewidth]{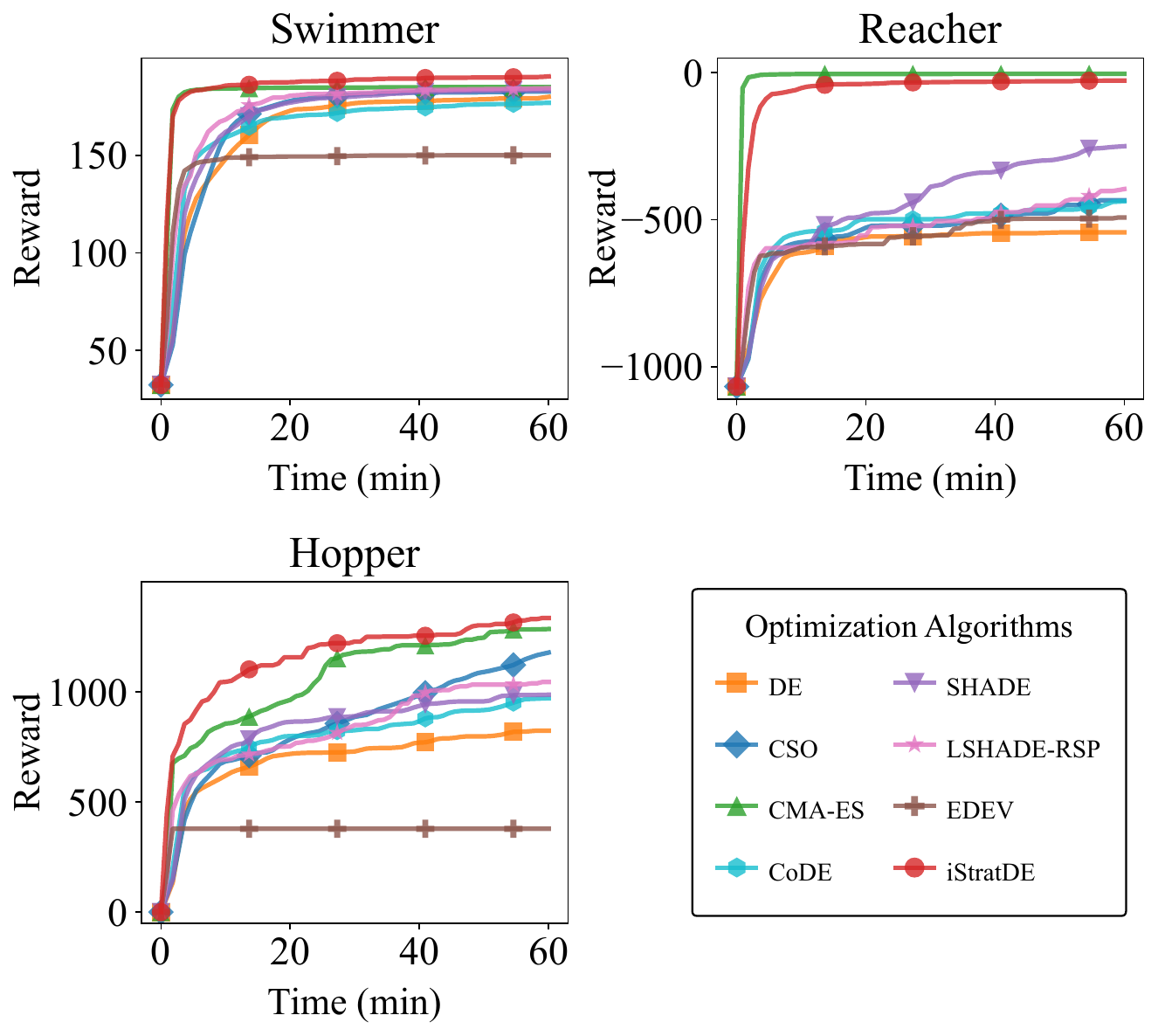}
\caption{Reward curves of iStratDE and other DE algorithms on Brax robot control tasks.}
\label{Figure_brax_all_convergence}
\end{figure}The policy models for these tasks adopt a similar neural network architecture, consisting of three fully connected layers with task-specific input and output dimensions.
Consequently, the parameter counts differ slightly, namely 1,410 for \textit{Swimmer}, 1,506 for \textit{Reacher}, and 1,539 for \textit{Hopper}.

For comparative evaluation, iStratDE was benchmarked against several representative algorithms.
This set includes mainstream DE variants such as DE \cite{DE1997}, CoDE \cite{CoDE2011}, SHADE \cite{SHADE2013}, LSHADE-RSP \cite{LSHADE-RSP2018}, and EDEV \cite{EDEV2018}.
In addition, two widely recognized algorithms with strong performance in this domain, CMA-ES \cite{CMAES} and CSO \cite{CSO}, were included as key reference baselines.
Each baseline algorithm was run independently with a population size of 100, and the termination criterion was fixed at 60 minutes to account for the computational demands of the simulations.

As shown in Fig.~\ref{Figure_brax_all_convergence}, iStratDE achieves competitive or superior performance across all three tasks, confirming its robustness in high-dimensional policy optimization.
Notably, iStratDE reliably converges to solutions near the performance upper bound, as its randomized assignment of mutation strategies and control parameters enhances exploration in search spaces susceptible to plateaus.
These findings highlight iStratDE's potential for tackling complex real-world tasks.\section{Conclusion}\label{section Conclusion}
This paper presented iStratDE, an enhanced variant of Differential Evolution (DE) that assigns randomized strategies and control parameters at the individual level. 
By cultivating population-wide heterogeneity, iStratDE achieves a more balanced exploration-exploitation trade-off and exhibits robust performance across both synthetic benchmarks and challenging robotic control tasks. 
These results demonstrate not only the practical effectiveness and scalability of iStratDE but also the value of designing evolutionary algorithms that prioritize diversity as a core principle.Several promising research directions emerge from this work. On the algorithmic side, more informed initialization (e.g., through problem-aware strategy pools or adaptive biasing) could further enhance sampling efficiency. 
On the theoretical side, a deeper study of iStratDE’s implicit elitism and asynchronous convergence could reveal general principles that extend beyond this method. 
More broadly, our findings highlight that population-level heterogeneity is not a mere design choice but a fundamental driver of robustness and adaptability in evolutionary search. This perspective suggests that future work should embrace heterogeneity as a unifying principle to tackle the growing complexity of real-world optimization problems.

\footnotesize


\newpage

\renewcommand{\algorithmicrequire}{\textbf{Input:}}
\renewcommand{\algorithmicensure}{\textbf{Output:}}


\ifCLASSINFOpdf
\else
\fi




\captionsetup{font={footnotesize}}
\captionsetup[table]{labelformat=simple, labelsep=newline, textfont=sc, justification=centering}

\title{Enhancing Differential Evolution through Individual-Level Strategy Diversity}

\markboth{Bare Demo of IEEEtran.cls for IEEE Journals}
{Shell \MakeLowercase{\textit{et al.}}: Bare Demo of IEEEtran.cls for IEEE Journals}

\setcounter{figure}{0} 
\setcounter{table}{0}

\onecolumn{}

\begin{center}
{\Huge
\renewcommand{\arraystretch}{1.1}\begin{tabular}{c}
Unleashing the Potential of Differential Evolution \\
through Individual-Level Strategy Diversity\\
(Supplementary Document)
\end{tabular}
}
\end{center}
\section{Supplementary Benchmark Results}

\subsection{Detailed Numerical Comparisons}

\begin{table}[htbp]
  \centering
  \caption{The following table details the results for 10-dimensional (10D) problems from the CEC 2022 benchmark suite. All peer Differential Evolution (DE) variants were configured with a population size of 100. Each experimental run had a uniform runtime of 60 seconds. Performance is reported as the mean and standard deviation (in parentheses) calculated over these multiple runs, presented in pairs. The symbols $+$, $-$, and $\approx$ indicate that the result is significantly better, significantly worse, or statistically similar to the iStratDE, respectively. The best mean values achieved are highlighted.}
  {
  \resizebox{\textwidth}{!}{
    \renewcommand{\arraystretch}{1.2}
    \begin{tabular}{cccccccccc}
    \toprule
    Func & iStratDE & DE & SaDE & JADE & CoDE & SHADE & LSHADE-RSP & EDEV & MetaDE \\
    \midrule
    $F_{1}$ & \multicolumn{1}{l}{\boldmath{\textbf{0.00E+00 (0.00E+00)}}\unboldmath{}} & \multicolumn{1}{l}{\boldmath{\textbf{0.00E+00 (0.00E+00)$\approx$}}\unboldmath{}} & \multicolumn{1}{l}{\boldmath{\textbf{0.00E+00 (0.00E+00)$\approx$}}\unboldmath{}} & \multicolumn{1}{l}{\boldmath{\textbf{0.00E+00 (0.00E+00)$\approx$}}\unboldmath{}} & \multicolumn{1}{l}{\boldmath{\textbf{0.00E+00 (0.00E+00)$\approx$}}\unboldmath{}} & \multicolumn{1}{l}{\boldmath{\textbf{0.00E+00 (0.00E+00)$\approx$}}\unboldmath{}} & \multicolumn{1}{l}{\boldmath{\textbf{0.00E+00 (0.00E+00)$\approx$}}\unboldmath{}} & \multicolumn{1}{l}{\boldmath{\textbf{0.00E+00 (0.00E+00)$\approx$}}\unboldmath{}} & \multicolumn{1}{l}{\boldmath{\textbf{0.00E+00 (0.00E+00)$\approx$}}\unboldmath{}} \\
    $F_{2}$ & \multicolumn{1}{l}{\boldmath{\textbf{0.00E+00 (0.00E+00)}}\unboldmath{}} & \multicolumn{1}{l}{5.89E+00 (2.40E+00)$-$} & \multicolumn{1}{l}{5.96E+00 (4.00E+00)$-$} & \multicolumn{1}{l}{5.47E+00 (3.63E+00)$-$} & \multicolumn{1}{l}{4.62E+00 (2.90E+00)$-$} & \multicolumn{1}{l}{4.69E+00 (4.15E+00)$-$} & \multicolumn{1}{l}{2.52E+00 (3.09E+00)$-$} & \multicolumn{1}{l}{6.94E+00 (2.41E+00)$-$} & \multicolumn{1}{l}{\boldmath{\textbf{0.00E+00 (0.00E+00)$\approx$}}\unboldmath{}} \\
    $F_{3}$ & \multicolumn{1}{l}{\boldmath{\textbf{0.00E+00 (0.00E+00)}}\unboldmath{}} & \multicolumn{1}{l}{\boldmath{\textbf{0.00E+00 (0.00E+00)$\approx$}}\unboldmath{}} & \multicolumn{1}{l}{\boldmath{\textbf{0.00E+00 (0.00E+00)$\approx$}}\unboldmath{}} & \multicolumn{1}{l}{\boldmath{\textbf{0.00E+00 (0.00E+00)$\approx$}}\unboldmath{}} & \multicolumn{1}{l}{\boldmath{\textbf{0.00E+00 (0.00E+00)$\approx$}}\unboldmath{}} & \multicolumn{1}{l}{\boldmath{\textbf{0.00E+00 (0.00E+00)$\approx$}}\unboldmath{}} & \multicolumn{1}{l}{\boldmath{\textbf{0.00E+00 (0.00E+00)$\approx$}}\unboldmath{}} & \multicolumn{1}{l}{\boldmath{\textbf{0.00E+00 (0.00E+00)$\approx$}}\unboldmath{}} & \multicolumn{1}{l}{\boldmath{\textbf{0.00E+00 (0.00E+00)$\approx$}}\unboldmath{}} \\
    $F_{4}$ & \multicolumn{1}{l}{\boldmath{\textbf{0.00E+00 (0.00E+00)}}\unboldmath{}} & \multicolumn{1}{l}{6.64E+00 (3.05E+00)$-$} & \multicolumn{1}{l}{9.68E-01 (8.55E-01)$-$} & \multicolumn{1}{l}{2.28E+01 (1.34E+01)$-$} & \multicolumn{1}{l}{1.06E+00 (7.12E-01)$-$} & \multicolumn{1}{l}{2.92E+00 (5.71E-01)$-$} & \multicolumn{1}{l}{3.26E+00 (1.81E+00)$-$} & \multicolumn{1}{l}{7.10E+00 (4.98E+00)$-$} & \multicolumn{1}{l}{\boldmath{\textbf{0.00E+00 (0.00E+00)$\approx$}}\unboldmath{}} \\
    $F_{5}$ & \multicolumn{1}{l}{\boldmath{\textbf{0.00E+00 (0.00E+00)}}\unboldmath{}} & \multicolumn{1}{l}{\boldmath{\textbf{0.00E+00 (0.00E+00)$\approx$}}\unboldmath{}} & \multicolumn{1}{l}{\boldmath{\textbf{0.00E+00 (0.00E+00)$\approx$}}\unboldmath{}} & \multicolumn{1}{l}{\boldmath{\textbf{0.00E+00 (0.00E+00)$\approx$}}\unboldmath{}} & \multicolumn{1}{l}{\boldmath{\textbf{0.00E+00 (0.00E+00)$\approx$}}\unboldmath{}} & \multicolumn{1}{l}{\boldmath{\textbf{0.00E+00 (0.00E+00)$\approx$}}\unboldmath{}} & \multicolumn{1}{l}{1.10E+02 (4.11E+02)$-$} & \multicolumn{1}{l}{\boldmath{\textbf{0.00E+00 (0.00E+00)$\approx$}}\unboldmath{}} & \multicolumn{1}{l}{\boldmath{\textbf{0.00E+00 (0.00E+00)$\approx$}}\unboldmath{}} \\
    $F_{6}$ & \multicolumn{1}{l}{6.44E-03 (3.72E-03)} & \multicolumn{1}{l}{2.49E-01 (1.91E-01)$-$} & \multicolumn{1}{l}{3.03E+00 (2.18E+00)$-$} & \multicolumn{1}{l}{7.85E-01 (8.47E-01)$-$} & \multicolumn{1}{l}{3.82E-03 (6.22E-03)$+$} & \multicolumn{1}{l}{1.29E+00 (2.44E+00)$-$} & \multicolumn{1}{l}{3.71E-01 (1.11E-01)$-$} & \multicolumn{1}{l}{5.80E-01 (6.43E-01)$-$} & \multicolumn{1}{l}{\boldmath{\textbf{5.50E-04 (3.96E-04)$+$}}\unboldmath{}} \\
    $F_{7}$ & \multicolumn{1}{l}{\boldmath{\textbf{0.00E+00 (0.00E+00)}}\unboldmath{}} & \multicolumn{1}{l}{7.24E-02 (2.28E-01)$-$} & \multicolumn{1}{l}{\boldmath{\textbf{0.00E+00 (0.00E+00)$\approx$}}\unboldmath{}} & \multicolumn{1}{l}{\boldmath{\textbf{0.00E+00 (0.00E+00)$\approx$}}\unboldmath{}} & \multicolumn{1}{l}{\boldmath{\textbf{0.00E+00 (0.00E+00)$\approx$}}\unboldmath{}} & \multicolumn{1}{l}{\boldmath{\textbf{0.00E+00 (0.00E+00)$\approx$}}\unboldmath{}} & \multicolumn{1}{l}{1.35E-03 (3.74E-03)$-$} & \multicolumn{1}{l}{1.09E+01 (9.86E+00)$-$} & \multicolumn{1}{l}{\boldmath{\textbf{0.00E+00 (0.00E+00)$\approx$}}\unboldmath{}} \\
    $F_{8}$ & \multicolumn{1}{l}{\boldmath{\textbf{1.10E-03 (2.98E-03)}}\unboldmath{}} & \multicolumn{1}{l}{8.43E-02 (2.08E-01)$-$} & \multicolumn{1}{l}{1.01E-01 (1.20E-01)$-$} & \multicolumn{1}{l}{1.75E+01 (4.40E+00)$-$} & \multicolumn{1}{l}{2.46E-03 (7.09E-03)$\approx$} & \multicolumn{1}{l}{1.45E+00 (4.99E+00)$-$} & \multicolumn{1}{l}{1.69E+00 (5.09E+00)$-$} & \multicolumn{1}{l}{9.68E+00 (9.89E+00)$-$} & \multicolumn{1}{l}{5.52E-03 (4.41E-03)$\approx$} \\
    $F_{9}$ & \multicolumn{1}{l}{\boldmath{\textbf{1.78E-06 (2.19E-07)}}\unboldmath{}} & \multicolumn{1}{l}{2.29E+02 (5.12E-06)$-$} & \multicolumn{1}{l}{2.29E+02 (3.75E-06)$-$} & \multicolumn{1}{l}{2.29E+02 (7.17E-06)$-$} & \multicolumn{1}{l}{2.29E+02 (7.94E-06)$-$} & \multicolumn{1}{l}{2.29E+02 (3.81E-06)$-$} & \multicolumn{1}{l}{2.29E+02 (9.43E-06)$-$} & \multicolumn{1}{l}{2.29E+02 (1.10E-05)$-$} & \multicolumn{1}{l}{3.36E+00 (1.77E+01)$-$} \\
    $F_{10}$ & \multicolumn{1}{l}{\boldmath{\textbf{0.00E+00 (0.00E+00)}}\unboldmath{}} & \multicolumn{1}{l}{1.00E+02 (4.94E-02)$-$} & \multicolumn{1}{l}{1.04E+02 (1.84E+01)$-$} & \multicolumn{1}{l}{1.00E+02 (6.81E-02)$-$} & \multicolumn{1}{l}{1.00E+02 (5.54E-02)$-$} & \multicolumn{1}{l}{1.07E+02 (2.60E+01)$-$} & \multicolumn{1}{l}{1.07E+02 (2.63E+01)$-$} & \multicolumn{1}{l}{1.00E+02 (6.33E-02)$-$} & \multicolumn{1}{l}{\boldmath{\textbf{0.00E+00 (0.00E+00)$\approx$}}\unboldmath{}} \\
    $F_{11}$ & \multicolumn{1}{l}{1.05E-05 (1.29E-05)} & \multicolumn{1}{l}{\boldmath{\textbf{0.00E+00 (0.00E+00)$+$}}\unboldmath{}} & \multicolumn{1}{l}{9.70E+00 (3.70E+01)$-$} & \multicolumn{1}{l}{\boldmath{\textbf{0.00E+00 (0.00E+00)$+$}}\unboldmath{}} & \multicolumn{1}{l}{\boldmath{\textbf{0.00E+00 (0.00E+00)$+$}}\unboldmath{}} & \multicolumn{1}{l}{2.67E+01 (9.98E+01)$-$} & \multicolumn{1}{l}{\boldmath{\textbf{0.00E+00 (0.00E+00)$+$}}\unboldmath{}} & \multicolumn{1}{l}{1.00E+01 (3.75E+01)$-$} & \multicolumn{1}{l}{\boldmath{\textbf{0.00E+00 (0.00E+00)$+$}}\unboldmath{}} \\
    $F_{12}$ & \multicolumn{1}{l}{\boldmath{\textbf{6.11E+01 (7.60E+01)}}\unboldmath{}} & \multicolumn{1}{l}{1.62E+02 (1.43E+00)$-$} & \multicolumn{1}{l}{1.63E+02 (1.55E+00)$-$} & \multicolumn{1}{l}{1.62E+02 (1.97E+00)$-$} & \multicolumn{1}{l}{1.59E+02 (1.02E+00)$-$} & \multicolumn{1}{l}{1.63E+02 (8.33E-01)$-$} & \multicolumn{1}{l}{1.64E+02 (1.09E+00)$-$} & \multicolumn{1}{l}{1.62E+02 (1.49E+00)$-$} & \multicolumn{1}{l}{1.39E+02 (4.63E+01)$-$} \\
    \midrule

    $+$ / $\approx$ / $-$ & --    & 1/3/8 & 0/4/8 & 1/4/7 & 2/5/5 & 0/4/8 & 1/2/9 & 0/3/9 & 2/8/2 \\
    \bottomrule
    \end{tabular}
  }
  \footnotesize
  \label{tab:my_detailed_results_10D_wilcoxon_generated}
  }
\end{table}
\begin{table}[htbp]
  \centering
  \caption{The following table details the results for 20-dimensional (20D) problems from the CEC 2022 benchmark suite. All peer Differential Evolution (DE) variants were configured with a population size of 100. Each experimental run had a uniform runtime of 60 seconds. Performance is reported as the mean and standard deviation (in parentheses) calculated over these multiple runs, presented in pairs. The symbols $+$, $-$, and $\approx$ indicate that the result is significantly better, significantly worse, or statistically similar to the iStratDE, respectively. The best mean values achieved are highlighted.}
  {
  \resizebox{\textwidth}{!}{%
    \renewcommand{\arraystretch}{1.2}
    \begin{tabular}{cccccccccc}
    \toprule
    Func & iStratDE & DE & SaDE & JADE & CoDE & SHADE & LSHADE-RSP & EDEV & MetaDE \\
    \midrule
    $F_{1}$ & \multicolumn{1}{l}{\boldmath{\textbf{0.00E+00 (0.00E+00)}}\unboldmath{}} & \multicolumn{1}{l}{\boldmath{\textbf{0.00E+00 (0.00E+00)$\approx$}}\unboldmath{}} & \multicolumn{1}{l}{\boldmath{\textbf{0.00E+00 (0.00E+00)$\approx$}}\unboldmath{}} & \multicolumn{1}{l}{\boldmath{\textbf{0.00E+00 (0.00E+00)$\approx$}}\unboldmath{}} & \multicolumn{1}{l}{\boldmath{\textbf{0.00E+00 (0.00E+00)$\approx$}}\unboldmath{}} & \multicolumn{1}{l}{\boldmath{\textbf{0.00E+00 (0.00E+00)$\approx$}}\unboldmath{}} & \multicolumn{1}{l}{\boldmath{\textbf{0.00E+00 (0.00E+00)$\approx$}}\unboldmath{}} & \multicolumn{1}{l}{\boldmath{\textbf{0.00E+00 (0.00E+00)$\approx$}}\unboldmath{}} & \multicolumn{1}{l}{\boldmath{\textbf{0.00E+00 (0.00E+00)$\approx$}}\unboldmath{}} \\
    $F_{2}$ & \multicolumn{1}{l}{\boldmath{\textbf{0.00E+00 (0.00E+00)}}\unboldmath{}} & \multicolumn{1}{l}{4.77E+01 (1.97E+00)$-$} & \multicolumn{1}{l}{3.27E+01 (2.31E+01)$-$} & \multicolumn{1}{l}{4.91E+01 (2.53E-05)$-$} & \multicolumn{1}{l}{4.91E+01 (1.52E-05)$-$} & \multicolumn{1}{l}{4.58E+01 (1.22E+01)$-$} & \multicolumn{1}{l}{3.00E+02 (9.65E+02)$-$} & \multicolumn{1}{l}{4.23E+01 (1.66E+01)$-$} & \multicolumn{1}{l}{1.26E-02 (3.74E-02)$-$} \\
    $F_{3}$ & \multicolumn{1}{l}{\boldmath{\textbf{0.00E+00 (0.00E+00)}}\unboldmath{}} & \multicolumn{1}{l}{\boldmath{\textbf{0.00E+00 (0.00E+00)$\approx$}}\unboldmath{}} & \multicolumn{1}{l}{\boldmath{\textbf{0.00E+00 (0.00E+00)$\approx$}}\unboldmath{}} & \multicolumn{1}{l}{\boldmath{\textbf{0.00E+00 (0.00E+00)$\approx$}}\unboldmath{}} & \multicolumn{1}{l}{\boldmath{\textbf{0.00E+00 (0.00E+00)$\approx$}}\unboldmath{}} & \multicolumn{1}{l}{\boldmath{\textbf{0.00E+00 (0.00E+00)$\approx$}}\unboldmath{}} & \multicolumn{1}{l}{7.58E+00 (2.84E+01)$\approx$} & \multicolumn{1}{l}{1.27E-04 (4.68E-04)$-$} & \multicolumn{1}{l}{\boldmath{\textbf{0.00E+00 (0.00E+00)$\approx$}}\unboldmath{}} \\
    $F_{4}$ & \multicolumn{1}{l}{9.62E+00 (2.59E+00)} & \multicolumn{1}{l}{2.07E+01 (9.21E+00)$-$} & \multicolumn{1}{l}{8.90E+00 (1.90E+00)$+$} & \multicolumn{1}{l}{7.00E+01 (3.24E+01)$-$} & \multicolumn{1}{l}{9.88E+00 (2.01E+00)$\approx$} & \multicolumn{1}{l}{1.27E+01 (2.63E+00)$-$} & \multicolumn{1}{l}{1.94E+01 (2.08E+01)$-$} & \multicolumn{1}{l}{2.28E+01 (6.47E+00)$-$} & \multicolumn{1}{l}{\boldmath{\textbf{2.02E+00 (8.56E-01)$+$}}\unboldmath{}} \\
    $F_{5}$ & \multicolumn{1}{l}{\boldmath{\textbf{0.00E+00 (0.00E+00)}}\unboldmath{}} & \multicolumn{1}{l}{\boldmath{\textbf{0.00E+00 (0.00E+00)$\approx$}}\unboldmath{}} & \multicolumn{1}{l}{1.51E+00 (1.90E+00)$-$} & \multicolumn{1}{l}{\boldmath{\textbf{0.00E+00 (0.00E+00)$\approx$}}\unboldmath{}} & \multicolumn{1}{l}{\boldmath{\textbf{0.00E+00 (0.00E+00)$\approx$}}\unboldmath{}} & \multicolumn{1}{l}{\boldmath{\textbf{0.00E+00 (0.00E+00)$\approx$}}\unboldmath{}} & \multicolumn{1}{l}{5.73E+02 (2.15E+03)$-$} & \multicolumn{1}{l}{1.69E-01 (3.02E-01)$-$} & \multicolumn{1}{l}{\boldmath{\textbf{0.00E+00 (0.00E+00)$\approx$}}\unboldmath{}} \\
    $F_{6}$ & \multicolumn{1}{l}{3.97E-01 (3.00E-01)} & \multicolumn{1}{l}{4.60E-01 (3.83E-01)$\approx$} & \multicolumn{1}{l}{1.65E+03 (1.35E+03)$-$} & \multicolumn{1}{l}{3.63E+02 (7.80E+02)$-$} & \multicolumn{1}{l}{6.51E-01 (1.25E+00)$-$} & \multicolumn{1}{l}{6.53E+01 (3.34E+01)$-$} & \multicolumn{1}{l}{1.89E+01 (7.08E+00)$-$} & \multicolumn{1}{l}{2.31E+03 (3.91E+03)$-$} & \multicolumn{1}{l}{\boldmath{\textbf{1.16E-01 (2.79E-02)$+$}}\unboldmath{}} \\
    $F_{7}$ & \multicolumn{1}{l}{1.60E+00 (2.59E+00)} & \multicolumn{1}{l}{7.11E+00 (9.51E+00)$-$} & \multicolumn{1}{l}{1.52E+01 (4.89E+00)$-$} & \multicolumn{1}{l}{4.23E+00 (6.32E+00)$-$} & \multicolumn{1}{l}{2.03E+00 (3.81E+00)$-$} & \multicolumn{1}{l}{1.03E+01 (7.34E+00)$-$} & \multicolumn{1}{l}{2.40E+01 (1.15E+01)$-$} & \multicolumn{1}{l}{2.72E+01 (7.52E+00)$-$} & \multicolumn{1}{l}{\boldmath{\textbf{5.17E-02 (6.21E-02)$+$}}\unboldmath{}} \\
    $F_{8}$ & \multicolumn{1}{l}{6.18E+00 (8.96E+00)} & \multicolumn{1}{l}{5.81E+00 (8.78E+00)$\approx$} & \multicolumn{1}{l}{2.20E+01 (3.46E-01)$-$} & \multicolumn{1}{l}{2.65E+01 (1.18E+00)$-$} & \multicolumn{1}{l}{1.45E+01 (8.75E+00)$-$} & \multicolumn{1}{l}{2.06E+01 (1.52E-01)$-$} & \multicolumn{1}{l}{2.20E+01 (2.05E+00)$-$} & \multicolumn{1}{l}{1.96E+01 (5.02E+00)$-$} & \multicolumn{1}{l}{\boldmath{\textbf{7.19E-01 (1.02E+00)$+$}}\unboldmath{}} \\
    $F_{9}$ & \multicolumn{1}{l}{\boldmath{\textbf{1.00E+02 (0.00E+00)}}\unboldmath{}} & \multicolumn{1}{l}{1.81E+02 (1.50E-05)$-$} & \multicolumn{1}{l}{1.81E+02 (5.88E-05)$-$} & \multicolumn{1}{l}{1.81E+02 (6.02E-05)$-$} & \multicolumn{1}{l}{1.81E+02 (0.00E+00)$-$} & \multicolumn{1}{l}{1.81E+02 (2.00E-05)$-$} & \multicolumn{1}{l}{2.39E+02 (2.19E+02)$-$} & \multicolumn{1}{l}{1.81E+02 (1.69E-04)$-$} & \multicolumn{1}{l}{1.07E+02 (1.98E+01)$-$} \\
    $F_{10}$ & \multicolumn{1}{l}{\boldmath{\textbf{0.00E+00 (0.00E+00)}}\unboldmath{}} & \multicolumn{1}{l}{1.31E+02 (8.79E+01)$-$} & \multicolumn{1}{l}{1.00E+02 (4.17E-02)$-$} & \multicolumn{1}{l}{1.03E+02 (4.35E+01)$-$} & \multicolumn{1}{l}{1.00E+02 (3.55E-02)$-$} & \multicolumn{1}{l}{1.08E+02 (2.94E+01)$-$} & \multicolumn{1}{l}{1.00E+02 (7.56E-02)$-$} & \multicolumn{1}{l}{1.12E+02 (6.28E+01)$-$} & \multicolumn{1}{l}{\boldmath{\textbf{0.00E+00 (0.00E+00)$\approx$}}\unboldmath{}} \\
    $F_{11}$ & \multicolumn{1}{l}{\boldmath{\textbf{1.41E-05 (2.37E-05)}}\unboldmath{}} & \multicolumn{1}{l}{3.47E+02 (4.99E+01)$-$} & \multicolumn{1}{l}{3.13E+02 (3.40E+01)$-$} & \multicolumn{1}{l}{3.13E+02 (3.40E+01)$-$} & \multicolumn{1}{l}{3.13E+02 (3.40E+01)$-$} & \multicolumn{1}{l}{3.27E+02 (4.42E+01)$-$} & \multicolumn{1}{l}{3.78E+02 (1.16E+02)$-$} & \multicolumn{1}{l}{3.07E+02 (2.49E+01)$-$} & \multicolumn{1}{l}{7.28E-05 (3.06E-04)$-$} \\
    $F_{12}$ & \multicolumn{1}{l}{2.31E+02 (1.59E+00)} & \multicolumn{1}{l}{2.39E+02 (3.71E+00)$-$} & \multicolumn{1}{l}{2.40E+02 (4.63E+00)$-$} & \multicolumn{1}{l}{2.37E+02 (4.60E+00)$-$} & \multicolumn{1}{l}{2.35E+02 (3.24E+00)$-$} & \multicolumn{1}{l}{2.40E+02 (3.56E+00)$-$} & \multicolumn{1}{l}{2.46E+02 (1.26E+01)$-$} & \multicolumn{1}{l}{2.41E+02 (3.93E+00)$-$} & \multicolumn{1}{l}{\boldmath{\textbf{2.29E+02 (6.08E-01)$+$}}\unboldmath{}} \\
    \midrule
    $+$ / $\approx$ / $-$ & --    & 0/5/7 & 1/2/9 & 0/3/9 & 0/4/8 & 0/3/9 & 0/2/10 & 0/1/11 & 5/4/3 \\
    \bottomrule
    \end{tabular}%
  }
  \footnotesize
  \label{tab:my_detailed_results_20D_wilcoxon_generated}
  }
\end{table}
\clearpage
\begin{table}[htbp]
  \centering
  \caption{The following table details the results for 10-dimensional (10D) problems from the CEC 2022 benchmark suite. Peer Differential Evolution (DE) variants were configured with a population size of 100. All algorithms were run for a uniform number of Function Evaluations (FEs). This FE budget was set based on the FEs achieved by iStratDE (using a population size of 100,000) within a 60-second execution. Performance is reported as the mean and standard deviation (in parentheses) calculated over multiple runs, presented in pairs. The symbols $+$, $-$, and $\approx$ indicate that the result is significantly better, significantly worse, or statistically similar to the iStratDE, respectively. The best mean values achieved are highlighted.}
  {
  \resizebox{\textwidth}{!}{%
    \renewcommand{\arraystretch}{1.2}
    \begin{tabular}{cccccccccc}
    \toprule
    Func & iStratDE & DE & SaDE & JADE & CoDE & SHADE & LSHADE-RSP & EDEV & MetaDE \\
    \midrule
    $F_{1}$ & \multicolumn{1}{l}{\boldmath{\textbf{0.00E+00 (0.00E+00)}}\unboldmath{}} & \multicolumn{1}{l}{\boldmath{\textbf{0.00E+00 (0.00E+00)$\approx$}}\unboldmath{}} & \multicolumn{1}{l}{\boldmath{\textbf{0.00E+00 (0.00E+00)$\approx$}}\unboldmath{}} & \multicolumn{1}{l}{\boldmath{\textbf{0.00E+00 (0.00E+00)$\approx$}}\unboldmath{}} & \multicolumn{1}{l}{\boldmath{\textbf{0.00E+00 (0.00E+00)$\approx$}}\unboldmath{}} & \multicolumn{1}{l}{\boldmath{\textbf{0.00E+00 (0.00E+00)$\approx$}}\unboldmath{}} & \multicolumn{1}{l}{\boldmath{\textbf{0.00E+00 (0.00E+00)$\approx$}}\unboldmath{}} & \multicolumn{1}{l}{\boldmath{\textbf{0.00E+00 (0.00E+00)$\approx$}}\unboldmath{}} & \multicolumn{1}{l}{\boldmath{\textbf{0.00E+00 (0.00E+00)$\approx$}}\unboldmath{}} \\
    $F_{2}$ & \multicolumn{1}{l}{\boldmath{\textbf{0.00E+00 (0.00E+00)}}\unboldmath{}} & \multicolumn{1}{l}{5.95E+00 (2.37E+00)$-$} & \multicolumn{1}{l}{5.95E+00 (4.48E+00)$-$} & \multicolumn{1}{l}{6.38E+00 (3.21E+00)$-$} & \multicolumn{1}{l}{5.36E+00 (2.15E+00)$-$} & \multicolumn{1}{l}{4.75E+00 (3.68E+00)$-$} & \multicolumn{1}{l}{3.87E+00 (3.25E+00)$-$} & \multicolumn{1}{l}{5.04E+00 (2.94E+00)$-$} & \multicolumn{1}{l}{\boldmath{\textbf{0.00E+00 (0.00E+00)$\approx$}}\unboldmath{}} \\
    $F_{3}$ & \multicolumn{1}{l}{\boldmath{\textbf{0.00E+00 (0.00E+00)}}\unboldmath{}} & \multicolumn{1}{l}{\boldmath{\textbf{0.00E+00 (0.00E+00)$\approx$}}\unboldmath{}} & \multicolumn{1}{l}{\boldmath{\textbf{0.00E+00 (0.00E+00)$\approx$}}\unboldmath{}} & \multicolumn{1}{l}{\boldmath{\textbf{0.00E+00 (0.00E+00)$\approx$}}\unboldmath{}} & \multicolumn{1}{l}{\boldmath{\textbf{0.00E+00 (0.00E+00)$\approx$}}\unboldmath{}} & \multicolumn{1}{l}{\boldmath{\textbf{0.00E+00 (0.00E+00)$\approx$}}\unboldmath{}} & \multicolumn{1}{l}{\boldmath{\textbf{0.00E+00 (0.00E+00)$\approx$}}\unboldmath{}} & \multicolumn{1}{l}{\boldmath{\textbf{0.00E+00 (0.00E+00)$\approx$}}\unboldmath{}} & \multicolumn{1}{l}{\boldmath{\textbf{0.00E+00 (0.00E+00)$\approx$}}\unboldmath{}} \\
    $F_{4}$ & \multicolumn{1}{l}{\boldmath{\textbf{0.00E+00 (0.00E+00)}}\unboldmath{}} & \multicolumn{1}{l}{7.90E+00 (3.35E+00)$-$} & \multicolumn{1}{l}{7.30E-01 (5.71E-01)$-$} & \multicolumn{1}{l}{2.49E+01 (1.21E+01)$-$} & \multicolumn{1}{l}{6.63E-01 (5.93E-01)$-$} & \multicolumn{1}{l}{2.92E+00 (5.71E-01)$-$} & \multicolumn{1}{l}{1.78E+00 (6.60E-01)$-$} & \multicolumn{1}{l}{4.61E+00 (1.76E+00)$-$} & \multicolumn{1}{l}{\boldmath{\textbf{0.00E+00 (0.00E+00)$\approx$}}\unboldmath{}} \\
    $F_{5}$ & \multicolumn{1}{l}{\boldmath{\textbf{0.00E+00 (0.00E+00)}}\unboldmath{}} & \multicolumn{1}{l}{\boldmath{\textbf{0.00E+00 (0.00E+00)$\approx$}}\unboldmath{}} & \multicolumn{1}{l}{5.97E-03 (2.23E-02)$\approx$} & \multicolumn{1}{l}{\boldmath{\textbf{0.00E+00 (0.00E+00)$\approx$}}\unboldmath{}} & \multicolumn{1}{l}{\boldmath{\textbf{0.00E+00 (0.00E+00)$\approx$}}\unboldmath{}} & \multicolumn{1}{l}{\boldmath{\textbf{0.00E+00 (0.00E+00)$\approx$}}\unboldmath{}} & \multicolumn{1}{l}{\boldmath{\textbf{0.00E+00 (0.00E+00)$\approx$}}\unboldmath{}} & \multicolumn{1}{l}{\boldmath{\textbf{0.00E+00 (0.00E+00)$\approx$}}\unboldmath{}} & \multicolumn{1}{l}{\boldmath{\textbf{0.00E+00 (0.00E+00)$\approx$}}\unboldmath{}} \\
    $F_{6}$ & \multicolumn{1}{l}{1.33E-02 (8.36E-03)} & \multicolumn{1}{l}{1.08E-01 (1.92E-01)$-$} & \multicolumn{1}{l}{3.42E+00 (3.10E+00)$-$} & \multicolumn{1}{l}{5.83E-01 (4.15E-01)$-$} & \multicolumn{1}{l}{\boldmath{\textbf{0.00E+00 (0.00E+00)$+$}}\unboldmath{}} & \multicolumn{1}{l}{1.46E+00 (4.16E+00)$-$} & \multicolumn{1}{l}{4.20E-06 (1.54E-05)$+$} & \multicolumn{1}{l}{4.30E-01 (5.44E-01)$-$} & \multicolumn{1}{l}{5.50E-04 (3.96E-04)$+$} \\
    $F_{7}$ & \multicolumn{1}{l}{\boldmath{\textbf{0.00E+00 (0.00E+00)}}\unboldmath{}} & \multicolumn{1}{l}{1.25E-01 (2.50E-01)$-$} & \multicolumn{1}{l}{\boldmath{\textbf{0.00E+00 (0.00E+00)$\approx$}}\unboldmath{}} & \multicolumn{1}{l}{\boldmath{\textbf{0.00E+00 (0.00E+00)$\approx$}}\unboldmath{}} & \multicolumn{1}{l}{\boldmath{\textbf{0.00E+00 (0.00E+00)$\approx$}}\unboldmath{}} & \multicolumn{1}{l}{\boldmath{\textbf{0.00E+00 (0.00E+00)$\approx$}}\unboldmath{}} & \multicolumn{1}{l}{\boldmath{\textbf{0.00E+00 (0.00E+00)$\approx$}}\unboldmath{}} & \multicolumn{1}{l}{1.31E+01 (9.28E+00)$-$} & \multicolumn{1}{l}{\boldmath{\textbf{0.00E+00 (0.00E+00)$\approx$}}\unboldmath{}} \\
    $F_{8}$ & \multicolumn{1}{l}{7.45E-03 (7.36E-03)} & \multicolumn{1}{l}{1.29E-01 (2.48E-01)$-$} & \multicolumn{1}{l}{1.54E-03 (5.78E-03)$+$} & \multicolumn{1}{l}{1.83E+01 (4.05E+00)$-$} & \multicolumn{1}{l}{\boldmath{\textbf{9.83E-05 (2.54E-04)$+$}}\unboldmath{}} & \multicolumn{1}{l}{1.49E-02 (2.07E-02)$-$} & \multicolumn{1}{l}{1.33E-01 (2.55E-01)$-$} & \multicolumn{1}{l}{9.35E+00 (1.00E+01)$-$} & \multicolumn{1}{l}{5.52E-03 (4.41E-03)$\approx$} \\
    $F_{9}$ & \multicolumn{1}{l}{\boldmath{\textbf{1.75E-06 (0.00E+00)}}\unboldmath{}} & \multicolumn{1}{l}{2.29E+02 (3.92E-06)$-$} & \multicolumn{1}{l}{2.29E+02 (8.77E-06)$-$} & \multicolumn{1}{l}{2.29E+02 (4.33E-06)$-$} & \multicolumn{1}{l}{2.29E+02 (6.19E-06)$-$} & \multicolumn{1}{l}{2.29E+02 (0.00E+00)$-$} & \multicolumn{1}{l}{2.29E+02 (8.65E-06)$-$} & \multicolumn{1}{l}{2.29E+02 (4.84E-06)$-$} & \multicolumn{1}{l}{3.36E+00 (1.77E+01)$-$} \\
    $F_{10}$ & \multicolumn{1}{l}{\boldmath{\textbf{0.00E+00 (0.00E+00)}}\unboldmath{}} & \multicolumn{1}{l}{1.00E+02 (4.65E-02)$-$} & \multicolumn{1}{l}{1.20E+02 (2.65E+01)$-$} & \multicolumn{1}{l}{1.08E+02 (2.93E+01)$-$} & \multicolumn{1}{l}{1.00E+02 (0.00E+00)$-$} & \multicolumn{1}{l}{1.03E+02 (2.58E+01)$-$} & \multicolumn{1}{l}{1.00E+02 (0.00E+00)$-$} & \multicolumn{1}{l}{1.01E+02 (3.04E+01)$-$} & \multicolumn{1}{l}{\boldmath{\textbf{0.00E+00 (0.00E+00)$\approx$}}\unboldmath{}} \\
    $F_{11}$ & \multicolumn{1}{l}{7.71E-06 (9.16E-06)} & \multicolumn{1}{l}{\boldmath{\textbf{0.00E+00 (0.00E+00)$+$}}\unboldmath{}} & \multicolumn{1}{l}{\boldmath{\textbf{0.00E+00 (0.00E+00)$+$}}\unboldmath{}} & \multicolumn{1}{l}{\boldmath{\textbf{0.00E+00 (0.00E+00)$+$}}\unboldmath{}} & \multicolumn{1}{l}{\boldmath{\textbf{0.00E+00 (0.00E+00)$+$}}\unboldmath{}} & \multicolumn{1}{l}{2.67E+01 (9.98E+01)$-$} & \multicolumn{1}{l}{\boldmath{\textbf{0.00E+00 (0.00E+00)$+$}}\unboldmath{}} & \multicolumn{1}{l}{4.10E+01 (9.25E+01)$-$} & \multicolumn{1}{l}{\boldmath{\textbf{0.00E+00 (0.00E+00)$+$}}\unboldmath{}} \\
    $F_{12}$ & \multicolumn{1}{l}{\boldmath{\textbf{6.35E+01 (7.77E+01)}}\unboldmath{}} & \multicolumn{1}{l}{1.62E+02 (1.52E+00)$-$} & \multicolumn{1}{l}{1.63E+02 (1.70E+00)$-$} & \multicolumn{1}{l}{1.62E+02 (2.31E+00)$-$} & \multicolumn{1}{l}{1.59E+02 (3.68E-01)$-$} & \multicolumn{1}{l}{1.63E+02 (8.33E-01)$-$} & \multicolumn{1}{l}{1.64E+02 (1.56E+00)$-$} & \multicolumn{1}{l}{1.63E+02 (9.38E-01)$-$} & \multicolumn{1}{l}{1.39E+02 (4.63E+01)$-$} \\
    \midrule
    $+$ / $\approx$ / $-$ & --    & 1/3/8 & 2/4/6 & 1/4/7 & 3/4/5 & 0/4/8 & 2/4/6 & 0/3/9 & 2/8/2 \\
    \bottomrule
    \end{tabular}%
  }
  \footnotesize
\label{tab:my_detailed_results_10D_wilcoxon_generated_same_FEs}
  }
\end{table}
\begin{table}[htbp]
  \centering
  \caption{The following table details the results for 20-dimensional (20D) problems from the CEC 2022 benchmark suite. Peer Differential Evolution (DE) variants were configured with a population size of 100. All algorithms were run for a uniform number of Function Evaluations (FEs). This FE budget was set based on the FEs achieved by iStratDE (using a population size of 100,000) within a 60-second execution. Performance is reported as the mean and standard deviation (in parentheses) calculated over multiple runs, presented in pairs. The symbols $+$, $-$, and $\approx$ indicate that the result is significantly better, significantly worse, or statistically similar to the iStratDE, respectively. The best mean values achieved are highlighted.}
  {
  \resizebox{\textwidth}{!}{
    \renewcommand{\arraystretch}{1.2}
    \begin{tabular}{cccccccccc}
    \toprule
    Func & iStratDE & DE & SaDE & JADE & CoDE & SHADE & LSHADE-RSP & EDEV & MetaDE \\
    \midrule
    $F_{1}$ & \multicolumn{1}{l}{\textbf{0.00E+00 (0.00E+00)}} & \multicolumn{1}{l}{\boldmath{\textbf{0.00E+00 (0.00E+00)$\approx$}}\unboldmath{}} & \multicolumn{1}{l}{\boldmath{\textbf{0.00E+00 (0.00E+00)$\approx$}}\unboldmath{}} & \multicolumn{1}{l}{\boldmath{\textbf{0.00E+00 (0.00E+00)$\approx$}}\unboldmath{}} & \multicolumn{1}{l}{\boldmath{\textbf{0.00E+00 (0.00E+00)$\approx$}}\unboldmath{}} & \multicolumn{1}{l}{\boldmath{\textbf{0.00E+00 (0.00E+00)$\approx$}}\unboldmath{}} & \multicolumn{1}{l}{\boldmath{\textbf{0.00E+00 (0.00E+00)$\approx$}}\unboldmath{}} & \multicolumn{1}{l}{\boldmath{\textbf{0.00E+00 (0.00E+00)$\approx$}}\unboldmath{}} & \multicolumn{1}{l}{\boldmath{\textbf{0.00E+00 (0.00E+00)$\approx$}}\unboldmath{}} \\
    $F_{2}$ & \multicolumn{1}{l}{\textbf{0.00E+00 (0.00E+00)}} & \multicolumn{1}{l}{4.70E+01 (1.90E+00)$-$} & \multicolumn{1}{l}{4.91E+01 (1.96E+00)$-$} & \multicolumn{1}{l}{4.68E+01 (1.27E+00)$-$} & \multicolumn{1}{l}{4.91E+01 (4.20E-06)$-$} & \multicolumn{1}{l}{4.55E+01 (1.96E+00)$-$} & \multicolumn{1}{l}{4.50E+01 (2.00E+00)$-$} & \multicolumn{1}{l}{4.47E+01 (1.55E+01)$-$} & \multicolumn{1}{l}{1.26E-02 (3.74E-02)$-$} \\
    $F_{3}$ & \multicolumn{1}{l}{\textbf{0.00E+00 (0.00E+00)}} & \multicolumn{1}{l}{\boldmath{\textbf{0.00E+00 (0.00E+00)$\approx$}}\unboldmath{}} & \multicolumn{1}{l}{\boldmath{\textbf{0.00E+00 (0.00E+00)$\approx$}}\unboldmath{}} & \multicolumn{1}{l}{\boldmath{\textbf{0.00E+00 (0.00E+00)$\approx$}}\unboldmath{}} & \multicolumn{1}{l}{\boldmath{\textbf{0.00E+00 (0.00E+00)$\approx$}}\unboldmath{}} & \multicolumn{1}{l}{\boldmath{\textbf{0.00E+00 (0.00E+00)$\approx$}}\unboldmath{}} & \multicolumn{1}{l}{\boldmath{\textbf{0.00E+00 (0.00E+00)$\approx$}}\unboldmath{}} & \multicolumn{1}{l}{3.04E-05 (5.32E-05)$-$} & \multicolumn{1}{l}{\boldmath{\textbf{0.00E+00 (0.00E+00)$\approx$}}\unboldmath{}} \\
    $F_{4}$ & \multicolumn{1}{l}{1.04E+01 (2.41E+00)} & \multicolumn{1}{l}{2.11E+01 (9.69E+00)$-$} & \multicolumn{1}{l}{8.56E+00 (2.46E+00)$+$} & \multicolumn{1}{l}{7.31E+01 (3.48E+01)$-$} & \multicolumn{1}{l}{9.95E+00 (1.82E+00)$\approx$} & \multicolumn{1}{l}{1.27E+01 (3.17E+00)$-$} & \multicolumn{1}{l}{8.69E+00 (2.34E+00)$+$} & \multicolumn{1}{l}{1.86E+01 (6.63E+00)$-$} & \multicolumn{1}{l}{\boldmath{\textbf{2.02E+00 (8.56E-01)$+$}}\unboldmath{}} \\
    $F_{5}$ & \multicolumn{1}{l}{\textbf{0.00E+00 (0.00E+00)}} & \multicolumn{1}{l}{\boldmath{\textbf{0.00E+00 (0.00E+00)$\approx$}}\unboldmath{}} & \multicolumn{1}{l}{6.87E-01 (7.89E-01)$-$} & \multicolumn{1}{l}{\boldmath{\textbf{0.00E+00 (0.00E+00)$\approx$}}\unboldmath{}} & \multicolumn{1}{l}{\boldmath{\textbf{0.00E+00 (0.00E+00)$\approx$}}\unboldmath{}} & \multicolumn{1}{l}{\boldmath{\textbf{0.00E+00 (0.00E+00)$\approx$}}\unboldmath{}} & \multicolumn{1}{l}{5.97E-03 (2.23E-02)$-$} & \multicolumn{1}{l}{6.57E-02 (1.81E-01)$-$} & \multicolumn{1}{l}{\boldmath{\textbf{0.00E+00 (0.00E+00)$\approx$}}\unboldmath{}} \\
    $F_{6}$ & \multicolumn{1}{l}{3.53E-01 (3.05E-01)} & \multicolumn{1}{l}{7.16E-01 (5.31E-01)$-$} & \multicolumn{1}{l}{4.05E+01 (2.41E+01)$-$} & \multicolumn{1}{l}{6.33E+01 (2.80E+01)$-$} & \multicolumn{1}{l}{1.05E+01 (1.04E+01)$-$} & \multicolumn{1}{l}{5.06E+01 (3.06E+01)$-$} & \multicolumn{1}{l}{5.52E+00 (3.44E+00)$-$} & \multicolumn{1}{l}{3.62E+01 (2.83E+01)$-$} & \multicolumn{1}{l}{\boldmath{\textbf{1.16E-01 (2.79E-02)$+$}}\unboldmath{}} \\
    $F_{7}$ & \multicolumn{1}{l}{1.09E+00 (1.07E+00)} & \multicolumn{1}{l}{3.09E+00 (6.76E+00)$-$} & \multicolumn{1}{l}{1.64E+00 (7.48E-01)$-$} & \multicolumn{1}{l}{4.74E+00 (6.19E+00)$-$} & \multicolumn{1}{l}{3.12E+00 (6.65E+00)$-$} & \multicolumn{1}{l}{7.80E+00 (7.76E+00)$-$} & \multicolumn{1}{l}{3.67E+00 (6.47E+00)$-$} & \multicolumn{1}{l}{2.30E+01 (6.07E+00)$-$} & \multicolumn{1}{l}{\boldmath{\textbf{5.17E-02 (6.21E-02)$+$}}\unboldmath{}} \\
    $F_{8}$ & \multicolumn{1}{l}{7.21E+00 (9.40E+00)} & \multicolumn{1}{l}{7.27E+00 (9.51E+00)$\approx$} & \multicolumn{1}{l}{2.04E+01 (8.57E-02)$-$} & \multicolumn{1}{l}{2.61E+01 (7.25E-01)$-$} & \multicolumn{1}{l}{1.72E+01 (6.83E+00)$-$} & \multicolumn{1}{l}{2.03E+01 (2.94E-01)$-$} & \multicolumn{1}{l}{1.50E+01 (8.79E+00)$-$} & \multicolumn{1}{l}{2.04E+01 (3.26E-01)$-$} & \multicolumn{1}{l}{\boldmath{\textbf{7.19E-01 (1.02E+00)$+$}}\unboldmath{}} \\
    $F_{9}$ & \multicolumn{1}{l}{\textbf{1.00E+02 (0.00E+00)}} & \multicolumn{1}{l}{1.81E+02 (4.27E-06)$-$} & \multicolumn{1}{l}{1.81E+02 (7.79E-06)$-$} & \multicolumn{1}{l}{1.81E+02 (4.47E-06)$-$} & \multicolumn{1}{l}{1.81E+02 (1.10E-05)$-$} & \multicolumn{1}{l}{1.81E+02 (8.82E-06)$-$} & \multicolumn{1}{l}{1.81E+02 (9.73E-06)$-$} & \multicolumn{1}{l}{1.81E+02 (8.14E-07)$-$} & \multicolumn{1}{l}{1.07E+02 (1.98E+01)$-$} \\
    $F_{10}$ & \multicolumn{1}{l}{\textbf{0.00E+00 (0.00E+00)}} & \multicolumn{1}{l}{1.25E+02 (8.32E+01)$-$} & \multicolumn{1}{l}{1.00E+02 (1.68E-02)$-$} & \multicolumn{1}{l}{1.01E+02 (3.81E+01)$-$} & \multicolumn{1}{l}{1.00E+02 (2.09E-02)$-$} & \multicolumn{1}{l}{1.07E+02 (2.97E+01)$-$} & \multicolumn{1}{l}{1.06E+02 (2.89E+01)$-$} & \multicolumn{1}{l}{1.17E+02 (4.28E+01)$-$} & \multicolumn{1}{l}{\boldmath{\textbf{0.00E+00 (0.00E+00)$\approx$}}\unboldmath{}} \\
    $F_{11}$ & \multicolumn{1}{l}{\textbf{1.41E-05 (2.37E-05)}} & \multicolumn{1}{l}{3.47E+02 (4.99E+01)$-$} & \multicolumn{1}{l}{3.13E+02 (3.40E+01)$-$} & \multicolumn{1}{l}{3.13E+02 (3.40E+01)$-$} & \multicolumn{1}{l}{3.27E+02 (4.42E+01)$-$} & \multicolumn{1}{l}{3.27E+02 (4.42E+01)$-$} & \multicolumn{1}{l}{3.27E+02 (4.42E+01)$-$} & \multicolumn{1}{l}{3.09E+02 (2.87E+01)$-$} & \multicolumn{1}{l}{7.28E-05 (3.06E-04)$-$} \\
    $F_{12}$ & \multicolumn{1}{l}{\textbf{2.31E+02 (1.40E+00)}} & \multicolumn{1}{l}{2.37E+02 (3.64E+00)$-$} & \multicolumn{1}{l}{2.42E+02 (6.10E+00)$-$} & \multicolumn{1}{l}{2.35E+02 (4.44E+00)$-$} & \multicolumn{1}{l}{2.34E+02 (3.38E+00)$-$} & \multicolumn{1}{l}{2.37E+02 (3.26E+00)$-$} & \multicolumn{1}{l}{2.40E+02 (4.38E+00)$-$} & \multicolumn{1}{l}{2.44E+02 (4.51E+00)$-$} & \multicolumn{1}{l}{\boldmath{\textbf{2.29E+02 (6.08E-01)$+$}}\unboldmath{}} \\
    \midrule
    $+$ / $\approx$ / $-$ & --    & 0/4/8 & 1/2/9 & 0/3/9 & 0/4/8 & 0/3/9 & 1/2/9 & 0/1/11 & 5/4/3 \\
    \bottomrule
    \end{tabular}%
    }
  \footnotesize
\label{tab:my_detailed_results_20D_wilcoxon_generated_same_FEs}
  }
\end{table}
\clearpage

\subsection{Convergence and Population Analysis}

\begin{figure*}[htpb]
\centering
\includegraphics[width=\columnwidth]{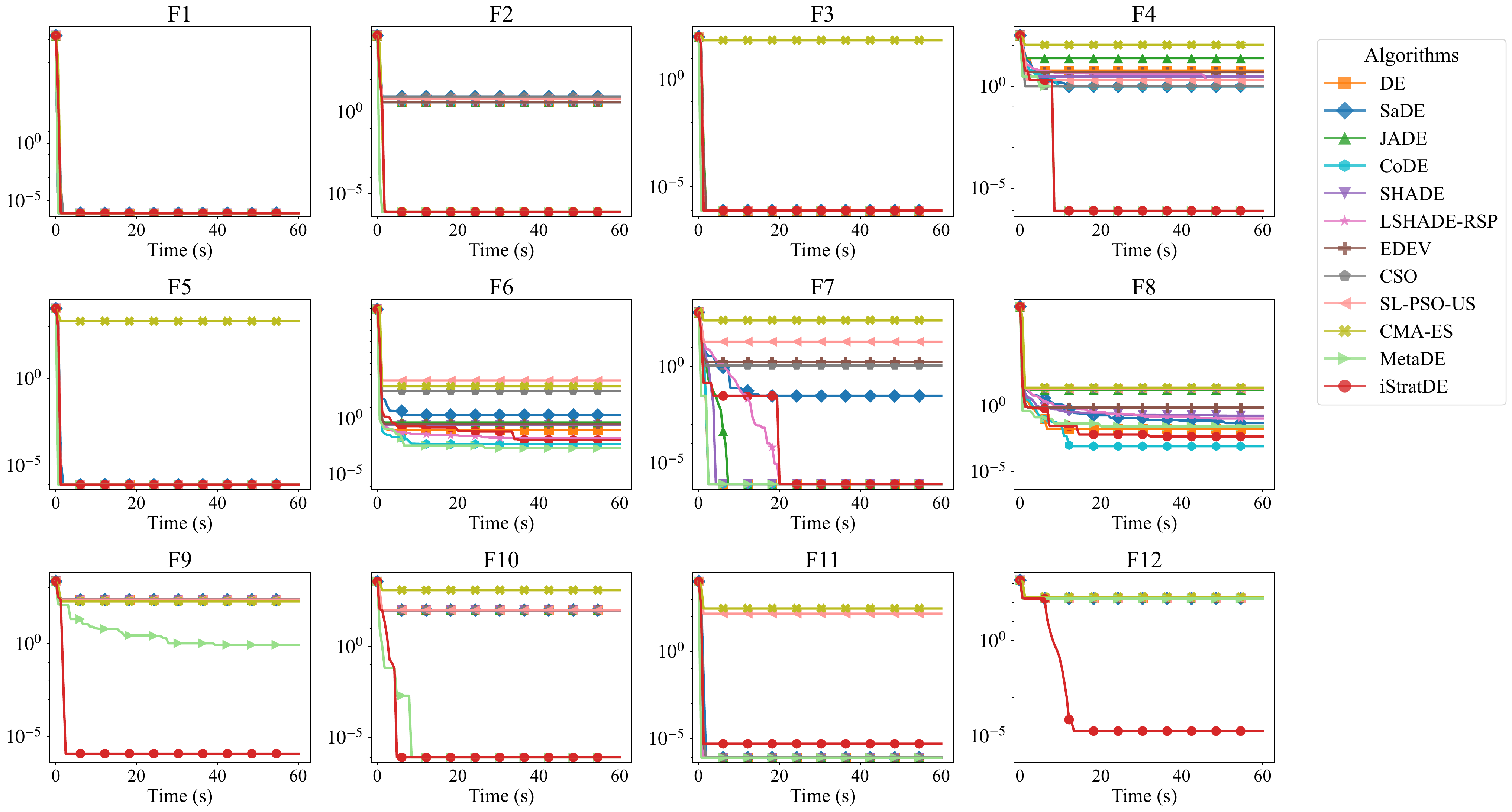}
\caption{Convergence curves on 10D problems in CEC2022 benchmark suite. The peer DE variants are set with population size of 100.}
\label{Figure_convergence_10D_all}
\end{figure*}
\begin{figure*}[htpb]
\centering
\includegraphics[width=\columnwidth]{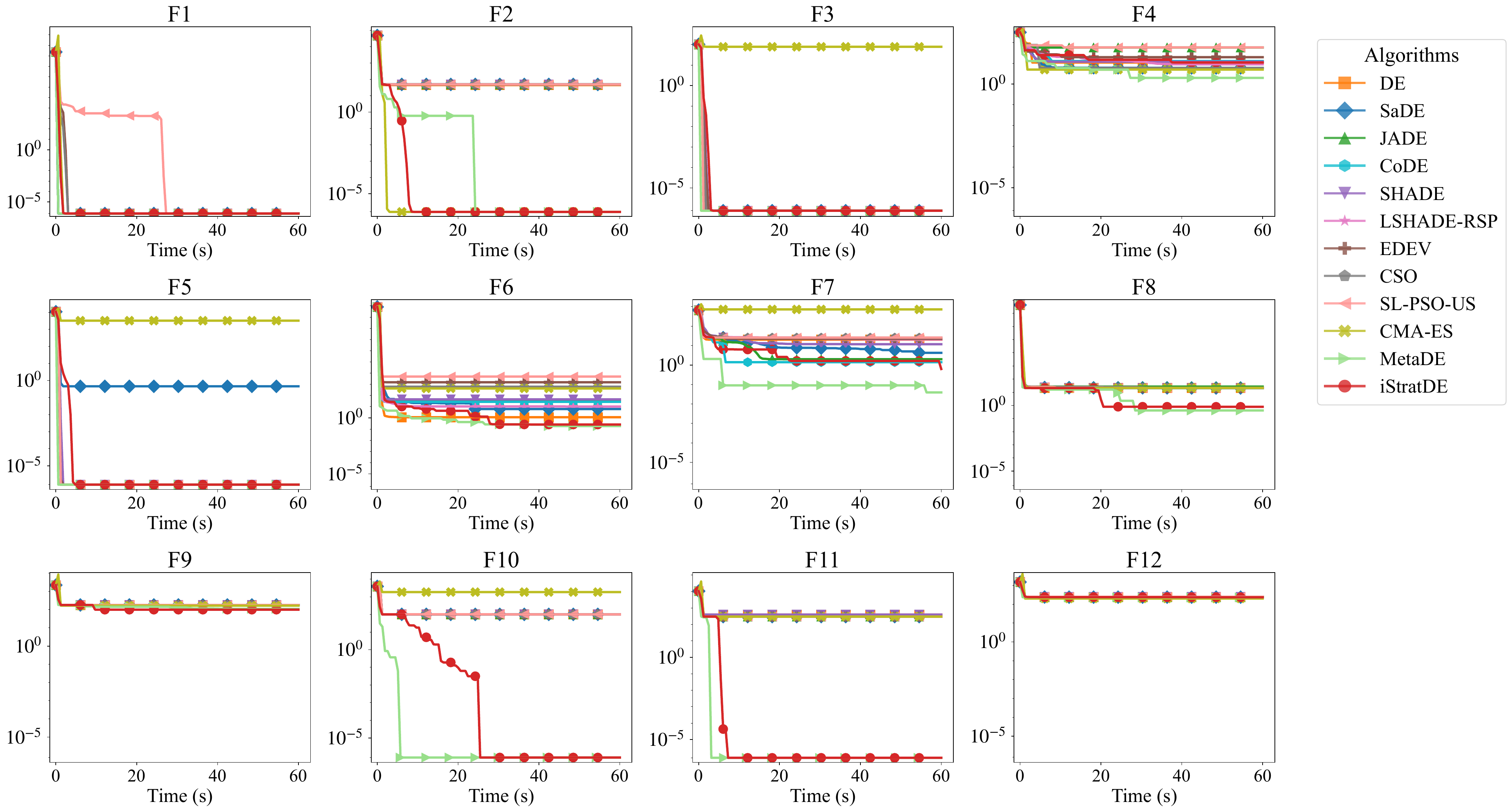}
\caption{Convergence curves on 20D problems in CEC2022 benchmark suite. The peer DE variants are set with population size of 100.}
\label{Figure_convergence_20D_all}
\end{figure*}
\begin{figure*}[htpb]
\centering
\includegraphics[width=0.9\columnwidth]{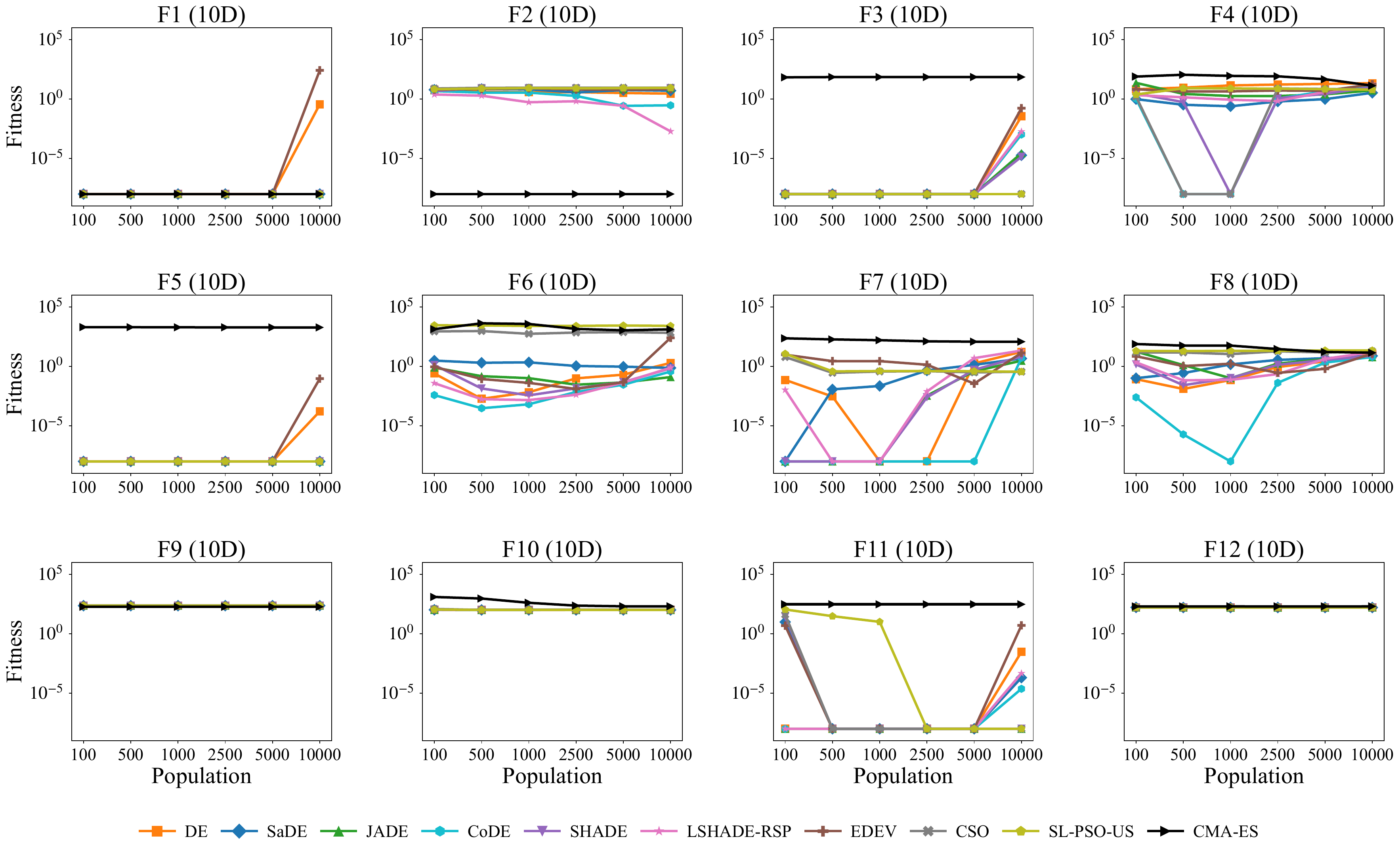}
\caption{The performance of the baseline algorithms on 10D problems in CEC2022 benchmark suite as the population changes.}
\label{Figure_baseline_pop_change_10D_all}
\end{figure*}
\begin{figure*}[htpb]
\centering
\includegraphics[width=0.9\columnwidth]{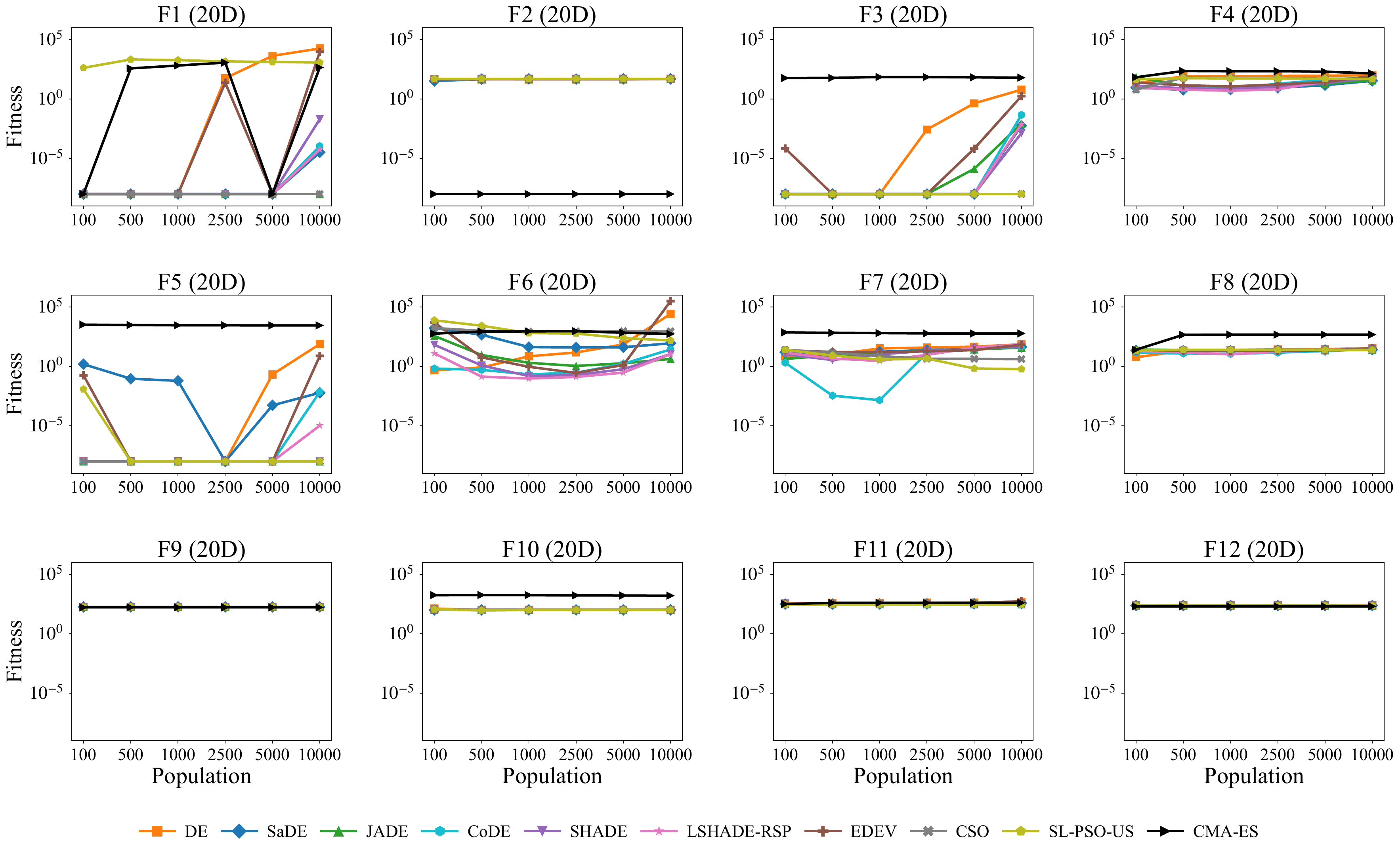}
\caption{The performance of the baseline algorithms on 20D problems in CEC2022 benchmark suite as the population changes.}
\label{Figure_baseline_pop_change_20D_all}
\end{figure*}
\clearpage

\section{Algorithmic Component Analysis}

\subsection{Strategy Diversity \& Utilization}

To evaluate strategy dominance, we analyzed the performance of iStratDE on the CEC2022 benchmark suite across 10 and 20 dimensions by comparing the frequency of successful updates with the composition of the top 10\% elite individuals. 
The results indicate that no single strategy universally dominates the evolutionary process. 
Specifically, strategies based on the \textit{Current} vector achieve high success rates through local search mechanisms, whereas the \textit{Best} and \textit{PBest} strategies predominate within the elite population, accounting for over 60\% of the composition and thereby driving global convergence. 
These findings confirm that fixed strategy approaches are insufficient, as they are prone to stagnation or suffer from low success rates. 
In contrast, the adaptive mechanism of iStratDE effectively leverages the synergy between these complementary strategies to balance exploration and exploitation.

\begin{figure}[htpb]
    \centering
    \includegraphics[width=\textwidth]{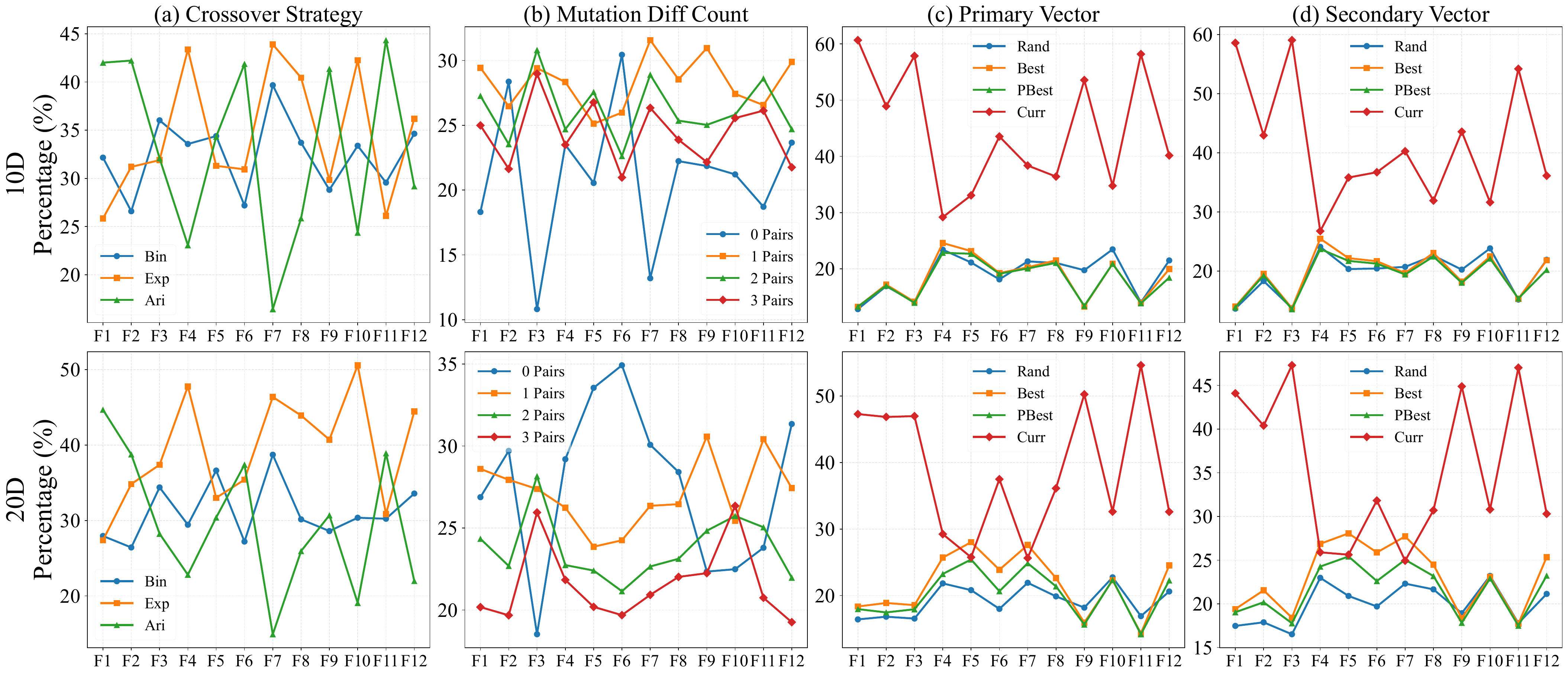}
    \caption{Proportion of successful updates contributed by different adaptive strategies across the CEC2022 benchmark suite. The plots illustrate the relative effectiveness of operator variants in generating offspring with better fitness than their parents. The comparison between 10-dimensional (top row) and 20-dimensional (bottom row) problems highlights how strategy performance varies with dimensionality. Subplots display the success contributions of: (a) Crossover methods, (b) Number of difference vectors in mutation, (c) Selection of the primary vector, and (d) Selection of the secondary vector.}
    \label{fig:success}
\end{figure}

\begin{figure}[htpb]
    \centering
    \includegraphics[width=\textwidth]{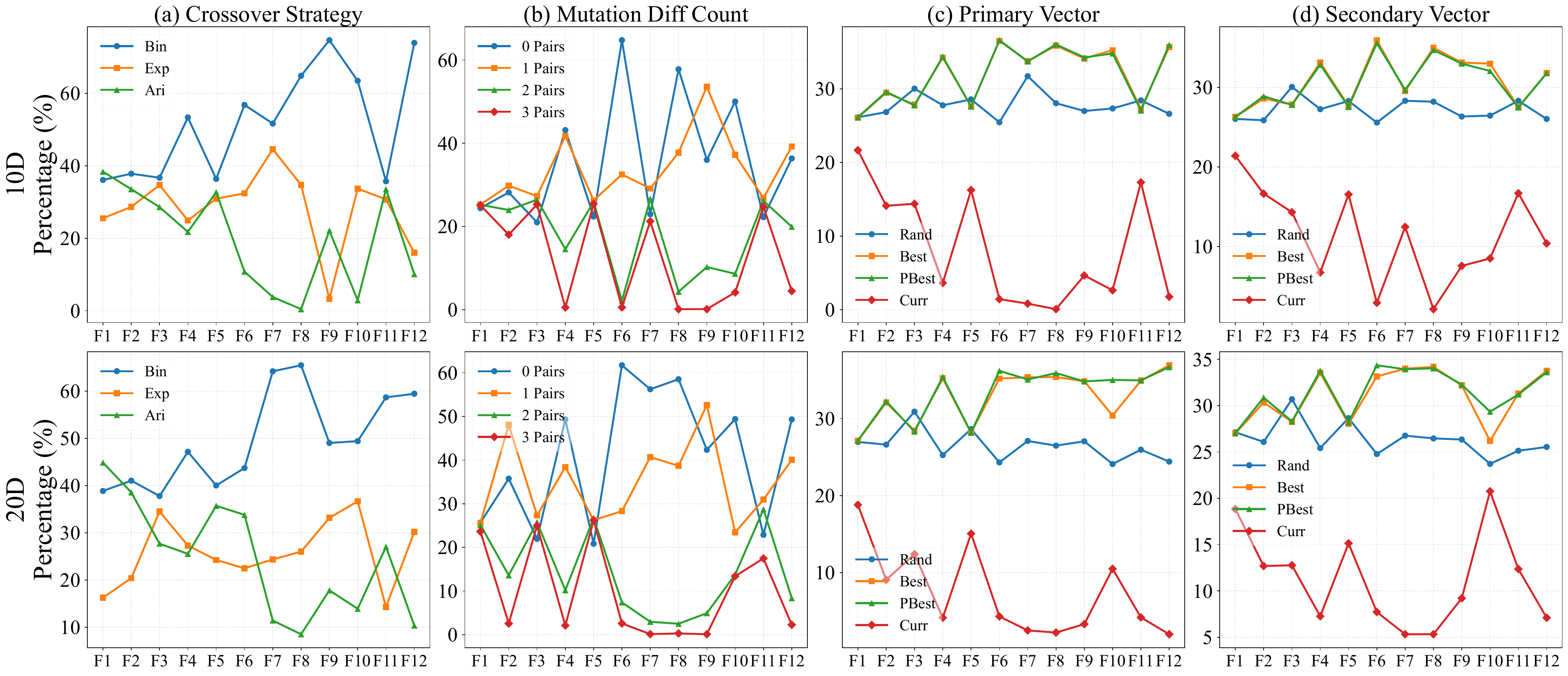}
    \caption{Strategy distribution within the top 10\% elite population. This figure analyzes the composition of strategies associated with the highest-quality solutions found by the algorithm. It reveals the correlation between specific operator choices and elite performance on 10D (top row) and 20D (bottom row) landscapes. The panels correspond to: (a) Crossover strategy, (b) Mutation difference count, (c) Primary vector selection, and (d) Secondary vector selection.}
    \label{fig:elite}
\end{figure}

\clearpage

\subsection{Ablation Studies}

The specific fixed probability distributions employed in \texttt{iStratDE(fix)} are defined based on an empirical analysis of the top 10\% of elite individuals, as follows:
\begin{equation}
    \begin{aligned}
        P_{\text{vector}} &= \{ \text{Rand}: 0.26, \ \text{Best}: 0.32, \ \text{PBest}: 0.32, \ \text{Current}: 0.10 \} \\
        P_{\text{diff\_pairs}} &= \{ 0\text{ extra}: 0.40, \ 1\text{ Extra}: 0.30, \ 2\text{ Extra}: 0.20, \ 3\text{ Extra}: 0.10 \} \\
        P_{\text{crossover}} &= \{ \text{Binomial}: 0.60, \ \text{Exponential}: 0.25, \ \text{Arithmetic}: 0.15 \}
    \end{aligned}
\end{equation}

\begin{table}[htbp]
  \centering
  \scriptsize
    \caption{Comparison of iStratDE and its variants on the 10-dimensional (10D) CEC 2022 benchmark suite. For all the experimental algorithms, the population size was set to 100,000, and all were run within a unified 60-second time limit. \texttt{iStratDE (fix)} employs a fixed strategy allocation based on the strategy proportions observed in the top 10\% of individuals from the original iStratDE's final population. \texttt{iStratDE (reallocate)} involves randomly re-assigning strategies to the population every 1000 generations. Results are reported as mean (standard deviation). The symbols $+$, $-$, and $\approx$ indicate that the result is significantly better, significantly worse, or statistically similar to the original iStratDE, respectively. The best mean values in each row are highlighted.}
  {
    \renewcommand{\arraystretch}{1.2}
    \begin{tabular}{cccc}
    \toprule
    Func & iStratDE & iStratDE (fix) & iStratDE (reallocate) \\
    \midrule
    $F_{1}$ & \multicolumn{1}{l}{\boldmath{\textbf{0.00E+00 (0.00E+00)}}\unboldmath{}} & \multicolumn{1}{l}{\boldmath{\textbf{0.00E+00 (0.00E+00)$\approx$}}\unboldmath{}} & \multicolumn{1}{l}{\boldmath{\textbf{0.00E+00 (0.00E+00)$\approx$}}\unboldmath{}} \\
    $F_{2}$ & \multicolumn{1}{l}{\boldmath{\textbf{0.00E+00 (0.00E+00)}}\unboldmath{}} & \multicolumn{1}{l}{\boldmath{\textbf{0.00E+00 (0.00E+00)$\approx$}}\unboldmath{}} & \multicolumn{1}{l}{\boldmath{\textbf{0.00E+00 (0.00E+00)$\approx$}}\unboldmath{}} \\
    $F_{3}$ & \multicolumn{1}{l}{\boldmath{\textbf{0.00E+00 (0.00E+00)}}\unboldmath{}} & \multicolumn{1}{l}{\boldmath{\textbf{0.00E+00 (0.00E+00)$\approx$}}\unboldmath{}} & \multicolumn{1}{l}{\boldmath{\textbf{0.00E+00 (0.00E+00)$\approx$}}\unboldmath{}} \\
    $F_{4}$ & \multicolumn{1}{l}{\boldmath{\textbf{0.00E+00 (0.00E+00)}}\unboldmath{}} & \multicolumn{1}{l}{1.55E-01 (3.61E-01)$-$} & \multicolumn{1}{l}{\boldmath{\textbf{0.00E+00 (0.00E+00)$\approx$}}\unboldmath{}} \\
    $F_{5}$ & \multicolumn{1}{l}{\boldmath{\textbf{0.00E+00 (0.00E+00)}}\unboldmath{}} & \multicolumn{1}{l}{\boldmath{\textbf{0.00E+00 (0.00E+00)$\approx$}}\unboldmath{}} & \multicolumn{1}{l}{\boldmath{\textbf{0.00E+00 (0.00E+00)$\approx$}}\unboldmath{}} \\
    $F_{6}$ & \multicolumn{1}{l}{6.44E-03 (3.72E-03)} & \multicolumn{1}{l}{1.70E-02 (1.58E-02)$-$} & \multicolumn{1}{l}{\boldmath{\textbf{2.75E-05 (9.20E-05)$+$}}\unboldmath{}} \\
    $F_{7}$ & \multicolumn{1}{l}{\boldmath{\textbf{0.00E+00 (0.00E+00)}}\unboldmath{}} & \multicolumn{1}{l}{2.04E-02 (1.78E-02)$-$} & \multicolumn{1}{l}{9.31E-04 (5.19E-03)$-$} \\
    $F_{8}$ & \multicolumn{1}{l}{\boldmath{\textbf{1.10E-03 (2.98E-03)}}\unboldmath{}} & \multicolumn{1}{l}{1.80E-01 (2.72E-01)$-$} & \multicolumn{1}{l}{2.13E-03 (2.63E-03)$\approx$} \\
    $F_{9}$ & \multicolumn{1}{l}{\boldmath{\textbf{1.78E-06 (2.19E-07)}}\unboldmath{}} & \multicolumn{1}{l}{1.22E+02 (1.04E+02)$-$} & \multicolumn{1}{l}{3.13E+00 (1.74E+01)$-$} \\
    $F_{10}$ & \multicolumn{1}{l}{\boldmath{\textbf{0.00E+00 (0.00E+00)}}\unboldmath{}} & \multicolumn{1}{l}{\boldmath{\textbf{0.00E+00 (0.00E+00)$\approx$}}\unboldmath{}} & \multicolumn{1}{l}{\boldmath{\textbf{0.00E+00 (0.00E+00)$\approx$}}\unboldmath{}} \\
    $F_{11}$ & \multicolumn{1}{l}{1.05E-05 (1.29E-05)} & \multicolumn{1}{l}{\boldmath{\textbf{0.00E+00 (0.00E+00)$+$}}\unboldmath{}} & \multicolumn{1}{l}{\boldmath{\textbf{0.00E+00 (0.00E+00)$+$}}\unboldmath{}} \\
    $F_{12}$ & \multicolumn{1}{l}{\boldmath{\textbf{6.11E+01 (7.60E+01)}}\unboldmath{}} & \multicolumn{1}{l}{1.49E+02 (3.84E+01)$-$} & \multicolumn{1}{l}{9.42E+01 (7.79E+01)$\approx$} \\
    \midrule
    $+$ / $\approx$ / $-$ & --    & 1/5/6 & 2/8/2 \\
    \bottomrule
    \end{tabular}
  \label{tab:istratde_variants_comparison}
  }
\end{table}
\begin{table}[htbp]
  \centering
  \scriptsize
    \caption{Comparison of iStratDE and its variants on the 10-dimensional (20D) CEC 2022 benchmark suite. For all the experimental algorithms, the population size was set to 100,000, and all were run within a unified 60-second time limit. \texttt{iStratDE (fix)} employs a fixed strategy allocation based on the strategy proportions observed in the top 10\% of individuals from the original iStratDE's final population. \texttt{iStratDE (reallocate)} involves randomly re-assigning strategies to the population every 1000 generations. Results are reported as mean (standard deviation). The symbols $+$, $-$, and $\approx$ indicate that the result is significantly better, significantly worse, or statistically similar to the original iStratDE, respectively. The best mean values in each row are highlighted.}
  {
    \renewcommand{\arraystretch}{1.2}
    \begin{tabular}{cccc}
    \toprule
    Func & iStratDE & iStratDE (fix) & iStratDE (reallocate) \\
    \midrule
    $F_{1}$ & \multicolumn{1}{l}{\boldmath{\textbf{0.00E+00 (0.00E+00)}}\unboldmath{}} & \multicolumn{1}{l}{\boldmath{\textbf{0.00E+00 (0.00E+00)$\approx$}}\unboldmath{}} & \multicolumn{1}{l}{\boldmath{\textbf{0.00E+00 (0.00E+00)$\approx$}}\unboldmath{}} \\
    $F_{2}$ & \multicolumn{1}{l}{\boldmath{\textbf{0.00E+00 (0.00E+00)}}\unboldmath{}} & \multicolumn{1}{l}{3.13E+01 (1.96E+01)$-$} & \multicolumn{1}{l}{\boldmath{\textbf{0.00E+00 (0.00E+00)$\approx$}}\unboldmath{}} \\
    $F_{3}$ & \multicolumn{1}{l}{\boldmath{\textbf{0.00E+00 (0.00E+00)}}\unboldmath{}} & \multicolumn{1}{l}{\boldmath{\textbf{0.00E+00 (0.00E+00)$\approx$}}\unboldmath{}} & \multicolumn{1}{l}{\boldmath{\textbf{0.00E+00 (0.00E+00)$\approx$}}\unboldmath{}} \\
    $F_{4}$ & \multicolumn{1}{l}{9.62E+00 (2.59E+00)} & \multicolumn{1}{l}{8.40E+00 (2.15E+00)$\approx$} & \multicolumn{1}{l}{\boldmath{\textbf{1.24E+00 (8.62E-01)$+$}}\unboldmath{}} \\
    $F_{5}$ & \multicolumn{1}{l}{\boldmath{\textbf{0.00E+00 (0.00E+00)}}\unboldmath{}} & \multicolumn{1}{l}{\boldmath{\textbf{0.00E+00 (0.00E+00)$\approx$}}\unboldmath{}} & \multicolumn{1}{l}{\boldmath{\textbf{0.00E+00 (0.00E+00)$\approx$}}\unboldmath{}} \\
    $F_{6}$ & \multicolumn{1}{l}{\boldmath{\textbf{3.97E-01 (3.00E-01)}}\unboldmath{}} & \multicolumn{1}{l}{2.01E+00 (9.47E-01)$-$} & \multicolumn{1}{l}{7.91E-01 (8.41E-01)$\approx$} \\
    $F_{7}$ & \multicolumn{1}{l}{1.60E+00 (2.59E+00)} & \multicolumn{1}{l}{5.34E+00 (7.88E+00)$-$} & \multicolumn{1}{l}{\boldmath{\textbf{8.41E-01 (3.46E+00)$\approx$}}\unboldmath{}} \\
    $F_{8}$ & \multicolumn{1}{l}{6.18E+00 (8.96E+00)} & \multicolumn{1}{l}{6.00E+00 (8.89E+00)$\approx$} & \multicolumn{1}{l}{\boldmath{\textbf{1.57E+00 (4.80E+00)$\approx$}}\unboldmath{}} \\
    $F_{9}$ & \multicolumn{1}{l}{\boldmath{\textbf{1.00E+02 (0.00E+00)}}\unboldmath{}} & \multicolumn{1}{l}{1.81E+02 (4.00E-05)$-$} & \multicolumn{1}{l}{1.71E+02 (2.67E+01)$-$} \\
    $F_{10}$ & \multicolumn{1}{l}{\boldmath{\textbf{0.00E+00 (0.00E+00)}}\unboldmath{}} & \multicolumn{1}{l}{9.39E+01 (2.40E+01)$-$} & \multicolumn{1}{l}{\boldmath{\textbf{0.00E+00 (0.00E+00)$\approx$}}\unboldmath{}} \\
    $F_{11}$ & \multicolumn{1}{l}{9.52E+00 (5.26E+01)} & \multicolumn{1}{l}{2.16E+02 (1.82E+02)$-$} & \multicolumn{1}{l}{\boldmath{\textbf{0.00E+00 (0.00E+00)$+$}}\unboldmath{}} \\
    $F_{12}$ & \multicolumn{1}{l}{\boldmath{\textbf{2.31E+02 (1.59E+00)}}\unboldmath{}} & \multicolumn{1}{l}{2.34E+02 (3.21E+00)$-$} & \multicolumn{1}{l}{2.32E+02 (1.89E+00)$\approx$} \\
    \midrule
    $+$ / $\approx$ / $-$ & --    & 0/5/7 & 2/9/1 \\
    \bottomrule
    \end{tabular}

  \label{tab:istratde_variants_comparison_20D}
  }
\end{table}
\clearpage
\begin{table}[htbp]
  \centering
  \caption{Comparison of iStratDE and its ablation variants on the 10-dimensional (10D) CEC 2022 benchmark suite. All algorithms were configured with a population size of 100,000 and a 60-second time limit. \texttt{No(Rand/Curr)} represents the variant where Rand and Current strategies are removed (retaining Best+PBest). \texttt{No(Best/PBest)} removes Best and PBest strategies (retaining Rand+Curr). The last three columns represent variants where a specific crossover strategy (Arithmetic, Binomial, or Exponential) is removed. The symbols $+$, $-$, and $\approx$ indicate that the result is significantly better, significantly worse, or statistically similar to the original iStratDE, respectively. The best mean values in each row are highlighted.}
  \label{tab:istratde_ablation_final}
  {
  \resizebox{\textwidth}{!}{
    \renewcommand{\arraystretch}{1.2}
    \begin{tabular}{ccccccc}
    \toprule
    Func & iStratDE & No(Rand/Curr) & No(Best/PBest) & No(Arith) & No(Bin) & No(Exp) \\
         & (Original) & (Best+PBest) & (Rand+Curr) & (Bin+Exp) & (Arith+Exp) & (Arith+Bin) \\
    \midrule
    $F_{1}$ & \boldmath{\textbf{0.00E+00 (0.00E+00)}}\unboldmath{} & \boldmath{\textbf{0.00E+00 (0.00E+00)}}\unboldmath{}$\approx$ & \boldmath{\textbf{0.00E+00 (0.00E+00)}}\unboldmath{}$\approx$ & \boldmath{\textbf{0.00E+00 (0.00E+00)}}\unboldmath{}$\approx$ & \boldmath{\textbf{0.00E+00 (0.00E+00)}}\unboldmath{}$\approx$ & \boldmath{\textbf{0.00E+00 (0.00E+00)}}\unboldmath{}$\approx$ \\
    $F_{2}$ & \boldmath{\textbf{0.00E+00 (0.00E+00)}}\unboldmath{} & \boldmath{\textbf{0.00E+00 (0.00E+00)}}\unboldmath{}$\approx$ & \boldmath{\textbf{0.00E+00 (0.00E+00)}}\unboldmath{}$\approx$ & \boldmath{\textbf{0.00E+00 (0.00E+00)}}\unboldmath{}$\approx$ & \boldmath{\textbf{0.00E+00 (0.00E+00)}}\unboldmath{}$\approx$ & \boldmath{\textbf{0.00E+00 (0.00E+00)}}\unboldmath{}$\approx$ \\
    $F_{3}$ & \boldmath{\textbf{0.00E+00 (0.00E+00)}}\unboldmath{} & \boldmath{\textbf{0.00E+00 (0.00E+00)}}\unboldmath{}$\approx$ & \boldmath{\textbf{0.00E+00 (0.00E+00)}}\unboldmath{}$\approx$ & \boldmath{\textbf{0.00E+00 (0.00E+00)}}\unboldmath{}$\approx$ & \boldmath{\textbf{0.00E+00 (0.00E+00)}}\unboldmath{}$\approx$ & \boldmath{\textbf{0.00E+00 (0.00E+00)}}\unboldmath{}$\approx$ \\
    $F_{4}$ & \boldmath{\textbf{0.00E+00 (0.00E+00)}}\unboldmath{} & \boldmath{\textbf{0.00E+00 (0.00E+00)}}\unboldmath{}$\approx$ & \boldmath{\textbf{0.00E+00 (0.00E+00)}}\unboldmath{}$\approx$ & 3.11E-02 (1.73E-01)$-$ & \boldmath{\textbf{0.00E+00 (0.00E+00)}}\unboldmath{}$\approx$ & \boldmath{\textbf{0.00E+00 (0.00E+00)}}\unboldmath{}$\approx$ \\
    $F_{5}$ & \boldmath{\textbf{0.00E+00 (0.00E+00)}}\unboldmath{} & \boldmath{\textbf{0.00E+00 (0.00E+00)}}\unboldmath{}$\approx$ & \boldmath{\textbf{0.00E+00 (0.00E+00)}}\unboldmath{}$\approx$ & \boldmath{\textbf{0.00E+00 (0.00E+00)}}\unboldmath{}$\approx$ & \boldmath{\textbf{0.00E+00 (0.00E+00)}}\unboldmath{}$\approx$ & \boldmath{\textbf{0.00E+00 (0.00E+00)}}\unboldmath{}$\approx$ \\
    $F_{6}$ & \boldmath{\textbf{6.44E-03 (3.72E-03)}}\unboldmath{} & 1.31E-02 (1.29E-02)$\approx$ & 2.22E-02 (5.72E-03)$-$ & 1.72E-02 (9.92E-03)$-$ & 2.15E-02 (2.26E-02)$-$ & 9.83E-03 (6.39E-03)$\approx$ \\
    $F_{7}$ & \boldmath{\textbf{0.00E+00 (0.00E+00)}}\unboldmath{} & 1.46E-02 (1.73E-02)$-$ & \boldmath{\textbf{0.00E+00 (0.00E+00)}}\unboldmath{}$\approx$ & \boldmath{\textbf{0.00E+00 (0.00E+00)}}\unboldmath{}$\approx$ & \boldmath{\textbf{0.00E+00 (0.00E+00)}}\unboldmath{}$\approx$ & \boldmath{\textbf{0.00E+00 (0.00E+00)}}\unboldmath{}$\approx$ \\
    $F_{8}$ & \boldmath{\textbf{1.10E-03 (2.98E-03)}}\unboldmath{} & 1.85E-02 (1.30E-02)$-$ & 5.32E-03 (6.83E-03)$-$ & 1.85E-03 (2.47E-03)$\approx$ & 5.82E-02 (3.64E-02)$-$ & 5.02E-03 (6.20E-03)$-$ \\
    $F_{9}$ & 1.78E-06 (2.19E-07) & 2.29E+02 (1.38E-05)$-$ & 2.41E-06 (2.43E-06)$\approx$ & 2.68E-06 (3.61E-06)$\approx$ & \boldmath{\textbf{1.75E-06 (0.00E+00)}}\unboldmath{}$\approx$ & 2.68E-06 (5.18E-06)$\approx$ \\
    $F_{10}$ & \boldmath{\textbf{0.00E+00 (0.00E+00)}}\unboldmath{} & \boldmath{\textbf{0.00E+00 (0.00E+00)}}\unboldmath{}$\approx$ & \boldmath{\textbf{0.00E+00 (0.00E+00)}}\unboldmath{}$\approx$ & \boldmath{\textbf{0.00E+00 (0.00E+00)}}\unboldmath{}$\approx$ & \boldmath{\textbf{0.00E+00 (0.00E+00)}}\unboldmath{}$\approx$ & \boldmath{\textbf{0.00E+00 (0.00E+00)}}\unboldmath{}$\approx$ \\
    $F_{11}$ & 1.05E-05 (1.29E-05) & 5.99E-06 (9.96E-06)$\approx$ & \boldmath{\textbf{0.00E+00 (0.00E+00)}}\unboldmath{}$+$ & 6.99E-06 (8.28E-06)$\approx$ & 1.35E-05 (1.65E-05)$\approx$ & 1.02E-05 (1.26E-05)$\approx$ \\
    $F_{12}$ & 6.11E+01 (7.60E+01) & 1.59E+02 (1.39E-05)$-$ & \boldmath{\textbf{1.12E+01 (3.76E+01)}}\unboldmath{}$+$ & 7.44E+01 (7.92E+01)$\approx$ & 5.46E+01 (7.23E+01)$\approx$ & 9.42E+01 (7.79E+01)$\approx$ \\
    \midrule
    $+$ / $\approx$ / $-$ & -- & 0/7/5 & 2/8/2 & 0/10/2 & 0/10/2 & 0/11/1 \\
    \bottomrule
    \end{tabular}
  }
}
\end{table}
\begin{table}[htbp]
  \centering
  \caption{Comparison of iStratDE and its ablation variants on the 20-dimensional (20D) CEC 2022 benchmark suite. All algorithms were configured with a population size of 100,000 and a 60-second time limit. \texttt{No(Rand/Curr)} represents the variant where Rand and Current strategies are removed. \texttt{No(Best/PBest)} removes Best and PBest strategies. The last three columns represent variants where a specific crossover strategy is removed. The symbols $+$, $-$, and $\approx$ indicate that the result is significantly better, significantly worse, or statistically similar to the original iStratDE, respectively. The best mean values in each row are highlighted.}
  \label{tab:istratde_ablation_20d}
  {
  \resizebox{\textwidth}{!}{
    \renewcommand{\arraystretch}{1.2}
    \begin{tabular}{ccccccc}
    \toprule
    Func & iStratDE & No(Rand/Curr) & No(Best/PBest) & No(Arith) & No(Bin) & No(Exp) \\
         & (Original) & (Best+PBest) & (Rand+Curr) & (Bin+Exp) & (Arith+Exp) & (Arith+Bin) \\
    \midrule
    $F_{1}$ & \boldmath{\textbf{0.00E+00 (0.00E+00)}}\unboldmath{} & \boldmath{\textbf{0.00E+00 (0.00E+00)}}\unboldmath{}$\approx$ & \boldmath{\textbf{0.00E+00 (0.00E+00)}}\unboldmath{}$\approx$ & \boldmath{\textbf{0.00E+00 (0.00E+00)}}\unboldmath{}$\approx$ & \boldmath{\textbf{0.00E+00 (0.00E+00)}}\unboldmath{}$\approx$ & \boldmath{\textbf{0.00E+00 (0.00E+00)}}\unboldmath{}$\approx$ \\
    $F_{2}$ & \boldmath{\textbf{0.00E+00 (0.00E+00)}}\unboldmath{} & 2.26E+01 (2.23E+01)$-$ & 2.08E-02 (1.94E-02)$-$ & \boldmath{\textbf{0.00E+00 (0.00E+00)}}\unboldmath{}$\approx$ & \boldmath{\textbf{0.00E+00 (0.00E+00)}}\unboldmath{}$\approx$ & \boldmath{\textbf{0.00E+00 (0.00E+00)}}\unboldmath{}$\approx$ \\
    $F_{3}$ & \boldmath{\textbf{0.00E+00 (0.00E+00)}}\unboldmath{} & \boldmath{\textbf{0.00E+00 (0.00E+00)}}\unboldmath{}$\approx$ & \boldmath{\textbf{0.00E+00 (0.00E+00)}}\unboldmath{}$\approx$ & \boldmath{\textbf{0.00E+00 (0.00E+00)}}\unboldmath{}$\approx$ & \boldmath{\textbf{0.00E+00 (0.00E+00)}}\unboldmath{}$\approx$ & \boldmath{\textbf{0.00E+00 (0.00E+00)}}\unboldmath{}$\approx$ \\
    $F_{4}$ & 9.62E+00 (2.59E+00) & 1.14E+01 (3.07E+00)$-$ & \boldmath{\textbf{6.20E+00 (1.52E+00)}}\unboldmath{}$+$ & 9.42E+00 (2.58E+00)$\approx$ & 1.05E+01 (2.09E+00)$\approx$ & 1.43E+01 (4.42E+00)$-$ \\
    $F_{5}$ & \boldmath{\textbf{0.00E+00 (0.00E+00)}}\unboldmath{} & \boldmath{\textbf{0.00E+00 (0.00E+00)}}\unboldmath{}$\approx$ & \boldmath{\textbf{0.00E+00 (0.00E+00)}}\unboldmath{}$\approx$ & \boldmath{\textbf{0.00E+00 (0.00E+00)}}\unboldmath{}$\approx$ & \boldmath{\textbf{0.00E+00 (0.00E+00)}}\unboldmath{}$\approx$ & \boldmath{\textbf{0.00E+00 (0.00E+00)}}\unboldmath{}$\approx$ \\
    $F_{6}$ & \boldmath{\textbf{3.97E-01 (3.00E-01)}}\unboldmath{} & 1.77E+00 (1.11E+00)$-$ & 9.00E-01 (5.50E-01)$-$ & 4.19E+00 (4.14E+00)$-$ & 1.70E+00 (8.62E-01)$-$ & 2.06E+00 (1.06E+00)$-$ \\
    $F_{7}$ & 1.60E+00 (2.59E+00) & \boldmath{\textbf{6.89E-01 (6.38E-01)}}\unboldmath{}$+$ & 1.42E+00 (8.69E-01)$\approx$ & 8.40E-01 (6.64E-01)$+$ & 2.65E+00 (1.50E+00)$\approx$ & 7.75E-01 (7.88E-01)$+$ \\
    $F_{8}$ & 6.18E+00 (8.96E+00) & 1.04E+01 (9.93E+00)$\approx$ & 5.02E+00 (2.59E+00)$\approx$ & 4.89E+00 (8.22E+00)$\approx$ & 5.07E+00 (8.19E+00)$\approx$ & \boldmath{\textbf{2.95E+00 (6.61E+00)}}\unboldmath{}$\approx$ \\
    $F_{9}$ & \boldmath{\textbf{1.00E+02 (0.00E+00)}}\unboldmath{} & 1.81E+02 (3.08E-05)$-$ & \boldmath{\textbf{1.00E+02 (2.07E-04)}}\unboldmath{}$\approx$ & \boldmath{\textbf{1.00E+02 (0.00E+00)}}\unboldmath{}$\approx$ & \boldmath{\textbf{1.00E+02 (0.00E+00)}}\unboldmath{}$\approx$ & \boldmath{\textbf{1.00E+02 (0.00E+00)}}\unboldmath{}$\approx$ \\
    $F_{10}$ & \boldmath{\textbf{0.00E+00 (0.00E+00)}}\unboldmath{} & 1.10E+01 (2.87E+01)$-$ & \boldmath{\textbf{0.00E+00 (0.00E+00)}}\unboldmath{}$\approx$ & \boldmath{\textbf{0.00E+00 (0.00E+00)}}\unboldmath{}$\approx$ & \boldmath{\textbf{0.00E+00 (0.00E+00)}}\unboldmath{}$\approx$ & 3.52E-02 (5.11E-02)$-$ \\
    $F_{11}$ & 9.52E+00 (5.26E+01) & 1.97E+02 (1.42E+02)$-$ & 9.17E-03 (5.47E-03)$+$ & \boldmath{\textbf{1.01E-06 (5.64E-06)}}\unboldmath{}$+$ & 5.06E-06 (1.76E-05)$+$ & 1.41E+02 (1.62E+02)$-$ \\
    $F_{12}$ & \boldmath{\textbf{2.31E+02 (1.59E+00)}}\unboldmath{} & 2.33E+02 (1.81E+00)$-$ & 2.32E+02 (8.12E-01)$\approx$ & \boldmath{\textbf{2.31E+02 (1.92E+00)}}\unboldmath{}$\approx$ & \boldmath{\textbf{2.31E+02 (1.71E+00)}}\unboldmath{}$\approx$ & 2.32E+02 (1.50E+00)$\approx$ \\
    \midrule
    $+$ / $\approx$ / $-$ & -- & 1/4/7 & 2/8/2 & 2/9/1 & 1/10/1 & 1/6/5 \\
    \bottomrule
    \end{tabular}
  }
}
\end{table}
\clearpage

\subsection{Sensitivity \& Sampling Analysis}
Tables \ref{tab:poolsize_sensitivity_reversed} and \ref{tab:poolsize_sensitivity_20d} present the detailed results of the Diversity Sensitivity Analysis conducted on the 10-dimensional and 20-dimensional CEC 2022 benchmark suites, respectively. The primary objective of this analysis is to empirically verify the correlation between the richness of the strategy pool and the algorithm's optimization capability. In this experiment, the \texttt{PoolSize} parameter serves as a quantifiable proxy for the degree of available strategy diversity. The fully randomized pool (\texttt{PoolSize} = 192), which represents the standard configuration of iStratDE, is established as the baseline for statistical comparison.

The results reveal a sharp contrast in performance across different diversity levels. As the \texttt{PoolSize} decreases, the algorithm's optimization capability deteriorates significantly. This trend is most pronounced in the extreme case of a single-strategy configuration, where the lack of structural variation leads to severe stagnation on complex multimodal functions. Statistical tests consistently confirm that reducing the strategy pool size results in significantly worse outcomes compared to the baseline. These findings provide strong empirical evidence that the large-scale sampling mechanism is not merely an initialization step, but a pivotal component that ensures the population possesses sufficient structural variety to escape local optima and locate global solutions.
\begin{table}[htbp]
  \centering
  \scriptsize
  \caption{Diversity Sensitivity Analysis on the 10-dimensional (10D) CEC 2022 benchmark suite with varying Pool Sizes (Reverse Order). The population size was set to 100,000. \texttt{PoolSize = 192} is positioned on the left and serves as the baseline for statistical comparison. The symbols $+$, $-$, and $\approx$ indicate that the result is significantly better, significantly worse, or statistically similar to the baseline (\texttt{PoolSize = 192}). Results are reported as mean (standard deviation). The best mean values in each row are highlighted.}
  {
  
    \renewcommand{\arraystretch}{1.2}
    \begin{tabular}{ccccc}
    \toprule
    Func & PoolSize = 192 (Base) & PoolSize = 20 & PoolSize = 5 & PoolSize = 1 \\
    \midrule
    $F_{1}$ & \multicolumn{1}{l}{\boldmath{\textbf{0.00E+00 (0.00E+00)}}\unboldmath{}} & \multicolumn{1}{l}{\boldmath{\textbf{0.00E+00 (0.00E+00)$\approx$}}\unboldmath{}} & \multicolumn{1}{l}{\boldmath{\textbf{0.00E+00 (0.00E+00)$\approx$}}\unboldmath{}} & \multicolumn{1}{l}{1.33E+01 (4.17E+01)$-$} \\
    $F_{2}$ & \multicolumn{1}{l}{\boldmath{\textbf{0.00E+00 (0.00E+00)}}\unboldmath{}} & \multicolumn{1}{l}{\boldmath{\textbf{0.00E+00 (0.00E+00)$\approx$}}\unboldmath{}} & \multicolumn{1}{l}{\boldmath{\textbf{0.00E+00 (0.00E+00)$\approx$}}\unboldmath{}} & \multicolumn{1}{l}{4.91E+00 (1.57E+01)$-$} \\
    $F_{3}$ & \multicolumn{1}{l}{\boldmath{\textbf{0.00E+00 (0.00E+00)}}\unboldmath{}} & \multicolumn{1}{l}{\boldmath{\textbf{0.00E+00 (0.00E+00)$\approx$}}\unboldmath{}} & \multicolumn{1}{l}{\boldmath{\textbf{0.00E+00 (0.00E+00)$\approx$}}\unboldmath{}} & \multicolumn{1}{l}{2.20E+00 (5.85E+00)$-$} \\
    $F_{4}$ & \multicolumn{1}{l}{\boldmath{\textbf{0.00E+00 (0.00E+00)}}\unboldmath{}} & \multicolumn{1}{l}{\boldmath{\textbf{0.00E+00 (0.00E+00)$\approx$}}\unboldmath{}} & \multicolumn{1}{l}{3.11E-02 (1.73E-01)$-$} & \multicolumn{1}{l}{3.89E+00 (6.23E+00)$-$} \\
    $F_{5}$ & \multicolumn{1}{l}{\boldmath{\textbf{0.00E+00 (0.00E+00)}}\unboldmath{}} & \multicolumn{1}{l}{\boldmath{\textbf{0.00E+00 (0.00E+00)$\approx$}}\unboldmath{}} & \multicolumn{1}{l}{\boldmath{\textbf{0.00E+00 (0.00E+00)$\approx$}}\unboldmath{}} & \multicolumn{1}{l}{1.61E+00 (4.50E+00)$-$} \\
    $F_{6}$ & \multicolumn{1}{l}{\boldmath{\textbf{4.14E-03 (3.57E-03)}}\unboldmath{}} & \multicolumn{1}{l}{6.14E-03 (6.51E-03)$-$} & \multicolumn{1}{l}{9.57E-03 (9.14E-03)$-$} & \multicolumn{1}{l}{9.20E+01 (3.90E+02)$-$} \\
    $F_{7}$ & \multicolumn{1}{l}{\boldmath{\textbf{0.00E+00 (0.00E+00)}}\unboldmath{}} & \multicolumn{1}{l}{\boldmath{\textbf{0.00E+00 (0.00E+00)$\approx$}}\unboldmath{}} & \multicolumn{1}{l}{2.03E-03 (7.20E-03)$-$} & \multicolumn{1}{l}{3.76E+00 (7.66E+00)$-$} \\
    $F_{8}$ & \multicolumn{1}{l}{\boldmath{\textbf{5.30E-04 (1.30E-03)}}\unboldmath{}} & \multicolumn{1}{l}{8.21E-03 (1.07E-02)$-$} & \multicolumn{1}{l}{3.00E-02 (4.19E-02)$-$} & \multicolumn{1}{l}{4.47E+00 (8.63E+00)$-$} \\
    $F_{9}$ & \multicolumn{1}{l}{\boldmath{\textbf{1.75E-06 (0.00E+00)}}\unboldmath{}} & \multicolumn{1}{l}{1.88E+01 (4.64E+01)$-$} & \multicolumn{1}{l}{8.73E+01 (1.08E+02)$-$} & \multicolumn{1}{l}{1.78E+02 (1.04E+02)$-$} \\
    $F_{10}$ & \multicolumn{1}{l}{\boldmath{\textbf{0.00E+00 (0.00E+00)}}\unboldmath{}} & \multicolumn{1}{l}{\boldmath{\textbf{0.00E+00 (0.00E+00)$\approx$}}\unboldmath{}} & \multicolumn{1}{l}{3.05E-05 (1.70E-04)$-$} & \multicolumn{1}{l}{5.34E+01 (5.01E+01)$-$} \\
    $F_{11}$ & \multicolumn{1}{l}{1.38E-05 (2.28E-05)} & \multicolumn{1}{l}{1.45E-05 (3.71E-05)$\approx$} & \multicolumn{1}{l}{\boldmath{\textbf{3.48E-06 (9.45E-06)$+$}}\unboldmath{}} & \multicolumn{1}{l}{1.36E+01 (3.76E+01)$-$} \\
    $F_{12}$ & \multicolumn{1}{l}{\boldmath{\textbf{5.76E+01 (7.15E+01)}}\unboldmath{}} & \multicolumn{1}{l}{7.07E+01 (7.65E+01)$-$} & \multicolumn{1}{l}{9.60E+01 (7.46E+01)$-$} & \multicolumn{1}{l}{1.55E+02 (4.59E+01)$-$} \\
    \midrule
    $+$ / $\approx$ / $-$ & -- & 0/8/4 & 1/4/7 & 0/0/12 \\
    \bottomrule
    \end{tabular}
  
  \label{tab:poolsize_sensitivity_reversed}
  }
\end{table}
\begin{table}[htbp]
  \centering
  \scriptsize
  \caption{Diversity Sensitivity Analysis on the 20-dimensional (20D) CEC 2022 benchmark suite with varying Pool Sizes (Reverse Order). The population size was set to 100,000. \texttt{PoolSize = 192} is positioned on the left and serves as the baseline for statistical comparison. The symbols $+$, $-$, and $\approx$ indicate that the result is significantly better, significantly worse, or statistically similar to the baseline (\texttt{PoolSize = 192}). Results are reported as mean (standard deviation). The best mean values in each row are highlighted.}
  {

    \renewcommand{\arraystretch}{1.2}
    \begin{tabular}{ccccc}
    \toprule
    Func & PoolSize = 192 (Base) & PoolSize = 20 & PoolSize = 5 & PoolSize = 1 \\
    \midrule
    $F_{1}$ & \multicolumn{1}{l}{\boldmath{\textbf{0.00E+00 (0.00E+00)}}\unboldmath{}} & \multicolumn{1}{l}{\boldmath{\textbf{0.00E+00 (0.00E+00)$\approx$}}\unboldmath{}} & \multicolumn{1}{l}{1.07E-09 (5.94E-09)$-$} & \multicolumn{1}{l}{9.12E+02 (2.44E+03)$-$} \\
    $F_{2}$ & \multicolumn{1}{l}{\boldmath{\textbf{0.00E+00 (0.00E+00)}}\unboldmath{}} & \multicolumn{1}{l}{\boldmath{\textbf{0.00E+00 (0.00E+00)$\approx$}}\unboldmath{}} & \multicolumn{1}{l}{4.71E+00 (1.39E+01)$-$} & \multicolumn{1}{l}{5.50E+01 (9.47E+01)$-$} \\
    $F_{3}$ & \multicolumn{1}{l}{\boldmath{\textbf{0.00E+00 (0.00E+00)}}\unboldmath{}} & \multicolumn{1}{l}{\boldmath{\textbf{0.00E+00 (0.00E+00)$\approx$}}\unboldmath{}} & \multicolumn{1}{l}{\boldmath{\textbf{0.00E+00 (0.00E+00)$\approx$}}\unboldmath{}} & \multicolumn{1}{l}{5.70E+00 (1.30E+01)$-$} \\
    $F_{4}$ & \multicolumn{1}{l}{\boldmath{\textbf{5.91E+00 (2.27E+00)}}\unboldmath{}} & \multicolumn{1}{l}{6.90E+00 (1.97E+00)$\approx$} & \multicolumn{1}{l}{7.73E+00 (2.63E+00)$\approx$} & \multicolumn{1}{l}{2.59E+01 (2.11E+01)$-$} \\
    $F_{5}$ & \multicolumn{1}{l}{\boldmath{\textbf{0.00E+00 (0.00E+00)}}\unboldmath{}} & \multicolumn{1}{l}{\boldmath{\textbf{0.00E+00 (0.00E+00)$\approx$}}\unboldmath{}} & \multicolumn{1}{l}{\boldmath{\textbf{0.00E+00 (0.00E+00)$\approx$}}\unboldmath{}} & \multicolumn{1}{l}{7.64E+01 (1.48E+02)$-$} \\
    $F_{6}$ & \multicolumn{1}{l}{9.02E-01 (7.28E-01)} & \multicolumn{1}{l}{\boldmath{\textbf{8.13E-01 (6.85E-01)$\approx$}}\unboldmath{}} & \multicolumn{1}{l}{2.37E+00 (3.31E+00)$-$} & \multicolumn{1}{l}{9.39E+04 (3.23E+05)$-$} \\
    $F_{7}$ & \multicolumn{1}{l}{\boldmath{\textbf{3.24E-01 (3.94E-01)}}\unboldmath{}} & \multicolumn{1}{l}{5.21E-01 (5.82E-01)$\approx$} & \multicolumn{1}{l}{9.90E-01 (1.07E+00)$-$} & \multicolumn{1}{l}{2.05E+01 (2.32E+01)$-$} \\
    $F_{8}$ & \multicolumn{1}{l}{\boldmath{\textbf{4.89E-01 (2.24E-01)}}\unboldmath{}} & \multicolumn{1}{l}{1.02E+00 (3.45E+00)$-$} & \multicolumn{1}{l}{4.49E+00 (7.57E+00)$-$} & \multicolumn{1}{l}{1.50E+01 (1.03E+01)$-$} \\
    $F_{9}$ & \multicolumn{1}{l}{\boldmath{\textbf{1.00E+02 (0.00E+00)}}\unboldmath{}} & \multicolumn{1}{l}{1.10E+02 (2.67E+01)$-$} & \multicolumn{1}{l}{1.48E+02 (3.97E+01)$-$} & \multicolumn{1}{l}{1.93E+02 (6.82E+01)$-$} \\
    $F_{10}$ & \multicolumn{1}{l}{\boldmath{\textbf{0.00E+00 (0.00E+00)}}\unboldmath{}} & \multicolumn{1}{l}{3.27E-03 (1.21E-02)$-$} & \multicolumn{1}{l}{1.28E-01 (3.59E-01)$-$} & \multicolumn{1}{l}{6.59E+01 (5.14E+01)$-$} \\
    $F_{11}$ & \multicolumn{1}{l}{\boldmath{\textbf{7.17E-05 (2.69E-04)}}\unboldmath{}} & \multicolumn{1}{l}{2.89E+01 (8.73E+01)$-$} & \multicolumn{1}{l}{1.00E+02 (1.50E+02)$-$} & \multicolumn{1}{l}{4.11E+02 (5.99E+02)$-$} \\
    $F_{12}$ & \multicolumn{1}{l}{\boldmath{\textbf{2.30E+02 (1.70E+00)}}\unboldmath{}} & \multicolumn{1}{l}{\boldmath{\textbf{2.30E+02 (2.13E+00)$\approx$}}\unboldmath{}} & \multicolumn{1}{l}{2.31E+02 (1.55E+00)$\approx$} & \multicolumn{1}{l}{2.87E+02 (1.29E+02)$-$} \\
    \midrule
    $+$ / $\approx$ / $-$ & -- & 0/8/4 & 0/4/8 & 0/0/12 \\
    \bottomrule
    \end{tabular}
  
  \label{tab:poolsize_sensitivity_20d}
  }
\end{table}
\clearpage

This experiment demonstrates that with a large population size ($N=100,000$), the theoretical advantages of Latin Hypercube Sampling (LHS) over random sampling are largely mitigated by the Law of Large Numbers. Specifically, we analyzed the distribution of $192$ distinct strategy configurations within the iStratDE framework. While LHS theoretically guarantees a more stratified distribution, the results demonstrate that simple random sampling achieves a comparable level of diversity and coverage in high-volume scenarios. Quantitative analysis of the sampling uniformity reveals only a marginal difference in the standard deviation of strategy counts between the two methods (e.g., $21.98$ for random sampling versus $17.97$ for LHS). Consequently, simple random sampling is preferred for its implementation simplicity, as it is robust enough to maintain population diversity with negligible performance differences, thereby rendering the additional computational complexity of LHS unnecessary.

\begin{table}[htbp]
  \centering
    \scriptsize
    \caption{Comparison of Random iStratDE and LHS iStratDE on the 10D CEC 2022 benchmark suite. Random iStratDE denotes the original iStratDE algorithm, which employs random sampling for the initialization of strategies and parameters, whereas LHS iStratDE utilizes Latin Hypercube Sampling (LHS) for initialization. Results are reported as mean (standard deviation). The symbols $+$, $-$, and $\approx$ indicate that the result of LHS iStratDE is significantly better than, significantly worse than, or statistically similar to Random iStratDE based on the Wilcoxon rank-sum test at a 0.05 significance level. The best mean values in each row are highlighted.}
  {
    \renewcommand{\arraystretch}{1.2}
    \begin{tabular}{ccc}
    \toprule
    Func & Random iStratDE & LHS iStratDE \\
    \midrule
    $F_{1}$ & \multicolumn{1}{c}{\boldmath{\textbf{0.00E+00 (0.00E+00)}}\unboldmath{}} & \multicolumn{1}{c}{\boldmath{\textbf{0.00E+00 (0.00E+00)}}\unboldmath{}$\approx$} \\
    $F_{2}$ & \multicolumn{1}{c}{\boldmath{\textbf{0.00E+00 (0.00E+00)}}\unboldmath{}} & \multicolumn{1}{c}{\boldmath{\textbf{0.00E+00 (0.00E+00)}}\unboldmath{}$\approx$} \\
    $F_{3}$ & \multicolumn{1}{c}{\boldmath{\textbf{0.00E+00 (0.00E+00)}}\unboldmath{}} & \multicolumn{1}{c}{\boldmath{\textbf{0.00E+00 (0.00E+00)}}\unboldmath{}$\approx$} \\
    $F_{4}$ & \multicolumn{1}{c}{\boldmath{\textbf{0.00E+00 (0.00E+00)}}\unboldmath{}} & \multicolumn{1}{c}{\boldmath{\textbf{0.00E+00 (0.00E+00)}}\unboldmath{}$\approx$} \\
    $F_{5}$ & \multicolumn{1}{c}{\boldmath{\textbf{0.00E+00 (0.00E+00)}}\unboldmath{}} & \multicolumn{1}{c}{\boldmath{\textbf{0.00E+00 (0.00E+00)}}\unboldmath{}$\approx$} \\
    $F_{6}$ & \multicolumn{1}{c}{6.30E-03 (3.11E-03)} & \multicolumn{1}{c}{\boldmath{\textbf{4.48E-03 (3.27E-03)}}\unboldmath{}$+$} \\
    $F_{7}$ & \multicolumn{1}{c}{\boldmath{\textbf{0.00E+00 (0.00E+00)}}\unboldmath{}} & \multicolumn{1}{c}{\boldmath{\textbf{0.00E+00 (0.00E+00)}}\unboldmath{}$\approx$} \\
    $F_{8}$ & \multicolumn{1}{c}{5.14E-04 (1.29E-03)} & \multicolumn{1}{c}{\boldmath{\textbf{3.68E-04 (1.37E-03)}}\unboldmath{}$\approx$} \\
    $F_{9}$ & \multicolumn{1}{c}{\boldmath{\textbf{1.75E-06 (0.00E+00)}}\unboldmath{}} & \multicolumn{1}{c}{\boldmath{\textbf{1.75E-06 (0.00E+00)}}\unboldmath{}$\approx$} \\
    $F_{10}$ & \multicolumn{1}{c}{\boldmath{\textbf{0.00E+00 (0.00E+00)}}\unboldmath{}} & \multicolumn{1}{c}{\boldmath{\textbf{0.00E+00 (0.00E+00)}}\unboldmath{}$\approx$} \\
    $F_{11}$ & \multicolumn{1}{c}{1.31E-05 (1.74E-05)} & \multicolumn{1}{c}{\boldmath{\textbf{7.91E-06 (1.17E-05)}}\unboldmath{}$\approx$} \\
    $F_{12}$ & \multicolumn{1}{c}{\boldmath{\textbf{4.90E+01 (6.95E+01)}}\unboldmath{}} & \multicolumn{1}{c}{5.77E+01 (7.51E+01)$\approx$} \\
    \midrule
    $+$ / $\approx$ / $-$ & --    & 1/11/0 \\
    \bottomrule
    \end{tabular}
  
  \label{tab:istratde_10D}
  }
\end{table}
\begin{table}[htbp]
  \centering
  \scriptsize
    \caption{Comparison of Random iStratDE and LHS iStratDE on the 20D CEC 2022 benchmark suite. Random iStratDE denotes the original iStratDE algorithm, which employs random sampling for the initialization of strategies and parameters, whereas LHS iStratDE utilizes Latin Hypercube Sampling (LHS) for initialization. Results are reported as mean (standard deviation). The symbols $+$, $-$, and $\approx$ indicate that the result of LHS iStratDE is significantly better than, significantly worse than, or statistically similar to Random iStratDE based on the Wilcoxon rank-sum test at a 0.05 significance level. The best mean values in each row are highlighted.}
  {
    \renewcommand{\arraystretch}{1.2}
    \begin{tabular}{ccc}
    \toprule
    Func & Random iStratDE & LHS iStratDE \\
    \midrule
    $F_{1}$ & \multicolumn{1}{c}{\boldmath{\textbf{0.00E+00 (0.00E+00)}}\unboldmath{}} & \multicolumn{1}{c}{\boldmath{\textbf{0.00E+00 (0.00E+00)}}\unboldmath{}$\approx$} \\
    $F_{2}$ & \multicolumn{1}{c}{\boldmath{\textbf{0.00E+00 (0.00E+00)}}\unboldmath{}} & \multicolumn{1}{c}{\boldmath{\textbf{0.00E+00 (0.00E+00)}}\unboldmath{}$\approx$} \\
    $F_{3}$ & \multicolumn{1}{c}{\boldmath{\textbf{0.00E+00 (0.00E+00)}}\unboldmath{}} & \multicolumn{1}{c}{\boldmath{\textbf{0.00E+00 (0.00E+00)}}\unboldmath{}$\approx$} \\
    $F_{4}$ & \multicolumn{1}{c}{7.15E+00 (1.87E+00)} & \multicolumn{1}{c}{\boldmath{\textbf{6.47E+00 (1.99E+00)}}\unboldmath{}$\approx$} \\
    $F_{5}$ & \multicolumn{1}{c}{\boldmath{\textbf{0.00E+00 (0.00E+00)}}\unboldmath{}} & \multicolumn{1}{c}{\boldmath{\textbf{0.00E+00 (0.00E+00)}}\unboldmath{}$\approx$} \\
    $F_{6}$ & \multicolumn{1}{c}{\boldmath{\textbf{8.28E-01 (9.41E-01)}}\unboldmath{}} & \multicolumn{1}{c}{9.33E-01 (8.44E-01)$\approx$} \\
    $F_{7}$ & \multicolumn{1}{c}{3.76E-01 (4.71E-01)} & \multicolumn{1}{c}{\boldmath{\textbf{3.53E-01 (5.07E-01)}}\unboldmath{}$\approx$} \\
    $F_{8}$ & \multicolumn{1}{c}{\boldmath{\textbf{1.07E+00 (3.47E+00)}}\unboldmath{}} & \multicolumn{1}{c}{1.68E+00 (4.79E+00)$\approx$} \\
    $F_{9}$ & \multicolumn{1}{c}{\boldmath{\textbf{1.00E+02 (0.00E+00)}}\unboldmath{}} & \multicolumn{1}{c}{\boldmath{\textbf{1.00E+02 (0.00E+00)}}\unboldmath{}$\approx$} \\
    $F_{10}$ & \multicolumn{1}{c}{\boldmath{\textbf{0.00E+00 (0.00E+00)}}\unboldmath{}} & \multicolumn{1}{c}{\boldmath{\textbf{0.00E+00 (0.00E+00)}}\unboldmath{}$\approx$} \\
    $F_{11}$ & \multicolumn{1}{c}{\boldmath{\textbf{9.38E+00 (5.22E+01)}}\unboldmath{}} & \multicolumn{1}{c}{\boldmath{\textbf{9.38E+00 (5.22E+01)}}\unboldmath{}$\approx$} \\
    $F_{12}$ & \multicolumn{1}{c}{\boldmath{\textbf{2.30E+02 (1.90E+00)}}\unboldmath{}} & \multicolumn{1}{c}{2.30E+02 (1.56E+00)$\approx$} \\
    \midrule
    $+$ / $\approx$ / $-$ & --    & 0/12/0 \\
    \bottomrule
    \end{tabular}
  
  \label{tab:istratde_20D}
  }
\end{table}
\clearpage

\section{Application on Robotic Control Tasks}
\begin{figure}[htbp]
    \centering
    \begin{minipage}[b]{0.24\textwidth}
        \centering
        \includegraphics[height=3cm]{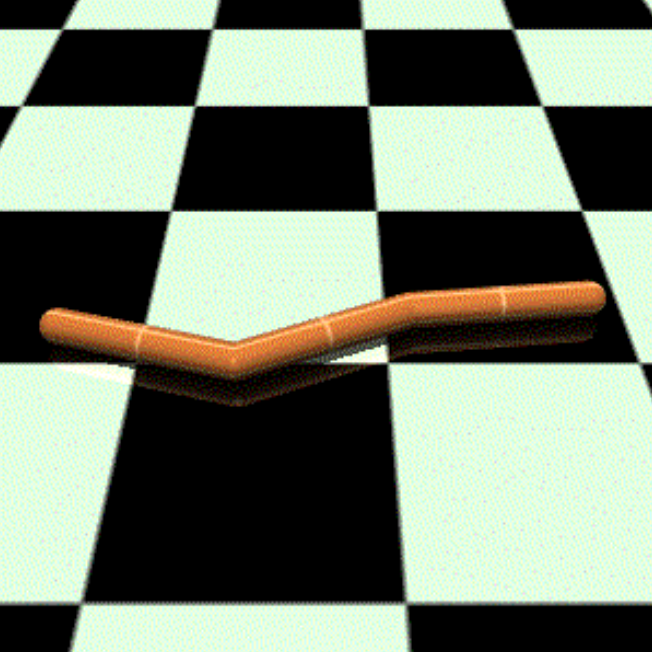}
        \caption*{Swimmer}
    \end{minipage}
    \hspace{0.125\textwidth}
    \begin{minipage}[b]{0.24\textwidth}
        \centering
        \includegraphics[height=3cm]{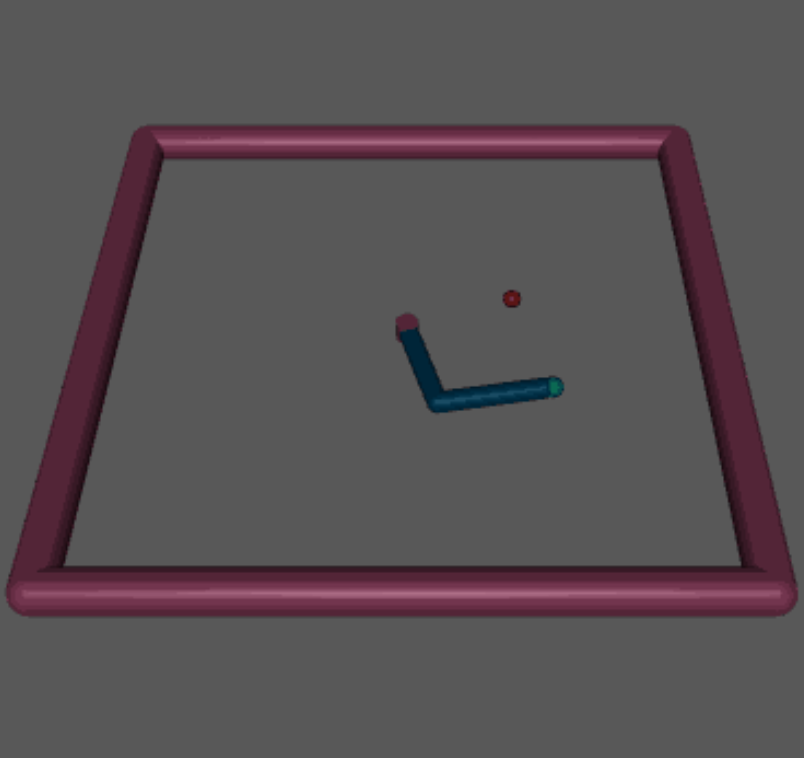}
        \caption*{Reacher}
    \end{minipage}
    \hspace{0.125\textwidth}
    \begin{minipage}[b]{0.24\textwidth}
        \centering
        \includegraphics[height=3cm]{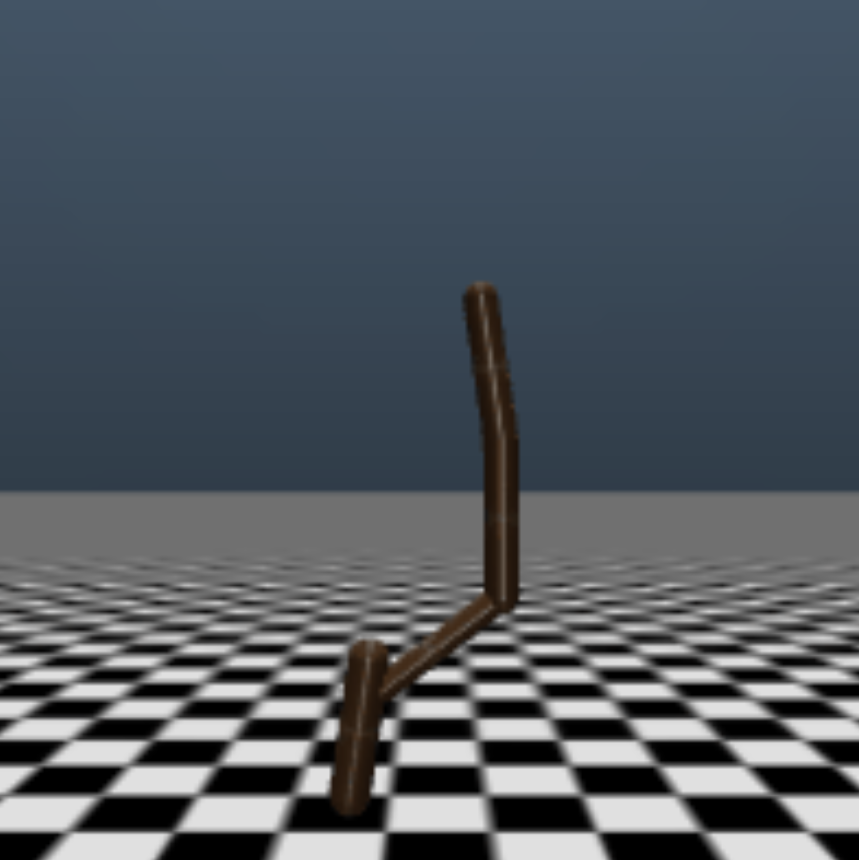}
        \caption*{Hopper}
    \end{minipage}
    \caption{Rendered environments of the three robotic control tasks: Swimmer, Reacher, and Hopper.}
    \label{fig:robotics_tasks} 
\end{figure}

The robotic control tasks in our experiments are implemented in the \emph{Brax} framework, 
a physics engine that supports fast and parallelizable simulation on accelerators such as GPUs and TPUs. 
Brax provides efficient environments for reinforcement learning research, which makes it suitable for testing evolutionary optimization methods. 
In these tasks, the objective is to maximize the cumulative reward, i.e., a larger fitness value indicates better performance. 
Each episode is capped at a maximum length of 500 time steps. 
All three tasks involve policy networks with approximately 1,500 parameters, which makes them high-dimensional optimization problems for evolutionary algorithms. 
We evaluate iStratDE on three representative continuous control tasks: \emph{Swimmer}, \emph{Hopper}, and \emph{Reacher}. 
Rendered images of these environments are shown in Fig.~\ref{fig:robotics_tasks}, and the specific architectures of their policy networks are detailed in Table~\ref{tab:Neural network structures}.

\begin{enumerate}[label=\arabic*)]
    \item \textbf{Swimmer:} The Swimmer task requires a snake-like agent with multiple joints to propel itself forward in a two-dimensional plane by generating coordinated oscillatory motions. 
    The goal is to maximize forward velocity while maintaining stability in movement. 
    The observation space has 8 dimensions, and the action space has 2 dimensions.
    \item \textbf{Reacher:} The Reacher task consists of a two-joint robotic arm that must reach a randomly sampled target position within a bounded workspace. 
    The objective is to minimize the distance between the arm’s end effector and the target location while maintaining smooth control. 
    The observation space has 11 dimensions, and the action space has 2 dimensions.
    \item \textbf{Hopper:} The Hopper task involves a one-legged robot that must learn to hop forward without falling. 
    This task is particularly challenging due to its underactuated dynamics, which require precise balance control and efficient energy use. 
    The observation space has 11 dimensions, and the action space has 3 dimensions.
\end{enumerate}\begin{table}[htbp]
\centering
\caption{Architectures of the policy networks used in the robotics control tasks.}
\label{tab:Neural network structures}
\begin{tabular}{cccccc}
\toprule
\textbf{Task} & \textbf{Dimension} & \textbf{Input} & \textbf{Hidden Layers} &   \textbf{Output}    & \textbf{Activation} \\
\midrule
Swimmer & 1410  & 8     & 32$\times$32 & 2     & \texttt{tanh} \\
Reacher & 1506  & 11    & 32$\times$32 & 2     & \texttt{tanh} \\
Hopper & 1539  & 11    & 32$\times$32 & 3     & \texttt{tanh} \\
\bottomrule
\end{tabular}%
\end{table}\begin{table}[htbp]
\centering
\caption{Rewards achieved by iStratDE and peer evolutionary algorithms on the Brax robotics control tasks (Swimmer, Hopper, and Reacher) with a 60-minute runtime. Values represent the mean reward and standard deviation (in parentheses) over multiple runs. The best mean rewards are highlighted in bold.}
\label{tab:comparative-reward-analysis-scientific}
{%
\resizebox{\textwidth}{!}{
\renewcommand{\arraystretch}{1.2}
\renewcommand{\tabcolsep}{2pt}
\begin{tabular}{ccccccccc}
\toprule
Task & \textbf{iStratDE} & \textbf{DE} & \textbf{CSO} & \textbf{CMA-ES} & \textbf{SHADE}&\textbf{CoDE}&\textbf{LSHADE-RSP}&\textbf{EDEV} \\
\midrule
Swimmer & \textbf{1.91E+02 (2.55E+00)} & 1.80E+02 (1.98E+00) & 1.83E+02 (1.10E+00) & 1.85E+02 (9.10E-01) & 1.84E+02 (2.33E+00) & 1.78E+02 (2.19E+00) & 1.85E+02 (2.36E+00) & 1.50E+02 (3.32E+01) \\
Hopper & \textbf{1.34E+03 (1.37E+02)} & 8.29E+02 (9.66E+01) & 1.18E+03 (3.58E+02) & 1.29E+03 (4.61E+02) & 9.88E+02 (1.12E+02) & 9.72E+02 (1.35E+02) & 1.06E+03 (1.40E+02) & 3.78E+02 (1.19E+02) \\
Reacher & -2.66E+01 (2.30E+01) & -5.43E+02 (9.51E+01) & -4.33E+02 (1.29E+02) & \textbf{-3.91E+00 (1.08E+00)} & -2.32E+02 (1.46E+02) & -4.37E+02 (1.34E+02) & -3.86E+02 (1.47E+02) & -4.92E+02 (1.04E+02) \\
\bottomrule
\end{tabular}%
}
}
\end{table}

\end{document}